\documentclass[12pt]{report}
\usepackage[letterpaper,top=2cm,bottom=2cm,left=2cm,right=2cm,marginparwidth=1.75cm,includeheadfoot]{geometry}

\usepackage[english]{babel}

\usepackage[autostyle]{csquotes}

\usepackage{multicol}
\usepackage{multirow}
\usepackage{comment}
\usepackage{array}
\usepackage{hyperref}
\usepackage{tabularx}
\usepackage{booktabs}
\usepackage{xspace} %
\usepackage{flushend}
\usepackage{color}
\usepackage{graphicx}

\usepackage{marginnote} 

\newcolumntype{M}[1]{>{\arraybackslash}m{#1}}
\newcolumntype{C}[1]{>{\centering \arraybackslash}m{#1}}
\newcolumntype{N}{@{}m{0pt}@{}}

\usepackage{amsthm}
\usepackage{amsmath}
\usepackage{amsfonts}
\usepackage{amssymb}
\usepackage{mathtools}
\usepackage{mathrsfs}

\usepackage{bbold}  %
\usepackage{wrapfig}
\usepackage{mdframed}

\usepackage{times}

\usepackage{caption}

\usepackage{tikz}
\usetikzlibrary{fit,calc,arrows,positioning,shapes,automata}

\usepackage{url}

\usepackage{doi}

\newcommand{\argmin}[1]{\ensuremath{\underset{#1}{\textrm{arg min}}\;}}
\newcommand{\argmax}[1]{\ensuremath{\underset{#1}{\textrm{arg max}}\;}}

\usepackage[numbers,sort&compress]{natbib}

\usepackage[textsize=footnotesize]{todonotes}
\usepackage{textcomp} %
\usepackage{nicefrac}       %
\usepackage{siunitx}

\usepackage{graphics}
\usepackage{tikz, pgfplots, pgfplotstable}

\usepackage{graphicx}
\usepackage{subcaption}

\DeclareSIUnit\px{px}

\tikzset{>=latex}

\pgfdeclarelayer{background}
\pgfsetlayers{background,main}

\pgfplotsset{compat=1.15}
\usetikzlibrary{shapes, shapes.geometric, arrows, calc, 3d, chains, arrows.meta, bending, quotes}
\usepgfplotslibrary{colorbrewer, colormaps}

\pgfplotsset{every axis/.style = {cycle list/Spectral}}

\newcommand{\quotes}[1]{``#1''}

\definecolor{mygreen}{HTML}{D2691E}
\definecolor{myred}{HTML}{ee0000}
\definecolor{mysepia}{HTML}{704214}
\definecolor{turquoise}{HTML}{10d59c}

\usepackage{acronym}
\acrodef{DTA}{Duckietown Autolab}
\acrodef{AI-DO}{AI Driving Olympics}
\acrodef{LF}{lane-following}
\acrodef{DUCKIENet}{Decentralized Urban Collaborative Benchmarking Network}
\acrodef{MPD}{Mean Position Deviation}
\acrodef{MOD}{Mean Orientation Deviation}
\acrodef{ML}{machine learning}

\newcommand{\expfigure}[5]{
\centering
    \begin{minipage}[c]{#1\textwidth}
        \centering
        \includegraphics[width=\textwidth,keepaspectratio=true]{#3}
    \end{minipage}%
    \begin{minipage}[c]{#2\textwidth}
        \centering
        \includegraphics[width=\textwidth,keepaspectratio=true]{#4}
    \end{minipage}
    \vspace{1mm}

    \footnotesize
    \begin{tabular}{@{}cccccccc@{}}
    \toprule
 Robots & Location & Repetitions & \begin{tabular}[c]{@{}c@{}}AVG\\ MPD\end{tabular} & \begin{tabular}[c]{@{}c@{}}STD\\ MPD\end{tabular} & \begin{tabular}[c]{@{}c@{}}AVG\\ MOD\end{tabular} & \begin{tabular}[c]{@{}c@{}}STD\\ MOD\end{tabular} \\ \midrule
    #5 \bottomrule
    \end{tabular}
}

\newcommand{\exptablesamebot}{%
 1 & ETHZ & 9 & -6.2 & 1.3 & -0.8 & 2.8 \\
}

\newcommand{\exptablediffbot}{%
 3 & ETHZ & $3\times3$ & 1.1 & 3.4 & -0.4 & 5.2 \\
}

\newcommand{\exptablediffsub}{%
 $1 \times 2$ & ETHZ / TTIC & $6\times2$ & 2.2 & 2.5 & -0.4 & 3.9 \\
}

\newcommand{\nlgsec}{Enabling robots to help humans navigate a priori unknown environments}
\newcommand{\nlusec}{Multimodal learning framework to estimate articulated objects models}
\newcommand{\autolabsec}{Reproducible and Accessible evaluation of robotic agents}
\newcommand{\aidosec}{The AI Driving Olympics at NeurIPS 2018}
\newcommand{\moocsec}{MOOC: Self-Driving Cars with Duckietown}
\newcommand{\sharcsec}{SHARC: SHared Autonomy for Remote Collaboration}

\acrodef{IDE}{Integrated Development Environment}
\acrodef{CMS}{Content Management System}
\acrodef{ToF}{Time-of-Flight}
\acrodef{IMU}{Inertial Measurement Unit}
\acrodef{ROS}{{R}obot {O}perating {S}ystem}
\acrodef{MOOC}{Massive Open Online Course}
\acrodef{NLP}{Natural Language Processing}
\acrodef{NLG}{Natural Language Generation}
\acrodef{NLU}{Natural Language Understanding}
\acrodef{ROV}{Remotely Operated Vehicle}
\acrodef{AUV}{Autonomous Underwater Vehicle}
\acrodef{XRF}{X-Ray Fluorescense}
\acrodef{VR}{Virtual Reality}
\acrodef{IRL}{Inverse Reinforcement Learning}

\title{Accessible Interfaces for the Development and Deployment of Robotic Platforms}
\author{Andrea F. Daniele\\ TTI-Chicago}

\begin{document}
    \definecolor{ttic_blue}{RGB}{55,93,137}
\begin{tikzpicture}[overlay,remember picture]
    \draw [line width=1mm] [ttic_blue]
        ($ (current page.north west) + (0.9in,-0.9in) $)
        rectangle
        ($ (current page.south east) + (-0.9in,0.9in) $);
\end{tikzpicture}
\begin{center}
\Large \MakeUppercase{{\bf Accessible Interfaces\\ for the Development and Deployment\\ of Robotic Platforms}}\\
\normalsize
\vspace{0.25in}
by\\Andrea F. Daniele\\
\vspace{1in}
A thesis submitted\\
in partial fulfillment of the requirements for\\
the degree of\\
\vspace{0.3in}
Doctor of Philosophy in Computer Science\\
\vspace{0.3in}
at the\\
\vspace{0.3in}
\MakeUppercase{Toyota Technological Institute at Chicago}\\
Chicago, Illinois\\
\vspace{0.3in}
January, 2023\\
\vspace{1.0in}
Thesis Committee:\\
\vspace{0.1in}
Matthew R.\ Walter (Thesis Advisor)\\
Gregory Shakhnarovich\\
Liam Paull\\
Richard Camilli\\
Stefanie Tellex\\
\end{center}
\thispagestyle{empty}
\newpage

    \pagenumbering{Roman}

    \chapter*{Acknowledgments}
\addcontentsline{toc}{chapter}{Acknowledgments}

Pursuing this Ph.D. was perhaps the best decision of my life. 
While academic achievements, paper publications, and research contributions are 
things to celebrate and be proud of, this Ph.D. was for me, first and foremost, 
\textbf{the opportunity} to step out of my comfort zone and face my demons. \\

My Ph.D. advisor, Matthew Walter, to whom I am incredibly grateful, is the person who gave 
me that opportunity, at a time in my academic career when I had nothing to show.
And while I am very thankful for his academic guidance and knowledge, I am most grateful for
his friendship and moral support throughout the years.\\

Thank you to all the incredible people I met at TTIC, from my fellow students to the faculty, 
admins, and staff, you quickly became my second family. Thank you to my colleagues in the robotics lab, 
especially to Zhongtian (Falcon) Dai and Charles (Chip) Schaff, inextinguishable sources of 
patience and wisdom.\\

Thank you to my committee for generously providing your time and feedback.\\

Thank you to my parents and the rest of my family. I worked hard to achieve this milestone, but
you worked twice as hard just to make sure I had what I needed to succeed.\\

Finally, thank you to my wife, Erika. Your love and support through hard and uncertain times 
were an unwavering beacon of light.

    \chapter*{Abstract}
\addcontentsline{toc}{chapter}{Abstract}

Accessibility is one of the most important features in the design of robots and their interfaces. 
Accessible interfaces allow untrained users to easily and intuitively tap into the full potential 
of a robotic platform. This thesis proposes methods that improve the accessibility of robots for 
three different target audiences: consumers, researchers, and learners.

In order for humans and robots to work together effectively, they both must be able to communicate 
with each other to convey information, establish a shared understanding of their collaborative tasks, 
and to coordinate their efforts. Natural languages offer a flexible, bidirectional, bandwidth-efficient 
medium that humans can readily use to interact with their robotic companions. 
We work on the problem of enabling robots to produce natural language utterances as well as understand them.
In particular, we tackle the problem of generating route instructions that are readily understandable 
by novice humans for the navigation of a priori unknown indoor environments. Our model first decides 
which information to share with the user, using a policy trained from human demonstrations via \acf{IRL}.
It then ``translates'' this information into a natural language instruction using a neural 
sequence-to-sequence model. 

We then move on to consider the related problem of enabling robots to 
understand natural language utterances in the context of learning to interact with their environment. 
In particular, we are interested in enabling robots to operate articulated objects (e.g., fridges, drawers) 
by leveraging kinematic models. We propose a multimodal learning framework that incorporates both vision and 
language acquired in situ, where we model linguistic information using a probabilistic graphical model that 
grounds natural language descriptions to their referent kinematic motion. 
Our architecture then fuses the two modalities to estimate the structure and parameters that define kinematic 
models of articulated objects.

We then turn our focus to the development of accessible and reproducible robotic platforms for scientific research. 
The majority of robotics research is accessible to all but a limited audience and usually takes place in idealized 
laboratory settings or unique uncontrolled environments. 

Owing to limited reproducibility, the value of a specific result is either open to interpretation or conditioned 
on specific environments or setups. In order to address these limitations, we propose a new concept for reproducible
robotics research that integrates development and benchmarking, so that reproducibility is obtained ``by design'' from 
the beginning of the research and development process. We first provide the overall conceptual objectives to achieve 
this goal and then a concrete instance that we have built: the DUCKIENet. We validate the system by analyzing the
repeatability of experiments and show that there is low variance across different robot hardware and labs. 

We then propose a framework, called SHARC (SHared Autonomy for Remote Collaboration), to improve accessibility for
underwater robotic intervention operations. Conventional underwater robotic manipulation requires a team of scientists
on-board a support vessel to instruct the pilots on the operations to perform. This limits the number of scientists 
that can work together on a single expedition, effectively hindering a robotic platform’s accessibility and driving 
up costs of operation. On the other hand, shared autonomy allows us to leverage human capabilities in perception and
semantic understanding of an unstructured environment, while relying on well-tested robotic capabilities for precise 
low-level control. SHARC allows multiple remote scientists to efficiently plan and execute high-level sampling procedures
using an underwater manipulator while deferring low-level control to the robot. A distributed architecture allows 
scientists to coordinate, collaborate, and control the robot while being on-shore and at thousands of kilometers away 
from one another and the robot.

Lastly, we turn our attention to the impact of accessible platforms in the context of educational robotics. 
While online learning materials and \acp{MOOC} are great tools for educating large audiences,
they tend to be completely virtual. Robotics has a hardware component that cannot and must not be ignored or replaced 
by simulators. This motivated our work in the development of the first hardware-based MOOC in AI and robotics. 
This course, offered for free on the edX platform, allows learners to study autonomy hands-on by making real robots 
make their own decisions and accomplish broadly defined tasks. We design a new robotic platform from the ground up 
to support this new learning experience. A fully browser-based experience, based on leading tools and technologies 
for code development, testing, validation, and deployment (e.g., ROS, Docker, VSCode), serves to maximize the 
accessibility of these educational resources.

    \tableofcontents

\listoffigures

\listoftables

    \newpage

    \pagenumbering{arabic}

    \chapter{Introduction}
    \label{sec:introduction}
    There are many important properties to consider when designing a new robotic platform or interface.
This work revolves around the strong belief that \textbf{accessibility} is one of them.

In this context, the word ``accessibility'' does not only refer to the user-friendliness of a robot 
or interface for untrained users.
In fact, accessible robots and their interfaces are not just easier to use, but easier to predict, debug, study with, 
develop on, and share.
This thesis proposes methods, systems, and tools that improve the accessibility of robots for 
three different target audiences: \textit{consumers}, \textit{researchers}, and \textit{learners}.

\section{Robot's accessibility for consumers}

Consumers are usually not expected to have any past experience or scientific background 
in robotics. Meanwhile, increasingly complex robots are being deployed in contexts where they are required 
to seamlessly work alongside people, whether it is serving as assistants in our homes~\citep{walters07}, 
helping students with language learning in the classroom~\cite{kanda04}, 
and acting as guides in public spaces~\cite{hayaski07}. In order for humans and robots to work together 
effectively, they both must be able to communicate with each other to convey information, establish a shared 
understanding of their collaborative task, or to coordinate their efforts~\cite{grosz96, fong01, clair15, sauppe15}.
Natural languages offer a flexible, bandwidth-efficient medium that humans can readily use to interact with 
their robotic companion~\cite{sung15}. 

Natural language interaction is also bidirectional, which means that robots must be able to express their 
beliefs and intentions by producing natural language utterances as well as parse and understand the beliefs 
and intentions verbally conveyed by its human partners.
The first problem is known as \acf{NLG}, while the second is known as \acf{NLU} and they constitute two 
sub-fields of the more general field known as \acf{NLP}.
Chapter~\ref{hri2017:sec:main} and~\ref{isrr2017:sec:main} will cover some of our past work in the fields 
of \acs{NLG} and \acs{NLU}, respectively.

\subsection{\nlgsec}
Modern robotics applications that involve human-robot interaction
require robots to be able to communicate with humans seamlessly
and effectively. Natural language provides a flexible and
efficient medium through which robots can exchange information
with their human partners. 
Significant advancements have been made
in developing robots capable of interpreting free-form
instructions, but less attention has been devoted to endowing
robots with the ability to generate natural language. 

In this work~\cite{daniele17}, we propose a
model that enables robots to generate natural
language instructions that allow humans to navigate a priori
unknown environments. We first decide which information to share
with the user according to their preferences, using a policy
trained from human demonstrations via inverse reinforcement
learning. We then ``translate'' this information into a natural
language instruction using a neural sequence-to-sequence model
that learns to generate free-form instructions from natural
language corpora. We evaluate our method on a benchmark route
instruction dataset and achieve a BLEU score of $72.18$\% 
compared to human-generated reference instructions. We
additionally conduct navigation experiments with human
participants demonstrating that our method generates
instructions that people follow as accurately and easily as those
produced by humans.

For more details, see Chapter~\ref{hri2017:sec:main}.

\subsection{\nlusec}
In order for robots to operate effectively in homes and workplaces, they must be able to manipulate the articulated objects common within environments built for and by humans. Kinematic models provide a concise representation of these objects that enable deliberate, generalizable manipulation policies. However, existing approaches to learning these models rely upon visual observations of an object's motion, and are subject to the effects of occlusions and feature sparsity. Natural language descriptions provide a flexible and efficient means by which humans can provide complementary information in a weakly supervised manner suitable for a variety of different  interactions (e.g., demonstrations and remote manipulation). 

In this work~\cite{daniele17a}, we present a multimodal learning framework that incorporates both vision and language information acquired in situ to estimate the structure and parameters that define kinematic models of articulated objects. The visual signal takes the form of an RGB-D image stream that opportunistically captures object motion in an unprepared scene. Accompanying natural language descriptions of the motion constitute the linguistic signal.  We model linguistic information using a probabilistic graphical model that grounds natural language descriptions to their referent kinematic motion. By exploiting the complementary nature of the vision and language observations, our method infers correct kinematic models for various multiple-part objects on which the previous state-of-the-art, visual-only system fails.  We evaluate our multimodal learning framework on a dataset comprised of a variety of household objects, and demonstrate a $23\%$ improvement in model accuracy over the vision-only baseline.

For more details, see Chapter~\ref{isrr2017:sec:main}.

\section{Robot's accessibility in scientific research}

The development of accessible and reproducible robotic platforms is also crucial for many scientific research 
communities. The majority of robotics research is accessible to all but a limited audience and usually takes 
place in idealized laboratory settings or unique uncontrolled environments. 

Owing to limited reproducibility, the value of a specific result is either open 
to interpretation or conditioned on specifics of the setup that are not necessarily reported as part of the 
presentation. As a result, the overwhelming majority of benchmarks and competitions in \ac{ML} do not involve 
physical robots. Yet, the interactive embodied setting is thought to be an essential scenario to study 
intelligence~\cite{pfeifer2001understanding, floreano2004evolution}.
Chapter~\ref{iros2020:sec:main} covers our work on the development of a platform for reproducible and accessible 
evaluation of robotic agents. In Chapter~\ref{aido2018:sec:main}, we put these ideas to the test, 
by reporting our findings from the 
``AI Driving Olympics'' (AI-DO), an international competition with the objective of evaluating the state of the 
art in machine learning and artificial intelligence for mobile robotics. 

Moreover, scientists from various fields employ robots to carry out experiments that are either too dangerous 
or unfeasible for humans to perform (e.g., underwater exploration, cave exploration).
In some of these scenarios, accessibility is not hindered by the poor design of the robots or their interfaces, but by
the logistical resources needed to support scientists and their experiments on the field.
The development of new technologies and tools aimed at easing these logistical constraints can have a significant 
impact on the costs, speed of execution, quality, and safety of an experiment.  
In Chapter~\ref{sharc2021:sec:main} we talk about
our work on the development of a distributed shared autonomy framework, called
SHARC, for the execution of deep underwater robotic manipulation procedures by a remote, geographically distributed 
team of scientists.

\subsection{\autolabsec}
As robotics matures and increases in complexity, it is more necessary than ever that robot autonomy research be \textit{reproducible}. Compared to other sciences, there are specific challenges to benchmarking autonomy, such as the complexity of the software stacks, the variability of the hardware and the reliance on data-driven techniques, amongst others. 

In this work~\cite{tani20}, we describe a new concept for reproducible robotics research that integrates development and benchmarking, so that reproducibility is obtained \quotes{by design} from the beginning of the research/development processes. We first provide the overall conceptual objectives to achieve this goal and then a concrete instance that we have built: the DUCKIENet. One of the central components of this setup is the Duckietown Autolab, a remotely accessible standardized setup that is itself also relatively low-cost and reproducible.
When evaluating agents, careful definition of interfaces allows users to choose among local versus remote evaluation using simulation, logs, or remote automated hardware setups.
We validate the system by analyzing the repeatability of experiments conducted using the DUCKIENet and show that there is low variance across different robot hardware and across different labs.

For more details, see Chapter~\ref{iros2020:sec:main}.

\subsection{\aidosec}
Despite recent breakthroughs, the ability of deep learning and reinforcement 
learning to outperform traditional approaches to control physically embodied 
robotic agents remains largely unproven.

To help bridge this gap, we present the ``AI Driving Olympics'' (AI-DO), 
a competition with the objective of evaluating the state of the art in machine 
learning and artificial intelligence for mobile robotics. 
Based on the simple and well-specified autonomous driving and navigation 
environment called ``Duckietown,'' the AI-DO includes a series of tasks of 
increasing complexity -- from simple lane-following to fleet management. 
For each task, we provide tools for competitors to use in the form of simulators, 
logs, code templates, baseline implementations and low-cost access to robotic hardware. 
We evaluate submissions in simulation online, on standardized hardware environments, 
and finally at the competition event. The first AI-DO, AI-DO 1, occurred at the 
Neural Information Processing Systems (NeurIPS) conference in December 2018. 
In this work~\cite{aido2018}, we describe the AI-DO 1, including the motivation and design 
objections, the challenges, the provided infrastructure, an overview of the 
approaches of the top submissions, and a frank assessment of what worked well 
as well as what needs improvement.

The results of AI-DO 1 highlight the need for better benchmarks, which are 
lacking in robotics, as well as improved mechanisms to bridge the gap between 
simulation and reality.

For more details, see Chapter~\ref{aido2018:sec:main}.

\subsection{\sharcsec}
Conventional robotic underwater intervention operations consists of one or more 
\acp{ROV} that are deployed to the ocean seafloor by a support vessel and 
tele-operated by pilots.
A team of scientists must also be present on-board the vessel
to instruct the pilots on the operations to perform as part of a 
scientific experiment or procedure.

Fully autonomous underwater manipulation has so far only been achieved in testing tanks and structured environments. In the short-term, field operations can benefit from recent advancements in the field of shared autonomy.
Shared autonomy allows us to leverage human capabilities in perception and semantic understanding of an unstructured environment, while relying on well-tested robotic 
capabilities for low-level precision control.

In this work, we propose SHARC: SHared Autonomy for Remote Collaboration framework.
SHARC allows multiple remote scientists to efficiently plan and execute high-level sampling procedures using an underwater manipulator while deferring low-level control to the robot. A distributed architecture allows scientists to coordinate, 
collaborate, and control the robot while being on-shore and at thousands of 
kilometers away from one another and the robot.
We tested SHARC in a field expedition, during which a group of three scientists
was able to perform the first in-situ XRF sampling of seafloor sediments while
being completely off-ship and geographically distributed.

For more details, see Chapter~\ref{sharc2021:sec:main}.

\section{Robot's accessibility in education}

In educational robotics, accessibility is perhaps the most impactful factor on a 
learner's experience.

Recent studies~\cite{us-homeschooling, us-homeschooling-trend, us-charachter-schools} in the US show a growing 
trend of people receiving education outside of traditional institutions (e.g., homeschooling, charter schools, 
study groups, continuing education). 
Increasing demographics and cost of education worldwide are factors suggesting that demand for alternative 
educational systems will likely keep increasing.

Nowadays, a wide variety of learning material and educational resources are available online. 
While online learning materials and \acp{MOOC} are great tools for teaching to a large audience, they tend to 
be completely virtual.
Robotics has a hardware component that cannot and must not be ignored or replaced by simulators.
In Chapter~\ref{mooc2021:sec:main} we cover our work on the development of the first hardware-based 
\ac{MOOC} in AI and robotics.

\subsection{\moocsec}
Autonomy and AI are all around us, revolutionizing our daily lives. In particular, autonomous vehicles have a huge potential to impact 
society in the near future.
With an exponential growth in number of potential applications comes the need to train a new generation of \acf{ML} engineers, researchers, and roboticists.

In March 2021, the Duckietown Foundation released the first hardware-based \acf{MOOC} in AI and robotics called ``Self-driving cars with Duckietown''. The course is offered for free on edX~\footnote{``Self-driving cars with Duckietown'' on edX: \url{https://www.edx.org/course/self-driving-cars-with-duckietown}}.
The course allows learners to study autonomy hands-on by making real robots make their own decisions and accomplish broadly defined tasks.
Step-by-step from the theory, to the implementation, to the deployment in simulation as well as on real robots.
A new robotic platform was specifically designed from the ground up to support this new learning experience.

In November 2022, the second edition of the course ``Self-driving cars with Duckietown'' was released. 
In an effort to further improve the accessibility of these educational resources, 
the second edition of the course featured a fully browser-based experience, based on leading tools and 
technologies for code development, testing, validation, and deployment (e.g., ROS, Docker, VSCode). 

For more details, see Chapter~\ref{mooc2021:sec:main}

    \newpage
    \chapter{\nlgsec}
    \label{hri2017:sec:main}
    This work was published in~\cite{daniele17}.\\
    
\begin{wrapfigure}{r}{0.5\textwidth}
  \centering
  \begin{tabular}{ M{8.4cm} N}
    \toprule
    \textbf{Input:} map and path & \\[3pt]
    \centering
    \vspace{0.1cm}
  	\includegraphics[width=0.92\linewidth]{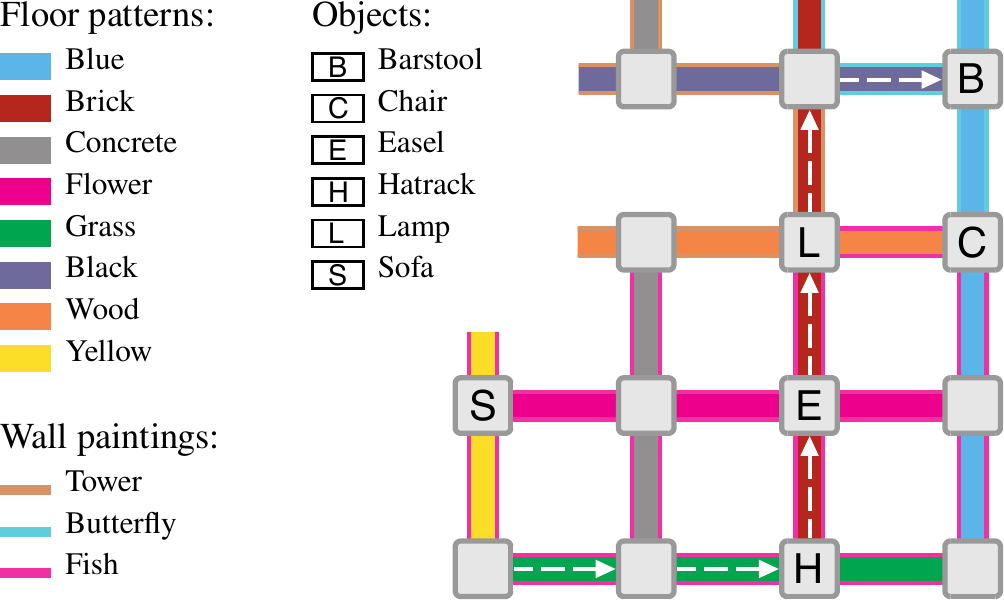}
  	\vspace{0.1cm}
  	& \\
    \midrule
    \textbf{Output:} route instruction & \\[3pt]
    ``turn to face the grass hallway. walk forward twice. face the easel. move until you see black floor to your right. face the stool. move to the stool''
  	& \\[26pt]
    \bottomrule
  \end{tabular}
  \caption{An example route instruction that our framework
    generates for the shown map and path.\label{hri2017:fig:generation_example}}
\end{wrapfigure}
Robots are increasingly being used as our partners, working with
and alongside people, whether it is serving as assistants
in our homes~\citep{walters07}, transporting cargo in
warehouses~\cite{correa10}, helping students with language learning in the
classroom~\cite{kanda04}, and acting as guides in public
spaces~\cite{hayaski07}. In order for 
humans and robots to work together effectively, robots 
must be able to communicate
with their
human partners to convey information, establish a shared understanding of their
collaborative task, or to coordinate their efforts~\cite{grosz96,
  fong01, clair15, sauppe15}.
Natural language provides an efficient,
flexible medium through which humans and robots can exchange
information. 
Consider, for example, a search-and-rescue operation
carried out by a human-robot team. 
The human may first issue spoken
commands (e.g., ``Search the rooms at the end of the hallway'') that
direct one or more robots to navigate throughout the building
searching for occupants~\cite{matuszek10, tellex2011understanding,
  mei2016listen}. In this process, the robot may engage the user in
dialogue to resolve any ambiguity in the task (e.g., to clarify which
hallway the user was referring to)~\cite{tellex12a, deits13, raman13,
  tellex2014asking, hemachandra15a}. The user's ability to trust their
robotic partners is also integral to effective collaboration~\cite{groom07}, and a
robot's ability to generate natural language explanations of its
progress (e.g., ``I have inspected two rooms'') and decision-making
processes have been shown to help establish trust~\cite{dzindolet03,
andrist13, wang16}.

In this work, we specifically consider the surrogate problem of
synthesizing natural language route instructions and describe a method
that generates free-form directions that people can accurately and
efficiently follow in environments unknown to them a priori 
(Figure~\ref{hri2017:fig:generation_example}). This problem has previously
been considered by the robotics community~\cite{morales2011modeling,goeddel2012dart,
oswald2014learning} and is important for human-robot
collaborative tasks, such as search-and-rescue, exploration, and
surveillance~\cite{kunze14}, and for robotic assistants, such as those
that serve as guides in museums, offices, etc. More generally, the problem is relevant beyond human-robot
interaction to the broader domain of indoor navigation. 
There are two primary challenges to generating effective, natural 
language route instructions, which are characteristic of the more general 
problem of free-form generation. 

The first challenge is \emph{content selection}, the problem of deciding what 
and how much information to convey to the user as part of the
directions. In general, the more detailed an instruction is, the less
ambiguous it is. However, verbose instructions can be unnatural and hard for
followers to remember and, thus, ineffective. Consequently, it is
important to balance the value of including particular information as
part of a route instruction with the cost that comes with
increasing the level of detail. Further, not all information is
equally informative. Existing commercial navigational solutions typically rely 
on a set of hand-crafted rules that consider only street names and
metric distances as valid candidates, the latter of which requires
that follower's keep track of their progress. In contrast, studies have shown that
people prefer route instructions that reference physical, salient
landmarks in the environment~\cite{waller2007landmarks}. However, no
standard exists with regards to what and how these landmarks should be
selected, as these depend on the nature of the environment and the
demographics of the follower~\cite{ward86, hund2012impact}. 

We propose a method that models the content selection problem as a
Markov decision process with a learned policy that decides what and
how much to include in a formal language specification of the task
(path). We learn this policy via inverse reinforcement learning from
demonstrations of route instructions provided by humans. This avoids
the need for hand-crafted selection rules, and allows our method to 
adapt to the preferences and communication style of the target
populations and to simultaneously choose to convey information that
minimizes the ambiguity of the instruction while avoiding verbosity.

The second challenge is \emph{surface realization}, which is
the task of synthesizing a natural language sentence that refers to
the selected content. Existing solutions rely on sentence templates,
generating sentences by populating manually defined fields (e.g.,
``turn $\langle$direction$\rangle$'') and then serializing these
sentences in a turn-by-turn fashion. As expected, the use of such templates reduces
coherence across sentences and limits the ability to adapt to
different domains (e.g., from outdoor to indoor navigation).  Additionally, while
the output is technically correct, the resulting sentences tend to be
rigid and unnatural.  Studies show that language generated by a robot
is most effective when it emulates the communication style that
people use~\cite{torrey13}. 

We address the surface realization problem through a neural
sequence-to-sequence model that ``translates'' a formal language
specification of the selected command into a natural language
sentence. Our model takes the form of an encoder-aligner-decoder
architecture that first encodes the formal path specification with a
recurrent neural network using long short-term memory
(LSTM-RNN)~\cite{hochreiter97}. The model then decodes (translates) the 
resulting abstraction of the input into a natural language sentence (word
sequence), using an alignment mechanism to further refine the selected
information and associate output words with the corresponding elements
in the input formal specification. The use of LSTMs as the hidden units
enables our model to capture the long-term dependencies that exist
among the selected information and among the words in the resulting
instruction. 
We train our surface realization model on
instruction corpora, enabling our method to generate free-form
 directions that emulate the style of human instructions, without 
 templates, specialized features, or linguistic resources.

We evaluate our method on the benchmark SAIL dataset of human-generated route
instructions~\cite{macmahon2006walk}. Instructions generated with our
method achieve a sentence-level BLEU score of $72.18\%$, indicating their similarity
with the reference set of human-provided instructions. We perform
a series of ablations and visualizations to better understand the
contributions of the primary components of our model. We additionally
conduct human evaluation experiments that demonstrate that our
method generates instructions that people are able to follow as
efficiently and accurately as those generated by humans. 

Our method is, to the best of our knowledge, the first framework
that generates natural language instructions that reflect
the preferences and rhetorical style of humans. We achieve this firstly via
 a decision process that learns to emulate human content selection
preferences from demonstrations; and secondly through a neural machine translation
model that learns to express this content via natural language. Human experiments reveal that our method
produces instructions that are as accurate, effective, and usable as
those generated by humans.

\section{Related Work}
\label{hri2017:sec:related_work}

Existing research related to the generation of route instructions
spans the fields of robotics, natural language processing, cognitive
science, and psychology. Early work in this area focuses on
understanding the way in which humans generate natural language route
instructions~\cite{ward86, allen97, lovelace1999elements} and the properties
that make ``good'' instructions easier for people to
follow~\cite{look2005location, richter2008simplest, waller2007landmarks}. 
These studies have shown that people prefer to
give directions as a sequence of turn-by-turn instructions and that
they favor physical objects and locations as intuitive landmarks.

Based on these studies, much of the existing research on
generating route instructions involves the use of hand-crafted rules
that are designed to emulate the manner in which people compose
navigation instructions~\cite{striegnitz11, curry15}. \citet{look2005location} 
compose route instructions using a set of templates and application rules 
engineered based upon a corpus of human-generated route instructions.
\citet{look2008cognitively} improves upon this work by incorporating
human cognitive spatial models to generate high-level route overviews
that augment turn-by-turn directions. Similarly,~\citet{dale2005using}
analyze a dataset of route instructions composed by people to derive a
set of hand-designed rules that mimic the content and style of human
directions. \citet{goeddel2012dart} describe a particle filter-based
method that employs a generative model of direction following to
produce templated  instructions that maximize the likelihood of
reaching the desired destination.
Meanwhile, \citet{morales2011modeling} propose a
system that enables robots to provide route instructions using a
combination of language and pointing gestures. Unlike our approach,
their method is limited to paths with no more than three direction changes  and employs four generation
templates, limiting the content and structure of the instructions.

The challenge with instruction generation systems that rely upon
hand-crafted rules is that it is difficult to design a policy that
generalizes to a wide variety of scenarios and followers, whose
preferences vary depending on such factors as their cultural
background~\cite{hund2012impact} and
gender~\cite{ward86}. \citet{cuayahuitl10} seek to improve upon this
using reinforcement learning with hand-crafted reward functions that
model the length of the instructions and the likelihood that they will
confuse a follower. They then learn a policy that reasons both over
the best route and the corresponding navigational
instructions. However, this approach still requires that domain
experts define the reward functions and specify model parameters. In
contrast,~\citet{oswald2014learning} model the problem of deciding
what to include in the instruction (i.e., the content selection
problem) as a Markov decision process and learn a policy from a
human-generated navigation corpus using maximum entropy inverse
reinforcement learning. Given the content identified by the policy,
their framework does not perform surface realization, and instead
generates instructions by matching the selected content with the
nearest match in a database of human-generated instructions. Our
method also uses inverse reinforcement learning for content
selection, but unlike their system, our method also learns to perform
surface realization directly from corpora, thus generating newly-composed 
natural language instructions.

Relatedly, much attention has been paid recently to the ``inverse''
problem of learning to follow (i.e., execute) natural language route
instructions. Statistical methods primarily formulate the problem of
converting instructions to actions as either a semantic parsing 
task~\cite{matuszek10,chen11,artzi2013weakly} or as a symbol grounding
problem~\cite{kollar2010toward, tellex2011understanding, landsiedel13,
howard14a, chung15}. Alternatively, \citet{mei2016listen} translate
free-form instructions to action sequences in an end-to-end fashion 
using an encoder-aligner-decoder.

Meanwhile, selective generation considers the more general problem of
converting a rich database to a natural language utterance, with
existing methods generally focusing on the individual problems of
content selection and surface realization.  \citet{barzilay2004catching}
perform content selection on collections of unannotated documents for
the sake of text summarization. \citet{barzilay2005collective}
formulate content selection as a collective classification problem,
simultaneously optimizing local label assignments and their pairwise
relations. \citet{liang2009learning} consider the related problem of
aligning elements of a database to textual description clauses. They
propose a generative semi-Markov model that simultaneously segments
text into utterances and aligns each utterance with its corresponding
entry in the database. Meanwhile, \citet{walker01} perform surface
realization via sentence planners that can be trained to generate
sentences for dialogue and context
planning. \citet{wong2007generation} effectively invert a semantic
parser to generate natural language sentences from formal meaning
representations using synchronous context-free grammars. Rather than
consider individual sub-problems, recent work focuses on solving
selective generation via a single framework~\cite{chen2008learning,
  kim2010generative, angeli2010simple, konstas2012unsupervised,
  mei2016what}. \citet{angeli2010simple} model  content selection
and surface realization as local decision problems via log-linear
models and employ templates for generation. \citet{mei2016what}
formulate selective generation as an end-to-end learning problem and
propose a recurrent neural network encoder-aligner-decoder model that
jointly learns to perform content selection and surface realization
from database-text pairs. Unlike our method, they do not reason
over the correctness or unambiguity of the summaries, nor do they
attempt to model human content selection preferences.

\section{Task Definition}
\label{hri2017:sec:task_definition}

We consider the problem of generating natural language instructions
that allow humans to navigate environments that are unknown to them a
priori.
The given map $m$ takes the form of a hybrid
metric-topologic-semantic representation
(Figure~\ref{hri2017:fig:generation_example}) that encodes the position of and
connectivity between a dense set of locations in the environment
(e.g., intersections) and the position and type of objects and
environment features (e.g., floor patterns). The path $p$ is a sequence
of poses (i.e., position and orientation) that
corresponds to the minimum distance route from a
given initial pose to a desired goal pose. We split the path
according to changes in direction, representing the path
$p = (p_1, p_2, \ldots p_M)$ as a sequence of intermediate segments
$p_i$.

Training data comes in the form of tuples $(m^{(i)}, p^{(i)},
\Lambda^{(i)})$ for $i=1, 2, \ldots, n$ drawn from human
demonstrations, where $m^{(i)}$ is a map of the environment,
$\Lambda^{(i)}$ is a human-generated natural language route
instruction, and $p^{(i)}$ is the path that a different human took
when following the instructions.
At test time, we consider only the
map and path pair as known and hold out the human-generated
instruction for evaluation. The dataset that we use for training,
validation, and testing comes from the benchmark SAIL
corpus~\cite{macmahon2006walk}.
\section{Model}
\label{hri2017:sec:model}

\begin{figure}[t]
  \centering
  \includegraphics[width=0.8\textwidth]{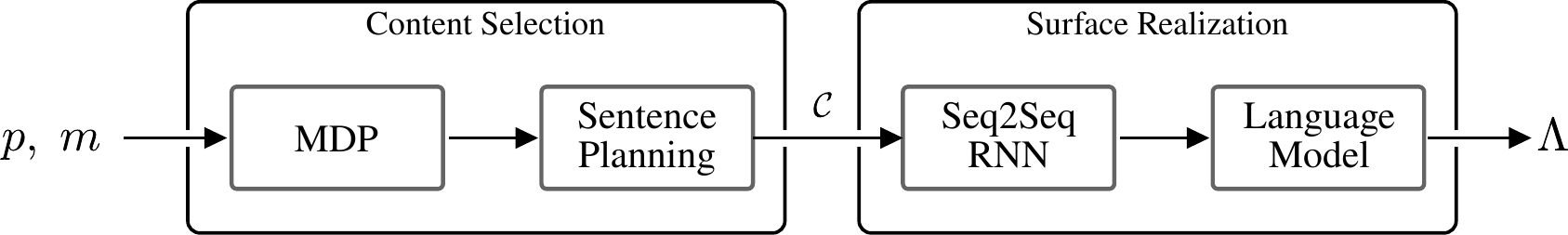}
  \caption{Our method generates natural language instructions for a
    given map and path.} \label{hri2017:fig:framework}
\end{figure}

Given a map and a path, our framework (Figure~\ref{hri2017:fig:framework})
performs content selection to decide what
information to share with the human follower and subsequently performs
surface realization to generate a natural language instruction
according to this selected content. Our method learns to perform
content selection and surface realization from human demonstrations,
so as to produce instructions that are similar to those generated by
humans.

Our framework only
assumes knowledge of the user's pose and
can be readily implemented on a mobile robot or body-worn device that
provides natural language guidance as the user navigates,
re-planning and issuing corrections as necessary. By maintaining
probabilistic models of information content and ambiguity, our method is amenable to incorporating user feedback (e.g.,
utterances and gaze) to revise the information shared
with the user (e.g., referencing landmarks based upon gaze). By
using inverse reinforcement learning, our method can also adapt the content selection policy
based on feedback to better learn user preferences.

\subsection{Compound Action Specifications} \label{hri2017:sec:cas}

In order to bridge the gap between the low-level nature of the input
paths and the natural language output, we encode paths using an
intermediate logic-based formal language. Specifically, we use the
Compound Action Specification (CAS)
representation~\cite{macmahon2006walk}, which provides a formal
abstraction of navigation commands for hybrid
metric-topologic-semantic maps. The CAS language consists
of five \textit{actions} (i.e., Travel, Turn, Face, Verify, and
Find), each of which is associated with a number of
\textit{attributes} (e.g.,
Travel.distance, Turn.direction) that together define specific commands. We
distinguish between CAS \emph{structures}, which are instructions with
the attributes left empty (e.g.,
\textit{Turn(direction=\textbf{None})}) thereby defining a class of
instructions, and CAS \emph{commands}, which correspond to
instantiated instructions with the attributes set to particular values
(e.g., \textit{Turn(direction=\textbf{Left})}).
For each English instruction $\Lambda^{(i)}$ in the dataset, we generate
the corresponding CAS command $c^{(i)}$ using the MARCO architecture 
\cite{macmahon2006walk}. For a complete
description of the CAS language, see~\citet{macmahon2006walk}.

\subsection{Content Selection} \label{hri2017:sec:content_selection}

There are many ways in which one can compose a CAS representation  of a
desired path, both in terms of the type of information that is conveyed 
(e.g., referencing
distances vs.\ physical landmarks), as well as the specific references
to use (e.g., different objects provide candidate landmarks). Humans
exhibit common preferences in terms of the type of information that is
shared (e.g., favoring visible landmarks over
distances)~\cite{waller2007landmarks}, yet the specific nature of this
information depends upon the environment and the followers'
demographics~\cite{ward86, hund2012impact}. Our goal is to learn these
preferences from a dataset of instructions generated by humans.

\subsubsection{MDP with Inverse Reinforcement Learning} \label{hri2017:sec:irl}

In similar fashion to \citet{oswald2014learning}, we formulate the content 
selection problem as a Markov decision
process (MDP) with the goal of then identifying an information selection
policy that maximizes long-term cumulative reward consistent with
human preferences (Figure~\ref{hri2017:fig:framework}). However, this reward
function is unknown a priori and is generally difficult to define. We
assume that humans optimize a common reward function when composing
instructions and employ inverse reinforcement learning (IRL) to learn a
policy that mimics the preferences that humans exhibit based upon a
set of human demonstrations.

An MDP is defined by the tuple $(S, A, R, P, \gamma)$, where $S$ is a
set of states, $A$ is a set of actions,
$R(s, a, s^\prime) \in \mathbb{R}$ is the reward received when
executing action $a \in A$ in state $s \in S$ and transitioning to
state $s^\prime \in S$, $P(s^\prime \vert a,s)$ is the probability of
transitioning from state $s$ to state $s^\prime$ when executing action
$a$, and $\gamma \in (0,1]$ is the discount factor. The policy
$\pi(a \vert s)$ corresponds to a distribution over actions given the
current state.  In the case of the route instruction domain, the state
$s$ defines the user's pose and path in the context of the map of the
environment. We represent the state in terms of $14$ \emph{context}
features that express characteristics such as changes in orientation
and position, the relative location of objects, and nearby environment
features (e.g., different floor colors). We encode the state $s$ as a
$14$-dimensional binary vector that indicates which context features
are active for that state. In this way, the state space $S$ is spanned 
by all possible instantiations of context features. Meanwhile, the action
space corresponds to the space of different CAS structures (i.e.,
without attribute values)  that can be used to define the path.

We seek a policy $\pi(a \vert s)$ that maximizes expected cumulative
reward. However, the reward function that defines the value of
particular characteristics of the instruction is unknown and difficult
to define. For that reason, we frame the task as an inverse
reinforcement learning  problem using human-provided route
instructions as demonstrations of the optimal policy. Specifically, we learn a
policy using the maximum entropy formulation of IRL~\cite{ziebart08},
which models user actions as a distribution over paths parameterized
as a log-linear model $P(a; \theta) \propto e^{-\theta^\top \xi(a)}$,
where $\xi(a)$ is a feature vector defined over actions and $\theta$
is a parameter vector. We consider $9$
instruction features (\emph{properties})  that
include features expressing the number of landmarks included in the
instruction, the frame of reference that is used, and the complexity
of the command. The feature vector $\xi(a)$ then takes the form of a
$9$-dimensional binary vector. 
Maximum entropy IRL then solves for
the distribution as
\begin{equation}
    \begin{split}
        &P(a; \theta^*) = \underset{\theta}{\text{arg max}} \; P(a; \theta) \log P(a; \theta)\\
        &\text{s.t.} \; \xi_g = \mathbb{E}[\xi(a)],
    \end{split}
\end{equation}
where $\xi_g$ denotes the features from the demonstrations and the
expectation is taken over the action distribution. For further details
regarding maximum entropy IRL, we refer the reader to \citet{ziebart08}.

\begin{figure*}[t]
    \centering
    \includegraphics[width=1.0\linewidth]{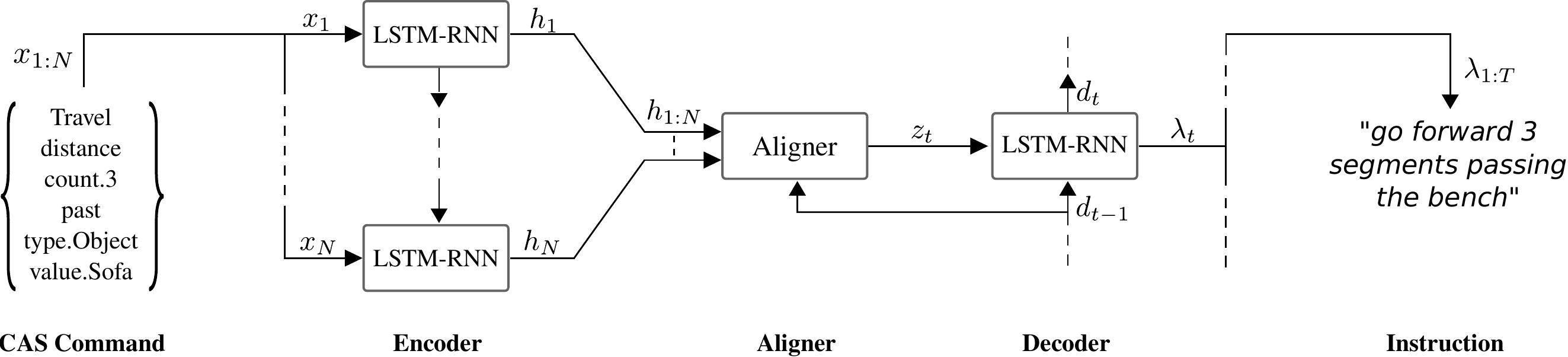}
    \caption{Our encoder-aligner-decoder model for surface realization.
    \label{hri2017:fig:seq2seq_model}}
\end{figure*}

The policy defines a distribution over CAS structure compositions (i.e., using the
\emph{Verify} action vs.\ the \emph{Turn} action) in terms of their feature encoding. 
We perform inference over this policy to identify the maximum a
posteriori property vector $\xi(a^*) = \text{arg max}_{\xi} \,
\pi$. As there is no way to invert the feature mapping, we then match
this vector $\xi(a^*)$ to
a database of CAS structures formed from our
training set. Rather than choosing the nearest match, which may result
in an inconsistent CAS structure, we retrieve the $k_c$ nearest
neighbors from the database using a weighted distance in terms of
mutual information~\cite{oswald2014learning} that expresses the
importance of different CAS features based upon the context. As
several of these may be valid, we employ spectral clustering using the
similarity of the CAS strings to
identify a set of candidate CAS structures $\mathcal{C}_s$.

\subsubsection{Sentence Planning} \label{hri2017:sec:sentence_planning}

Given the set of candidate CAS structures $\mathcal{C}_s$, our method next
chooses the attribute
values such that the final CAS commands are both valid and not
ambiguous. We express the
likelihood that a command $c \in \mathcal{C}_s$ is a valid representation
of a particular path
$p$ defined on a map $m$ as
\begin{equation} \label{hri2017:eqn:probability_cas}
    P( c \vert p, m ) = 
    \frac{ \delta( c \vert p, m ) }
    { \sum_{j=1}^{K} \delta( c \vert \hat{p}_j, m ) }.
\end{equation}
The index $j$ iterates over all  possible paths that have the same
starting pose as path $p$. We define $\delta( c \; | \; p, m )$ as
\begin{equation*}
    \delta( c \vert p, m ) = 
        \left\{
        \begin{array}{ll}
          1 & \text{if} \;\; \eta( c ) = \phi( c, p, m ) \\
          0 & \text{otherwise}
        \end{array}
        \right.
\end{equation*}
where $\eta( c )$ is the number of attributes defined in the command $c$, and
$\phi( c, p, m )$ is the number of attributes in the command $c$ that are
valid relative to the path $p$ and map $m$.

For each candidate CAS structure $c \in \mathcal{C}_s$, we generate multiple 
CAS commands by iterating over the possible attributes values. We evaluate the
correctness and ambiguity of each configuration according to
Equation~\ref{hri2017:eqn:probability_cas}. A command is deemed valid if its
likelihood is greater than a threshold $P_t$. Since the number of possible 
configurations for a
structure increases exponentially with respect to the number of
attributes, we assign attributes using greedy search. The iteration 
is constrained to use only objects and properties of the environment
visible to the follower. The result is a set $\mathcal{C}$ of valid CAS commands.

\subsection{Surface Realization} \label{hri2017:sec:surface_realization}

Having identified a set of CAS commands suitable to the given path,
our method then proceeds to generate the corresponding natural
language route instruction. We formulate this problem as one of
``translating'' the instruction specification in the formal CAS
language into its natural language equivalent.\footnote{Related
  work~\cite{matuszek10,artzi2013weakly,mei2016listen} similarly models the inverse
  task of language understanding as a machine translation problem.} 
We perform this translation using an encoder-aligner-decoder model
(Figure~\ref{hri2017:fig:seq2seq_model}) that enables our framework to generate
natural language instructions by learning from examples of
human-generated instructions, without the need for specialized
features, resources, or templates.

\subsubsection{Sequence-to-Sequence Model} \label{hri2017:sec:seq2seq}

We formulate the problem of generating natural language route
instructions as inference over a probabilistic model $P(\lambda_{1:T}
\vert x_{1:N})$, where $\lambda_{1:T} = (\lambda_1, \lambda_2, \ldots,
\lambda_T)$ is the sequence of words in the instruction and
$x_{1:N} = (x_1, x_2, \ldots x_N)$ is the sequence of tokens in the
CAS command. The CAS sequence includes a token for each action (e.g.,
$Turn$, $Travel$) and a set of tokens with the form \emph{attribute.value} for 
each attribute-value pair; for example, 
\emph{Turn(direction=Right)} is represented by the sequence 
(\emph{Turn}, \emph{direction.Right}). Generating an instruction sequence then corresponds to
inference over this model.
\begin{subequations} \label{hri2017:eqn:argmax}
\begin{align}
    \lambda_{1:T}^* &= 
    	\underset{\lambda_{1:T}}{\textrm{arg max }} P(\lambda_{1:T}|x_{1:N})\\
    &= 
    	\underset{\lambda_{1:T}}{\textrm{arg max }} \prod\limits_{t=1}^T
    	P(\lambda_t \vert \lambda_{1:t-1}, x_{1:N})
\end{align}
\end{subequations}

We formulate this task as a sequence-to-sequence learning problem, whereby
we use a recurrent neural network (RNN) to first encode the input
CAS command 
\begin{subequations}
   \begin{align}
     h_j &= f(x_j, h_{j-1})\\
     z_t &= b(h_1, h_2, \ldots h_N),
   \end{align}
\end{subequations}
where $h_j$ is the encoder hidden state for CAS token $j$, and $f$ and $b$ are
nonlinear functions that we define later.  An aligner computes the context vector $z_t$ that
encodes the language instruction at time $t \in \{1,\ldots,T\}$.
An RNN decodes the context vector $z_t$ to arrive at the desired
likelihood (Eqn.~\ref{hri2017:eqn:argmax})
\begin{equation} \label{hri2017:eqn:cond_prob}
      P(\lambda_t \vert \lambda_{1:t-1}, x_{1:N}) = g(d_{t-1}, z_t),
\end{equation}
where $d_{t-1}$ is the decoder hidden state at time $t-1$, and $g$ is a nonlinear
function. \\

\textbf{Encoder}   Our encoder (Figure~\ref{hri2017:fig:seq2seq_model}) takes as input the sequence
of tokens in the CAS command $x_{1:N}$. We transform each token $x_i$ into
a $k_e-$dimensional binary vector using a word embedding representation 
\cite{mikolov2013distributed}. We feed this sequence into an RNN
encoder that employs LSTMs as the recurrent unit due to their
ability to learn long-term dependencies among the instruction
sequences without being prone to vanishing or exploding
gradients. The LSTM-RNN encoder summarizes the relationship between
elements of the CAS command and yields a sequence of hidden states
$h_{1:N} = (h_1, h_2, \ldots, h_N)$, where $h_j$ encodes CAS words up
to and including $x_j$. In practice, we reverse the input sequence
before feeding it into the neural encoder, which has been demonstrated
to improve performance for other neural translation tasks 
\cite{sutskever14}.

Our encoder  is similar to that of \citet{graves13}
\begin{subequations} \label{hri2017:eqn:encoder}
    \begin{align}
      \begin{pmatrix}
          i^{e}_j\\ 
          f^{e}_j\\
          o^{e}_j\\ 
          g^{e}_j 
      \end{pmatrix} 
   &= 
     \begin{pmatrix}
         \sigma\\
         \sigma\\
         \sigma\\ 
         \tanh 
     \end{pmatrix}
      T^{e} 
      \begin{pmatrix}
          x_j\\ 
          h_{j-1}
      \end{pmatrix}\\
      c^{e}_j&=f^{e}_j \odot c^{e}_{j-1} + i^{e}_j \odot g^{e}_j \\
      h_j &=o^{e}_j \odot \tanh(c^{e}_j) \label{hri2017:eqn:encoder_h}
    \end{align}
\end{subequations}
where $T^{e}$ is an affine transformation, $\sigma$ is the logistic sigmoid
that restricts its input to $[0,1]$, $i^{e}_j$, $f^{e}_j$, and $o^{e}_j$
are the input, output, and forget gates of the LSTM, respectively, and
$c^e_j$ is the memory cell activation vector.
The memory cell $c^e_j$
summarizes the LSTM's previous memory $c^e_{j-1}$ and the current input,
which are modulated by the forget and input gates, respectively. \\

\textbf{Aligner}  Having encoded the input CAS command into a sequence
of hidden annotations $h_{1:N}$, the decoder then seeks to generate a
natural language instruction as a sequence of words.
We employ an alignment mechanism~\cite{bahdanau14} (Figure~\ref{hri2017:fig:seq2seq_model}) that
permits our model to match and focus on particular elements of the CAS sequence
that are salient to the current word in the output
instruction. We compute the context vector as
\begin{equation} \label{hri2017:eqn:attention_model}
    z_t = \sum_{j}\alpha_{tj} h_j.
\end{equation}
The weight $\alpha_{tj}$ associated with the j-th hidden state is
\begin{equation}
    \alpha_{tj} = \exp(\beta_{tj})/\sum_k \exp(\beta_{tk}),
\end{equation}
where the alignment term $\beta_{tj}=f(d_{t-1},h_j)$ expresses the
degree to which the CAS element at position $j$ and those around it
match the output at time $t$. The term $d_{t-1}$ represents the decoder 
hidden state at the previous 
time step. The alignment is modeled as a one-layer neural perceptron
\begin{equation}
  \beta_{tj} = v^\top\tanh(Wd_{t-1}+Vh_{j}),
\end{equation}
where $v$, $W$, and $V$ are learned parameters.\\

\textbf{Decoder}  Our model employs an LSTM decoder
(Figure~\ref{hri2017:fig:seq2seq_model}) that takes as input the
context vector $z_t$ and the decoder hidden state at the previous time step 
$d_{t-1}$, and outputs the conditional probability distribution 
\mbox{$P_{\lambda,t} = P(\lambda_t \vert \lambda_{1:t-1}, x_{1:N})$}
over the next token as a deep output layer
\begin{subequations}
    \begin{align}
      \begin{pmatrix}
          i^d_t\\
          f^d_t\\
          o^d_t\\
          g^d_t
      \end{pmatrix}
   &= 
     \begin{pmatrix}
         \sigma\\ 
         \sigma\\ 
         \sigma\\ 
         \tanh 
     \end{pmatrix} T^d 
      \begin{pmatrix}
          d_{t-1}\\ 
          z_t
      \end{pmatrix}\\
      c^d_t&=f^d_t \odot c^d_{t-1} + i^d_t \odot g^d_t \\
      d_{t}&=o_t^d \odot \tanh(c^d_t)\\
      q_t &= L_0 ( L_dd_t + L_zz_t )\\
      P_{\lambda,t} &= \textrm{softmax}\left(q_t\right)
    \end{align}
\end{subequations}
where $L_0$, $L_d$, and $L_z$ are parameters to be learned.\\

\textbf{Training}  We train our encoder-aligner-decoder
model so as to predict the natural language instruction $\lambda_{1:T}^*$ for a given 
input sequence $x_{1:N}$ using a training set of human-generated
reference instructions. We use the negative
log-likelihood of the reference instructions at each time step $t$ as
our loss function.
\\

\textbf{Inference}  Given a CAS command represented as a sequence of
tokens $x_{1:N}$, we generate a route instruction as the sequence of
maximum a posteriori words $\lambda_{1:T}^*$  under our learned
model (Eqn.~\ref{hri2017:eqn:argmax}). We use beam search to perform approximate
inference, but have empirically found greedy search to often perform
better.\footnote{This phenomenon has been observed by
  others~\cite{angeli2010simple, mei2016what}, and we attribute it to
  training the model in a greedy fashion.} For that reason, we generate
candidates using both greedy and beam search.

\subsubsection{Language Model} \label{hri2017:sec:language_model}

The inference procedure results in multiple candidate instructions for
a given segment, and additional candidates may exist when there are
multiple CAS specifications. We rank these candidate instructions
using a language model (LM) trained on large amounts of English data. 
We formulate this LM as an LSTM-RNN~\cite{sundermeyer2012lstm} 
that assigns a perplexity score
to each of the corresponding instructions. 

Given the CAS specifications for a segmented path $p = (p_1, p_2,
\ldots p_M)$, we generate the final instruction $\Lambda$  by
sequencing the $M$ sentences 
$\{\Lambda_1^\star,\ldots,\Lambda_M^\star\}$ (i.e., one for each path segment)
\begin{equation}
    \Lambda_i^\star = \underset{\Lambda_{ij}}{\textrm{arg min }} L(\Lambda_{ij}),
\end{equation}
where $\Lambda_{ij}$ is the j-th candidate for the i-th segment and $L(\Lambda_{ij})$
is the perplexity score assigned by the language model to the sentence $\Lambda_{ij}$.
\section{Experimental Setup}
\label{hri2017:sec:experiment_setup}

\subsection{Dataset} \label{hri2017:sec:dataset}

We train and evaluate our system using the publicly available SAIL
route instruction dataset collected by~\citet{macmahon2006walk}. We
use the original data without correcting typos or wrong
instructions (e.g., confusing ``left'' and ``right''). The dataset 
consists of $3213$ demonstrations arranged in
$706$ paragraphs produced by 6 instructors for $126$ different paths
throughout $3$ virtual environments, where each demonstration provides
a map-path-command tuple $(m^{(i)}, p^{(i)},
\Lambda^{(i)})$.
We partition the dataset into separate training ($70\%$), validation 
($10\%$), and test ($20\%$) sets. 
We utilize command-instruction pairs $(c^{(i)},
\Lambda^{(i)})$ from the training, 
validation and test sets for training, hyper-parameter
tuning, and testing of our encoder-aligner-decoder model, respectively. 
We use path-command pairs $(p^{(i)},
c^{(i)})$ from the training set for IRL and pairs from the
validation set to tune the hyper-parameters of the content selection 
model. Finally, we use path-instruction pairs $(p^{(i)},
\Lambda^{(i)})$ from the test 
set for evaluations with human participants.

\subsubsection{Data Augmentation}

The SAIL dataset is significantly smaller than those typically used to
train neural sequence-to-sequence models. In order to overcome this
scarcity, we augment the original dataset using a set of rules.
We use the augmented dataset to train the neural 
encoder-aligner-decoder model and the original dataset to 
train the content selection model.

\subsection{Implementation Details}
\label{hri2017:sec:implementation}

We implement and test the proposed model using the following
values for the system parameters: $k_c = 100$, $P_t = 0.99$,
$k_e = 128$, and $L_t=95.0$. The encoder-aligner-decoder
consists of $2$ layers for the encoder and decoder with $128$ LSTM
units per layer. The language model similarly includes a $2$-layer, $128$ LSTM units,
recurrent neural network. The size of
the CAS and English vocabularies is $88$ and
$435$, respectively, based upon the SAIL dataset. All
parameters are chosen based on the performance on 
the validation set.
We train our model using Adam~\cite{kingma2014adam} for
optimization. At test time, we perform approximate inference using a
beam width of two. Our method requires an average of $33$\,s
to generate
instructions for a path of $9$ movements on a
laptop with an Intel Core i7 $2.0$ GHz CPU and $8$\,GB of RAM.

\subsection{Automatic Evaluation}

To the best of our knowledge, we are the first to use the SAIL dataset
for the purposes of generating route instructions. Consequently, we evaluate 
our method by comparing our generated instructions with a
reference set of human-generated commands from the SAIL dataset using
the BLEU score (a $4$-gram matching-based
precision)~\citep{papineni2002bleu}.  For each
command-instruction pair
($c^{(i)},\Lambda^{(i)}$) in
the validation set, we first feed the command $c^{(i)},$ into our model to
obtain the generated instruction $\Lambda^*$, and secondly use
$\Lambda^{(i)}$ and $\Lambda^*$ as the reference and
hypothesis, respectively, to compute the BLEU score. We consider the
average BLEU score at the individual sentence (macro-average
precision) and the full-corpus  (micro-average precision) levels.

\subsection{Human Evaluation} \label{hri2017:sec:study}

The use of BLEU score indicates the similarity between instructions
generated via our method and those produced by humans, but it does not
provide a complete measure of the quality of the instructions (e.g.,
instructions that are correct but different in prose will receive a
low BLEU score). In an effort to further evaluate the accuracy and
usability of our method, we conducted a set of human evaluation experiments 
in which we asked $42$ novice participants on Amazon Mechanical Turk 
($21$ females and $21$ males, ages
$18$--$64$, all native English speakers) to follow natural language
route instructions chosen randomly from two equal-sized sets of
instructions generated by our method and by humans for $50$ distinct
paths of various lengths. The paths and corresponding human-generated
instructions were randomly sampled from the SAIL test set.  Given a
route instruction, human participants were asked to navigate to the
best of their ability using their keyboard within a first-person,
three-dimensional virtual world representative of the three
environments from the SAIL corpus.
After attempting to follow each instruction, each
participant was given a survey comprised of eight questions, three
requesting demographic information and five requesting feedback on
their experience and the quality of the instructions that they
followed. We collected data for a total of $441$ experiments ($227$ 
using human annotated instructions and $214$ using machine-generated 
instructions). 
The system randomly assigned the experiments
to discourage the participants from learning the environments or becoming
familiar with the style of a particular instructor. No participants
experienced the same scenario with both human annotated and 
machine-generated instructions.

\section{Results} \label{hri2017:sec:results}

We evaluate the performance of our architecture by scoring 
the generated instructions using the $4$-gram BLEU score commonly used
as an automatic evaluation mechanism for machine
translation. Comparing to the human-generated instructions, our
method achieves sentence- and corpus-level 
BLEU scores of $74.67\%$ and $60.10\%$, respectively, on the validation set. 
On the test set, the
method achieves sentence- and corpus-level BLEU scores of $72.18\%$
and $45.39\%$, respectively. Figure~\ref{hri2017:fig:generation_example} shows an
example of a route instruction  generated by our system for a given map and path.

\subsection{Aligner Ablation}

\begin{wraptable}{r}{0.5\textwidth}
    \vspace{-0.4cm}
    \centering
    \begin{tabularx}{0.5\textwidth}{r c c}
      \toprule
      & Full Model & No Aligner\\
      \midrule
      sentence-level BLEU & $\mathbf{74.67}$  & $74.40$ \\
      corpus-level BLEU & $\mathbf{60.10}$  & $57.40$ \\
      \bottomrule
    \end{tabularx}
    \caption{Aligner ablation results.}
    \label{hri2017:tab:ablation_test}
\end{wraptable}

Our model employs an aligner in order to learn to focus on particular CAS tokens
that are salient to words in the output. We evaluate the
contribution of the aligner by implementing and training an
alternative model in which the last encoder hidden state is fed
to the decoder. 

Table~\ref{hri2017:tab:ablation_test} compares
the performance of the two models on the original validation set. The
inclusion of an aligner results in a slight increase in the BLEU score of the generated
instructions relative to the human-provided references, and is also
useful as a means of visualizing the inner workings of our model (as
discussed later). 
Additionally, we empirically
find that the aligner improves our model's ability to learn the association
between CAS elements and words in the output, thereby yielding better instructions.

\subsection{Language Model Ablation}

\begin{wraptable}{r}{0.58\textwidth}
    \centering
    \vspace{-0.4cm}
    \begin{tabularx}{0.57\textwidth}{r l}
      \toprule
      LM-score & Candidate \\
      \midrule
      $105.00$ & ``so so a straight chair to your left'' \\
      $27.65$ & ``turn so that the chair is on your left side'' \\
      \midrule
      $101.00$ & ``keep going till the blue flor id on your left'' \\
      $11.00$ & ``move until you see blue floor to your right'' \\
      \bottomrule
    \end{tabularx}
    \caption{Language model ablation outputs.}
    \label{hri2017:tab:lm_ablation_test}
    \vspace{-0.6cm}
\end{wraptable}

Our method employs a language model to rank  instructions generated
for the different candidate CAS commands with different beam width settings.
In practice, the language model, trained on large amounts of English data, 
helps to remove grammatically incorrect sentences produced by the sequence-to-sequence
model, which is only trained on the smaller pairwise dataset. 
Table~\ref{hri2017:tab:lm_ablation_test} presents two instruction
candidates generated by our encoder-aligner-decoder model for two
 CAS commands. Our language model
successfully assigns high perplexity  to the incorrect
instructions, with the chosen instruction being grammatically correct.

\subsection{Aligner Visualization}

\begin{figure}[!t]
    \centering
    \includegraphics[width=0.82\linewidth]{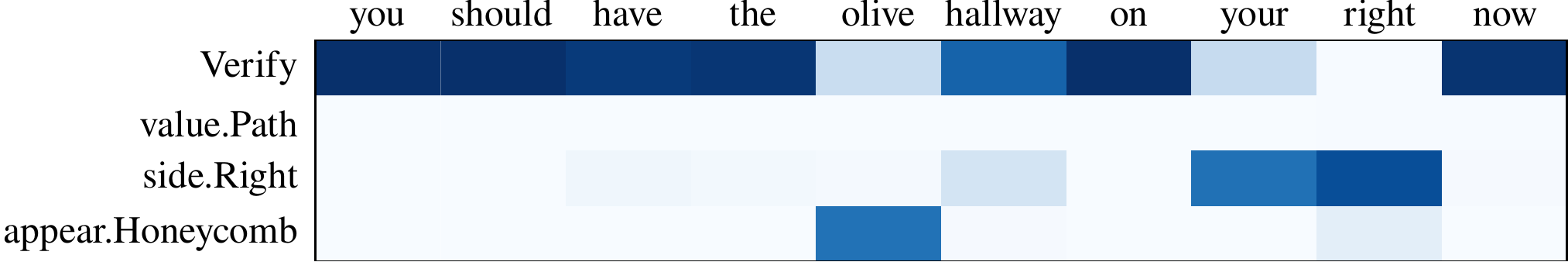}\\[6pt]
    \includegraphics[width=0.82\linewidth]{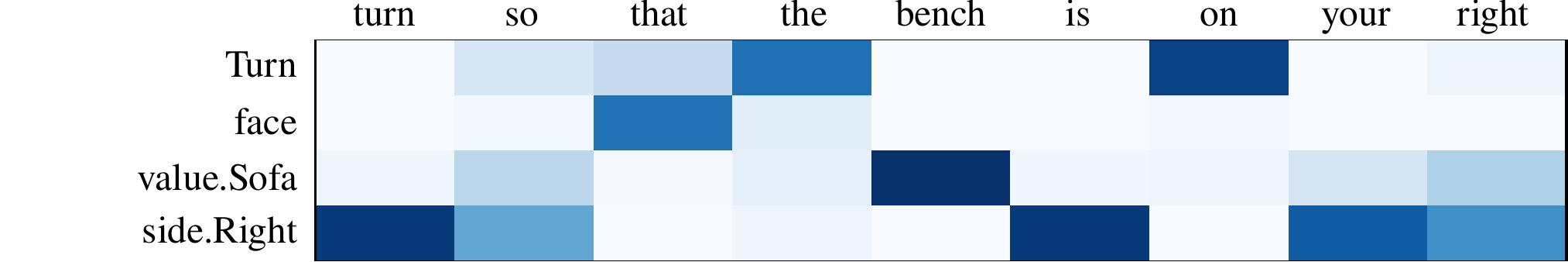}
    \caption{Alignment visualization for two
      pairs of CAS and natural language instructions.
		\label{hri2017:fig:attention_heatmap}}
\end{figure}

Figure~\ref{hri2017:fig:attention_heatmap} presents heat maps that visualize the
alignment between a CAS command that serves as input for surface realization (rows)
and the generated  instruction (columns) for two different
scenarios drawn from the SAIL validation set. The visualizations
demonstrate that our method learns to align elements of the formal CAS
command with their corresponding words in the generated
instruction. For example, the network learns the association between
the honeycomb textured floor and its color (top); that ``bench'' refers to
sofa objects (bottom); and that the phrase ``you should have'' indicates a
verification action (top).

\subsection{Human Evaluation}

\begin{figure}[t]
    \centering
    \includegraphics[width=0.82\linewidth]{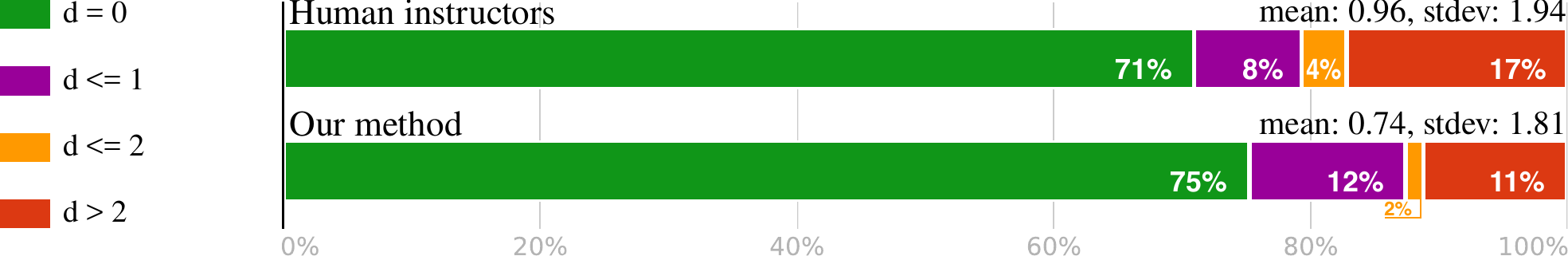}
    \caption{Participants' distances from the goal.
    \label{hri2017:fig:study_performance}}
\end{figure}

We evaluate the accuracy with which human participants followed the
natural language
instructions in terms of the Manhattan
distance $d$ between the desired destination (i.e., the last pose of the 
target path) and the participant's location when s/he finished the scenario. 
Figure~\ref{hri2017:fig:study_performance} compares the accuracy of the
participants' paths when following human-generated instructions with those corresponding to
instructions that our method produced. We report the fraction of times
participants finished within different distances from the
goal.\footnote{We note that the $d=0$ accuracy for the human-generated
  instructions is consistent with that reported
  elsewhere~\cite{chen11}.} The results demonstrate that participants
reached the desired position $4\%$ more often when following
instructions generated using our method compared to the human
instruction baseline. When they didn't reach the destination,
participants reached a location within one vertex away $8\%$ more
often given our instructions. Meanwhile our method yields a failure
rate ($d>2$) that is $6\%$ lower, 
though the difference is not
statistically significant, as confirmed by a p-value of $0.131$
relative to a standard significance level of $0.05$.
Of scenarios in which
participants reached the destination, the total time required to
interpret and follow our method's instructions is
$9.52$\,s less than those generated by humans.

\begin{figure}[!t]
    \centering
    
    \begin{subfigure}[b]{\textwidth}
    \centering
    \includegraphics[width=0.85\textwidth,clip=false]{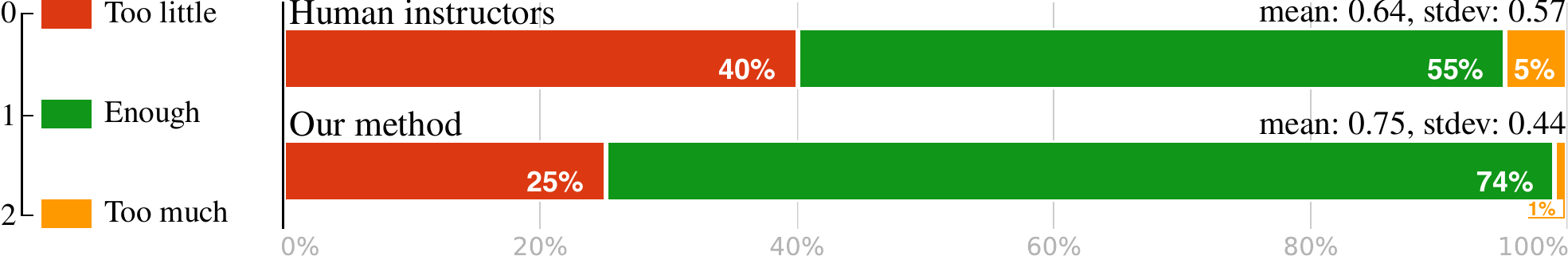}
    \caption{{\small ``Define the amount of information provided''~(p-value:~$0.041$)}}
    \label{hri2017:fig:survey_information}
    \end{subfigure}
    \vspace{5pt}
    
    \begin{subfigure}[b]{\textwidth}
    \centering
    \includegraphics[width=0.85\textwidth,clip=false]{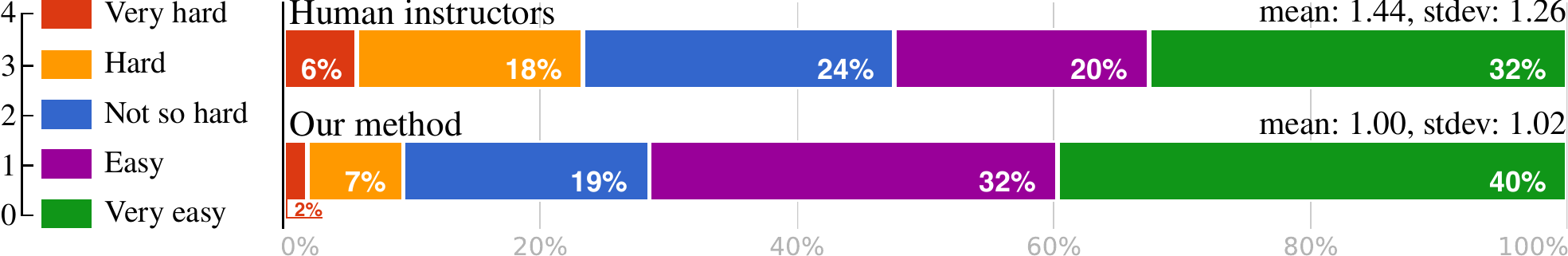}
    \caption{{\small ``Evaluate the task in terms of difficulty''~(p-value:~$0.007$)}}
    \label{hri2017:fig:survey_difficulty}
    \end{subfigure}
    \vspace{5pt}
    
    \begin{subfigure}[b]{\textwidth}
    \centering
    \includegraphics[width=0.85\textwidth,clip=false]{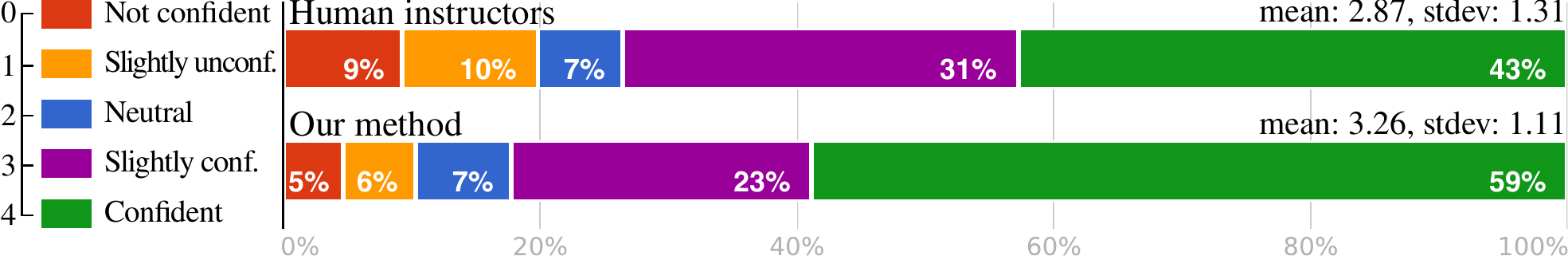}
    \caption{{\small ``Your confidence about the path just followed''~(p-value:~$0.002$)}}
    \label{hri2017:fig:survey_confidence}
    \end{subfigure}
    \vspace{5pt}
    
    \begin{subfigure}[b]{\textwidth}
    \centering
    \includegraphics[width=0.85\textwidth,clip=false]{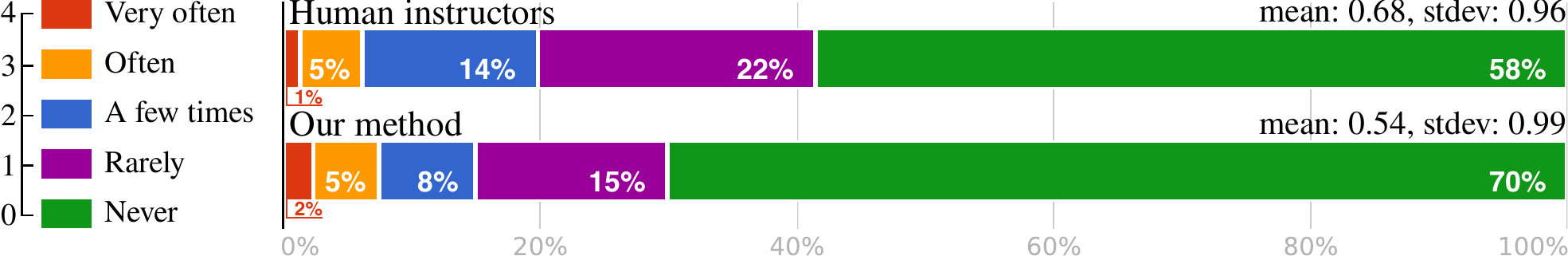}
    \caption{{\small ``{How many times did you have to backtrack?}''~(p-value:~$0.155$)}}
    \label{hri2017:fig:survey_backtrack}
    \end{subfigure}
    \vspace{5pt}
    
    \begin{subfigure}[b]{\textwidth}
    \centering
    \includegraphics[width=0.85\textwidth,clip=false]{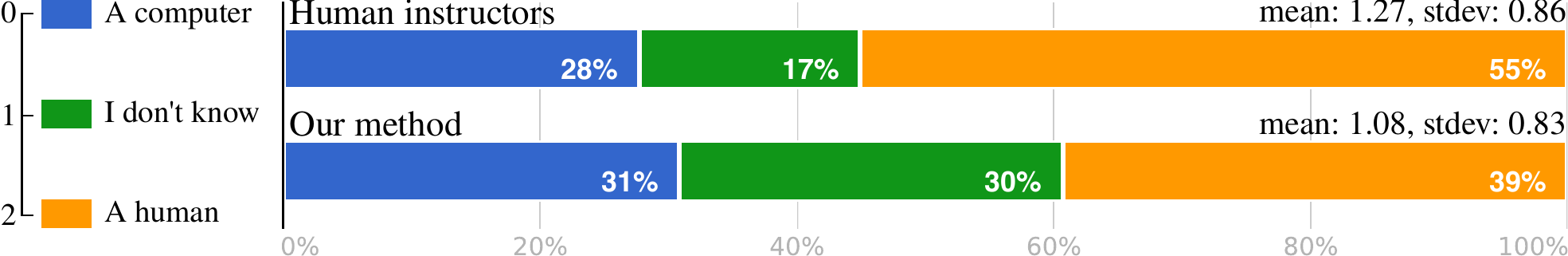}
    \caption{{\small ``{Who do you think generated the instructions?}''~(p-value:~$0.036$)}}
    \label{hri2017:fig:survey_who}
    \end{subfigure}
    
    \caption{Participants' survey response statistics.}
    \label{hri2017:fig:survey_results}
\end{figure}

Figure~\ref{hri2017:fig:survey_results} presents the participants' responses to
the survey questions that query their experience following the
instructions. By using IRL to learn a content selection policy for
constructing CAS structures, our method generates instructions that
convey enough information to follow the command and were rated as
providing too little information $15\%$ less frequently than the
human-generated baseline
(Figure~\ref{hri2017:fig:survey_information}). Meanwhile, participants felt that
our instructions were easier to follow
(Figure~\ref{hri2017:fig:survey_difficulty}) than the human-generated baselines
($72\%$ vs.\ $52\%$ rated as ``easy'' or ``very easy'' for our method
vs.\ the baseline).  Participants were more confident in their ability
to follow our method's instructions (Figure~\ref{hri2017:fig:survey_confidence})
and felt that they had to backtrack less often
(Figure~\ref{hri2017:fig:survey_backtrack}). Meanwhile, both types of
instructions were confused equally often as being machine-generated
(Figure~\ref{hri2017:fig:survey_who}), however participants were less sure of
who generated our instructions.

\begin{figure}[h]
  \vspace{-.3cm}
  \centering
  \vspace{12pt}
  \begin{tabular}{ M{10cm} N}
    \toprule
    \textbf{Map and Path} & \\[3pt]
    \centering   %
    \vspace{-.45cm}
	\includegraphics[width=\linewidth]{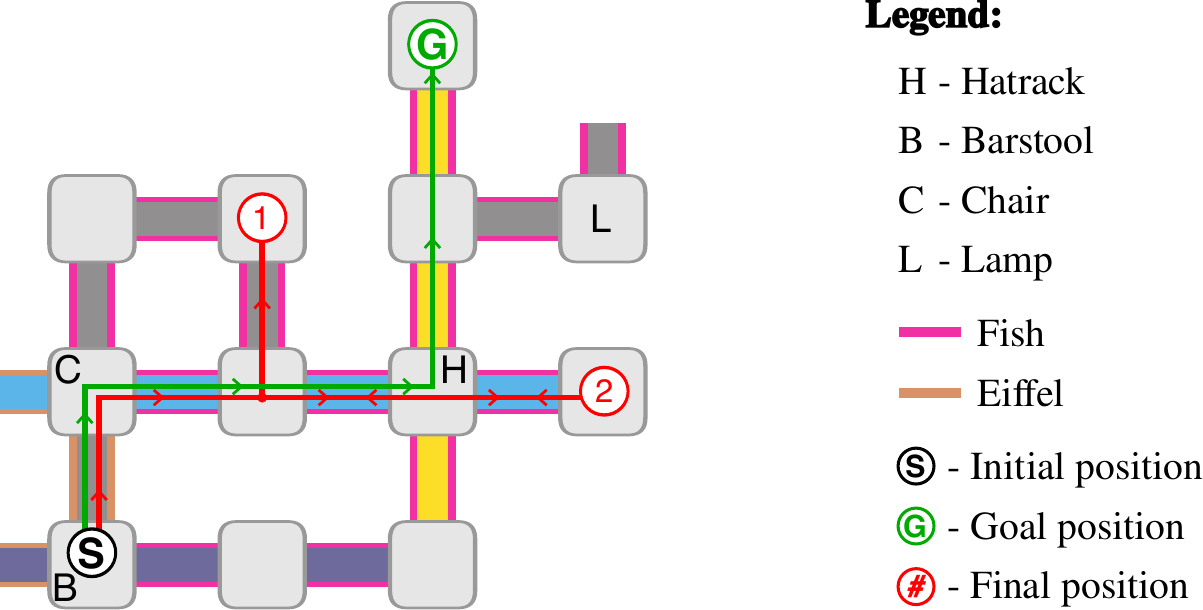} & \\
    \midrule
    \textbf{Instructions} & \\[3pt]
    \textbf{Human}: ``head toward the blue floored hallway. make a right on 
    it. go down till you see the fish walled areas. make a
	left in the fish walled hallway and go to the very end''
  	& \\
  	\midrule
   	\textbf{Our's}: ``turn to face the white hallway. walk forward once. 
   	turn right. walk forward twice. turn left. move to the wall''
   	& \\
    \bottomrule
  	\end{tabular}
	\caption{
	Paths followed by five participants 
	given human-generated (red) and our (green)  directions.
	}
	\label{hri2017:fig:generation_examples}
\end{figure}

Figure~\ref{hri2017:fig:generation_examples} compares the paths that
participants took when following our instructions 
with those that they
took given the reference human-generated directions for one scenario.
In particular, five 
participants failed to reach the destination when provided with the
human-generated instruction. Two of the participants went  directly to
location $1$, two participants navigated to location $2$, and one participant went to
location $2$ before backtracking and taking a right to location
$1$. We attribute the failures to the ambiguity in the human-generated
instruction that references ``fish walled areas,'' which could
correspond to most of the hallways in this portion of the map (as
denoted by the pink-colored lines). On the other hand, each of the
five participants followed the intended path (shown in green) and
reached the goal when following the instruction generated using our
method.

\section{Conclusion}
\label{hri2017:sec:conclusion}

We presented a model for natural language generation in the context of
providing indoor route instructions that exploits a structured
approach to produce unambiguous, easy to follow and grammatically
correct human-like route instructions. Currently, our model generates
natural language route instructions for the shortest path to the
goal.  However, there are situations in which a longer path may
afford instructions that are more
straightforward~\cite{richter2008simplest} or that increase the
likelihood of reaching the
destination~\cite{haque2006algorithms}. Another interesting direction
for future work would be to integrate a model of
instruction followers~\cite{mei2016listen} with our
architecture in an effort to learn to generate instructions that are
easier to follow. Such an approach would permit training the model in
a reinforcement learning setting, directly optimizing over task performance.

    \newpage
    \chapter{\nlusec}
    \label{isrr2017:sec:main}
    This work was published in~\cite{daniele17a}.\\
    \begin{wrapfigure}{r}{0.5\textwidth}
     \centering
     \includegraphics[width=0.45\textwidth]{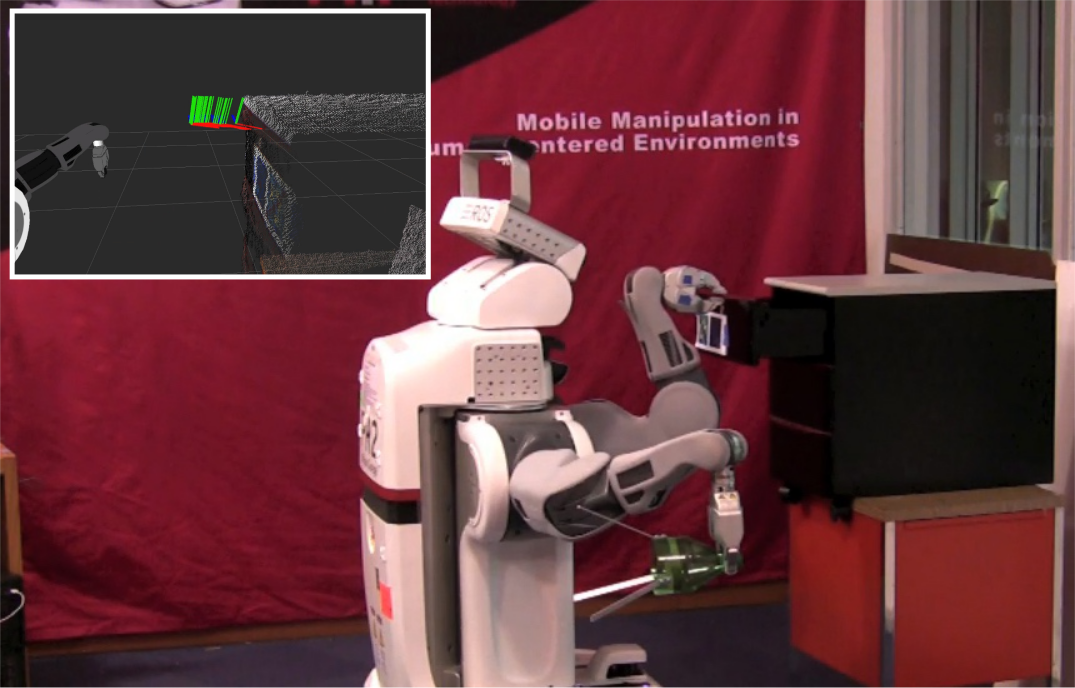}
     \caption{A robot autonomously opens a drawer by recalling a learned kinematic model of the object, which is visualized in the inset.}
     \label{isrr2017:fig:pr2}
\end{wrapfigure}

As robots move off factory floors and into our homes and workplaces,
they face the challenge of interacting with the articulated objects
frequently found in environments built by and for humans (e.g., drawers, ovens, refrigerators, and faucets). Typically, this interaction is
predefined in the form of a manipulation policy that must be
(manually) specified for each object that the robot is expected to
interact with. Such an approach may be reasonable for robots that interact with a small number of objects, but human environments contain a large number of diverse objects. In an effort to improve efficiency and
generalizability, recent work employs visual demonstrations to learn
representations that describe the motion of these parts in the form of
kinematic models that express the rotational, prismatic, and rigid
relationships between object parts~\citep{sturm11, huang12, katz13, pillai14, byravan17}. These structured object-relative models, which constrain the object's motion manifold, are suitable for trajectory controllers~\cite{jain10, sturm11}, provide a common representation amenable to transfer between objects~\cite{sung15}, and allow for manipulation policies that are more efficient and
deliberate than reactive policies (Figure~\ref{isrr2017:fig:pr2}). However, such visual cues may be too time-consuming to provide or may not be readily available, such as when a user is remotely commanding a robot over a bandwidth-limited channel (e.g., for disaster relief). Further, reliance solely on vision makes these methods sensitive to common errors in data association, object
segmentation, and tracking (e.g., tracking features over time and associating them with the correct object part) that occur as a result of clutter,
occlusions, and a dearth of visual features. Consequently, most existing systems require scenes to be free of distractors and that object parts be labeled with fiducial markers.

\begin{figure}[!t]
    \centering
    \includegraphics[width=0.9\textwidth]{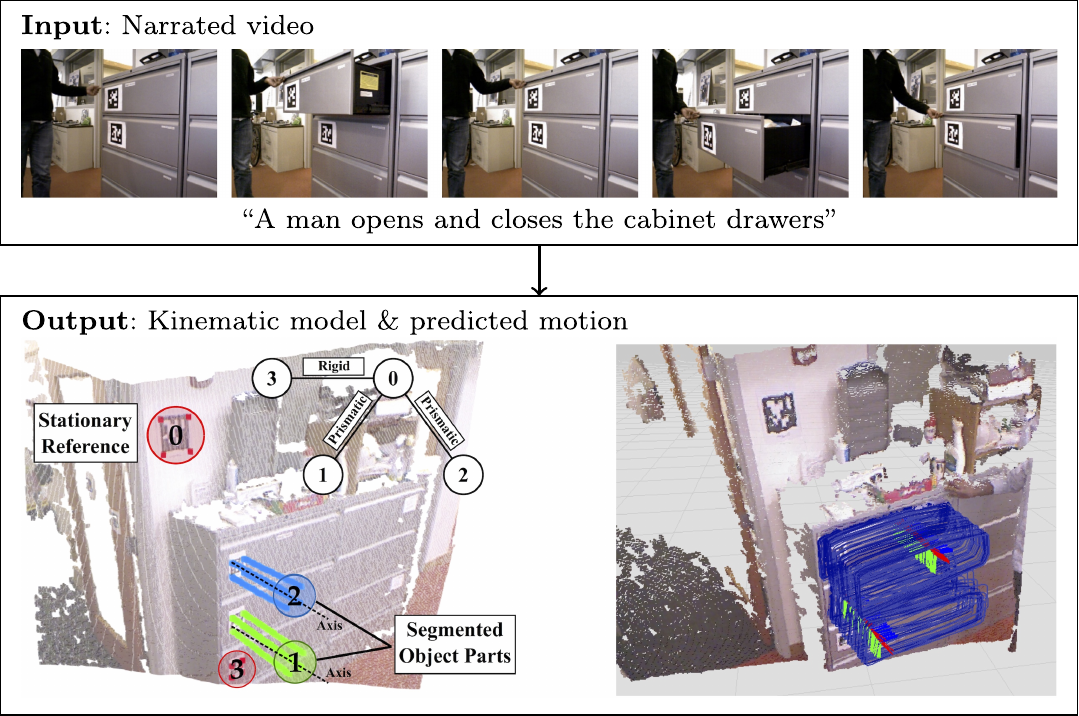}
    \caption{Our framework learns the kinematic model that governs the
      motion of articulated objects (lower-left) from narrated RGB-D
      videos. The method can then use this learned model to
      subsequently predict the motion of an object's parts (lower-right).} \label{isrr2017:fig:teaser}
\end{figure}

Natural language utterances offer a flexible, bandwidth-efficient medium that humans can readily use to convey knowledge of an object's operation~\cite{sung15}. When paired with visual observations, free-form descriptions of an articulated motion also provide a source of
information that is complementary to visual input. In similar fashion to existing work on learning manipulation policies from Internet videos~\cite{yang15}, the ability to learn kinematic models from narrated demonstrations, either given in situ or via instructional videos~\cite{yu14, malmaud15, sener15}, provides an efficient, intuitive means for people to convey information to robots, as the authors have shown in the context of map learning~\cite{walter13}. Natural language descriptions that accompany these demonstrations can be used to overcome some of the limitations of using
visual-only observations, e.g., by providing cues regarding the number
of parts that comprise the object or the motion type (e.g.,
rotational) between a pair of parts. However, fusing visual and linguistic observations is challenging. For one, language and vision provide disparate observations of motion and exhibit different statistical properties. Secondly, the two are often prone to uncertainty. RGB-D observations are subject to occlusions (e.g., as the human interacts with the object) and the objects often lack texture (e.g., the drawers in Figure~\ref{isrr2017:fig:teaser}), which makes feature detection challenging and feature correspondences subject to noise. Meanwhile, free-form descriptions exhibit variability and are prone to errors (e.g., confusing ``left'' and ``right''). Further, language also tends to be ambiguous with respect to the corresponding referents (i.e., object parts and their motion) in the scene. For example ``open'' can suggest both rotational and prismatic motion, and inferring the correct grounding requires reasoning over the full description (e.g., ``open the door'' vs.\ ``open the  drawers'').

In order to overcome these challenges, we present a multimodal learning framework that estimates the kinematic structure
and parameters of complex multi-part objects using both vision and
language input. We address the challenges associated with language understanding through a probabilistic language model that captures the compositional and hierarchical structure of natural language descriptions. Additionally, our method maintains a distribution over a sparse, structured model of an object's kinematics, which provides a common representation with which to fuse disparate linguistic and visual observations. %

Our effort is inspired by the recent attention that has been paid to
the joint use of vision and language as complementary signals for
multiview learning in robotics~\cite{kollar13, zender08, pronobis12, walter13, hemachandra14, sung15} and scene understanding~\citep{krishnamurthy13,
  ramanathan2014linking, guadarrama14, karpathy-15,
  vinyals15, donahue-14,
  srivastava15, alayrac17}. We
leverage the joint advantages of these two modalities in order to
estimate the structure and parameters that define kinematic models of
complex, multi-part objects such as doors, desks, chairs, and
appliances from narrated examples such as those conveyed in instructional videos or through demonstrations~\cite{argall09} in the form of a ``guided tour of manipulation'' (Figure~\ref{isrr2017:fig:teaser}), which provides an efficient and flexible means for humans to share information with robots.

Our multimodal learning framework first extracts noisy observations of
the object parts and their motion separately from the vision- and
language-based observations. It then fuses these observations to learn a
probabilistic model over the kinematic structure and model parameters
that best explain the motion observed in the vision and language
streams. Integral to this process is an appropriate means of
representing the ambiguous nature of observations gleaned from natural
language descriptions. We treat language understanding as a symbol grounding problem and employ a probabilistic language model~\cite{howard-14} that captures the uncertainty in the mapping between words in the description and their corresponding referents in the scene, namely the object parts and their relative motion. We fuse these language-based observations with those extracted from vision to estimate a joint distribution over the structure and parameters that define the kinematics of each object.

The contributions of this work include a multimodal approach to learning kinematic
models from vision and language signals and the integration of a probabilistic language model that grounds natural language descriptions into a structured representation of an object's articulation manifold. By jointly reasoning over vision and language cues, our framework is able to formulate a complete object model without the need for an expressed environment model. Our method requires no prior knowledge about the objects and operates in situ, without the need for environment preparation (i.e., fiducials). Evaluations on a dataset of video-text
pairs demonstrate improvements over the previous
state-of-the-art, which only uses visual information.

\section{Related Work} \label{isrr2017:sec:related}

Recent work considers the problem of learning articulated models based
upon visual observations of demonstrated motion. Several methods
formulate this problem as bundle adjustment, using
structure-from-motion methods to first segment an articulated object
into its compositional parts and to then estimate the parameters of
the rotational and prismatic degrees-of-freedom that describe inter-part
motion~\citep{yan06,huang12}. These methods are prone to erroneous
estimates of the pose of the object's parts and of the inter-part
models as a result of outliers in visual feature
matching. Alternatively, \citet{katz10} propose an active learning framework
that allows a robot to interact with articulated objects to induce
motion. This method operates in a deterministic manner, first assuming
that each part-to-part motion is prismatic. Only when the residual
error exceeds a threshold does it consider the alternative rotational
model. Further, they estimate the models based upon interactive
observations acquired in a structured environment free of clutter,
with the object occupying a significant portion of the RGB-D sensor's
field-of-view. \citet{katz13} improve upon the complexity of this method
while preserving the accuracy of the inferred models. This method is
prone to over-fitting to the observed motion and may result in overly
complex models to match the observations. \citet{hausman15} similarly
enable a robot to interact with the object and describe a
probabilistic model that integrates observations of fiducials with
manipulator feedback. Meanwhile, \citet{sturm11} propose a
probabilistic approach that simultaneously reasons over the likelihood
of observations while accounting for the learned model
complexity. Their method requires that the number of parts that
compose the object be known in advance and that fiducials be placed on
each part to enable the visual observation of motion. More recently,
\citet{pillai14} propose an extension to this work that uses novel
vision-based motion segmentation and tracking that enables model
learning in situ, without prior knowledge of the number of parts or the
need for fiducial markers. Our approach builds upon this method
with the addition of natural language descriptions of motion as an
additional observation mode in a multimodal learning framework. Meanwhile, \citet{schmidt14} propose a framework that addresses the similar problem of tracking articulated models that uses an articulated variation of the signed distance function to identify the model that best fits observed depth data.
Related, \citet{martin14} consider the problem of perceiving articulated objects and demonstrate how an estimate of kinematic structure facilitates tracking and manipulation. The authors improve upon this work by integrating shape reconstruction, pose estimation, and kinematic estimation~\cite{martin16}.

Related, there has been renewed attention to enabling robots to interpret natural language instructions that command navigation~\citep{kollar-10, matuszek-10, chen-11, artzi-13, mei16}  and manipulation~\citep{tellex-11, howard-14, misra16, paul16a} through symbol grounding and semantic parsing methods. While most existing grounded language acquisition methods abstract away perception by assuming a known symbolic world model, other work jointly reasons over language and sensing~\cite{matuszek-12, guadarrama13, duvallet14, hemachandra-15} for instruction following. Meanwhile, multimodal learning methods have been proposed that use language and vision to formulate spatial-semantic maps of a robot's environment~\citep{zender08, pronobis12, walter13, hemachandra14} and to learn object manipulation policies~\cite{sung15}. Particularly relevant to our work, \citet{sung15} learn a neural embedding of text, vision, and motion trajectories to transfer manipulation plans between similarly operating objects. \citet{kollar13} extend their framework that jointly learns a semantic parsing of language and vision~\cite{krishnamurthy13} to enable robots to learn object and spatial relation classifiers from textual descriptions paired with images. We similarly use language and vision in a joint learning framework, but for the challenging task of learning object articulation in terms of kinematic motion
models. Beyond robotics, there is a long history of work that exploits the complementary nature of vision and language in the context of multiview learning, dating back to the seminal SHRDLU program~\cite{winograd72}. This includes work for such tasks as image and video caption synthesis~\citep{ordonez11, karpathy-15, vinyals15, xu-15, donahue-14, srivastava15}, large-vocabulary object retrieval~\cite{guadarrama14}, visual coreference resolution~\citep{kong2014what, ramanathan2014linking}, and visual question-answering~\cite{antol-15}. Particularly related to our work are methods that use instructional videos paired with language (text or speech) for weakly supervised learning~\cite{yu14, malmaud15}, extracting procedural knowledge~\cite{sener15}, and identifying manipulating actions~\cite{sung15, alayrac17}.

\section{Multimodal Learning Framework} \label{isrr2017:sec:approach}

\begin{figure}[!t]
    \centering
    \includegraphics[width=\textwidth]{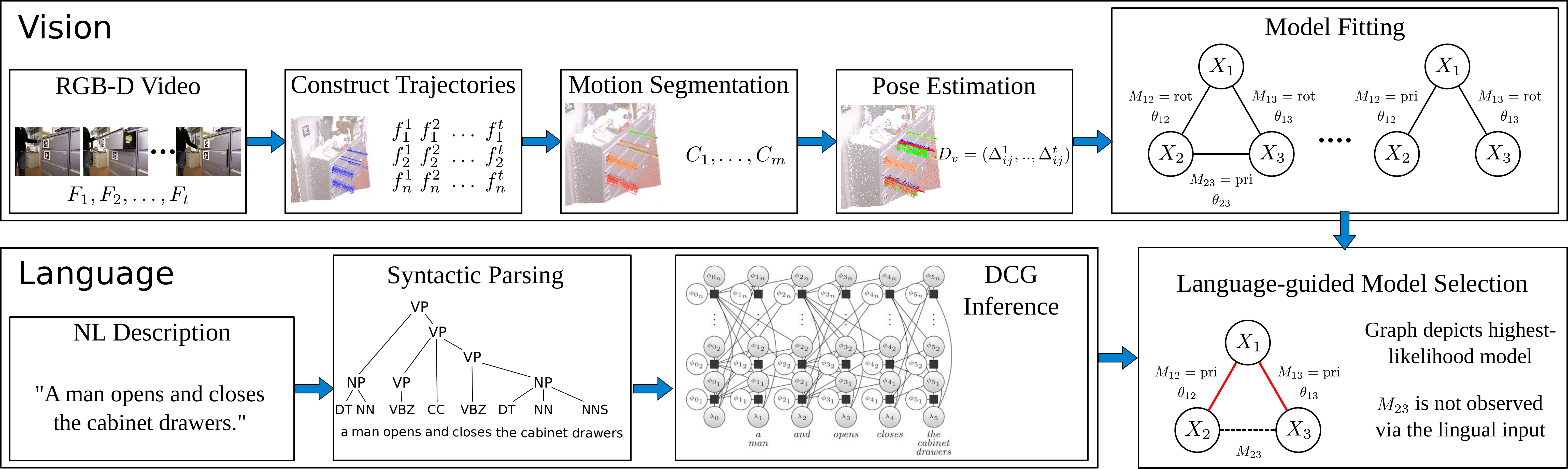}
    \caption{Our multimodal articulation learning framework first
    identifies clusters of visual features that correspond to
    individual object parts. It then uses these feature trajectories
    to estimate the model
    parameters, assuming an initial estimate of the kinematic type
    associated with each edge in
    the graph. The method grounds natural language descriptions of the motion to their corresponding referents in the kinematic model and parameters through a probabilistic language model, visualized as a factor graph. The vision and language observations are then fused to learn a distribution over the object's kinematic model.}
    \label{isrr2017:fig:architecture}
\end{figure}
Given an RGB-D video paired with the corresponding natural language
description (alternatively, an instruction or caption) of an
articulated object's motion, our goal is to infer the structure and
parameters of the object's kinematic model. Adopting the formulation
proposed by~\citet{sturm11}, we represent this model as a
graph, where each vertex denotes a different part of the object (or the
stationary background) and edges denote the existence of constrained
motion (e.g., a linkage) between two parts (Figure~\ref{isrr2017:fig:teaser}). More formally, we estimate
a \emph{kinematic graph} $G = (V_G, E_G)$ that consists of vertices $V_G$ for each object part and edges
$E_G \subset V_G \times V_G$ between parts whose relative motion is
kinematically constrained. Associated with each edge $(ij) \in E_G$ is its kinematic type
$M_{ij} \in \{\textrm{rotational},
\textrm{prismatic},\textrm{rigid}\}$
as well as the corresponding parameters $\theta_{ij}$, such as the
axis of rotation and the range of motion (see
Figure~\ref{isrr2017:fig:architecture}, lower-right). We take as
input vision $D_v$ and language $D_l$ observations of the type and
parameters of the edges in the graph. Our method then uses this
vision-language observation pair $D_z = \{D_v,D_l\}$ to infer the maximum a posteriori kinematic structure and model
parameters that constitute the kinematic graph:
\begin{subequations}
    \begin{align}
      \hat{G} &= \argmax{G} p(G \vert D_z) \\
              &= \argmax{G} p(\{M_{ij}, \theta_{ij} \vert (ij) \in E_G\} \vert D_z)\\
              &= \argmax{G} \prod_{(ij)\in E_G}p(M_{ij}, \theta_{ij}
                \vert D_z)
    \end{align}
\end{subequations}

Due to the complexity of joint inference, we adopt the procedure
described by \citet{sturm11} and use a two-step inference procedure
that alternates between model parameter fitting and model structure
selection steps (Figure~\ref{isrr2017:fig:architecture}). In the first step, we assume
a particular kinematic model type between each object $i$ and $j$
(e.g., prismatic), and then estimate the kinematic parameters based on
the vision data (relative transformation between the two objects) and
the assumed model type $M_{ij}$. We make one such assumption for each
possible model type for each object pair.

In the model selection step, we then use the natural language
description to infer the kinematic graph structure that best expresses
the observation. While our previous work~\citep{pillai14} provides
visual observations of motion without the need for fiducials, it
relies upon feature tracking and segmentation that can fail when the
object parts lack texture (e.g., metal door handles) or when the scene
is cluttered. Our system incorporates language as an additional, complementary observation of the motion, in order to improve
the robustness and accuracy of model selection.

\subsection{Vision-guided Model Fitting} \label{isrr2017:sec:vision-model}

We parse a given RGB-D video of the objects' motion (either performed by a human or the robot via teleoperation) to arrive at a visual observation of the trajectory of each object part~\cite{pillai14}.

The method (Figure~\ref{isrr2017:fig:architecture}, ``Construct Trajectories'') first identifies a set of 3D feature
trajectories $\{(f_1^1, f_1^2, \ldots, f_1^t), \ldots(f_n^1, f_n^2, \ldots, f_n^t)\}$ that correspond to different elements in the scene,
including the object parts, background, and clutter. Importantly, both the number of elements and the assignment of points in the RGB-D video to these elements are assumed to be unknown a priori. Further, many of the objects that we encounter lack the amount of texture typically required of SIFT~\cite{lowe04} and KLT~\cite{bouguet01} features. Consequently, we utilize dense trajectories~\cite{wang11} through a strategy that involves dense sampling (via the Shi-Tomasi criterion) for feature extraction followed by dense optical flow for propagation. We prune trajectories after a fixed length and subsequently sample new features to reduce feature drift.

Having extracted a set of features trajectories, the next step is then to group features that correspond to the same scene element via motion segmentation (Figure~\ref{isrr2017:fig:architecture}, ``Motion Segmentation''). For this purpose, we evaluate the relative displacement between pairs of feature trajectories along with the angle between their normals. We model the relative displacement and angle as Gaussian in order to account for measurement noise. We then employ density-based clustering~\cite{ester96} to identify rigidly associated feature trajectories. These clusters $\{C_1, C_2, \ldots, C_m\}$ denote the parsing of the scene into its requisite elements, namely the inferred object parts and background.

Next, we estimate the 6-DOF pose $x_i^t$ of each cluster at each point in time according to the set of features $Z_i^t$ assigned to each cluster $C_i$ at time $t$ (Figure~\ref{isrr2017:fig:architecture}, ``Pose Estimation''). We treat this as a pose graph estimation problem, whereby we consider the relative transformation $\Delta_i^{t-1,t}$ between successive time steps for each cluster based on the known correspondences between features $Z_i^{t-1}$ and $Z_i^t$.\footnote{We employ RANSAC~\cite{fischler81} to improve robustness to erroneous corresponences.} We optimize the pose graph using iSAM~\cite{kaess08}, which models the relative transformations as observations (constraints) in a factor graph with nodes that denote cluster poses.

The resulting 6-DOF pose trajectories constitute the visual observation of the
motion $D_v$. Our framework uses these trajectories to estimate the
parameters of a candidate kinematic model during the model fitting
step. Specifically, we find the kinematic parameters that best
explain the visual data given the assumed model
\begin{equation} \label{isrr2017:eqn:model-fitting}
    \hat{\theta}_{ij} = \argmax{\theta_{ij}} p(D_v \vert \hat{M}_{ij}, \theta_{ij}),
\end{equation}
where
$D_v = (\Delta_{ij}^1, ..., \Delta_{ij}^t), \forall (ij) \in E_G $ is
the sequence of observed relative transformations between the poses of
two object parts $i$ and $j$, and $\hat{M}_{ij}$ is the current estimate of their
model type. We perform this optimization over the joint kinematic
structure defined by the edges in the graph~\citep{sturm11}.

\subsection{Language-guided Model Selection}

Methods that solely rely on visual input are sensitive to the effects of scene clutter and the lack of texture, which can result in erroneous estimates for the structure and parameters of the kinematic model~\citep{pillai14}.  An alternative is to exploit audial information, provided in the form of utterances provided by the operator, to help guide the process for inferring the relationships between objects in the environment.   Specifically, we consider a natural language description $D_l$ that
describes the motion observed in the video.  Given this description, we infer the maximum a posteriori set of affordances or relationships between pairwise objects in the utterance.  Note that we do not assume that valid
captions provide an unambiguous description of all affordances, but rather consider a distribution over the language observation, which provides robustness to noisy, incomplete, or incorrect descriptions.

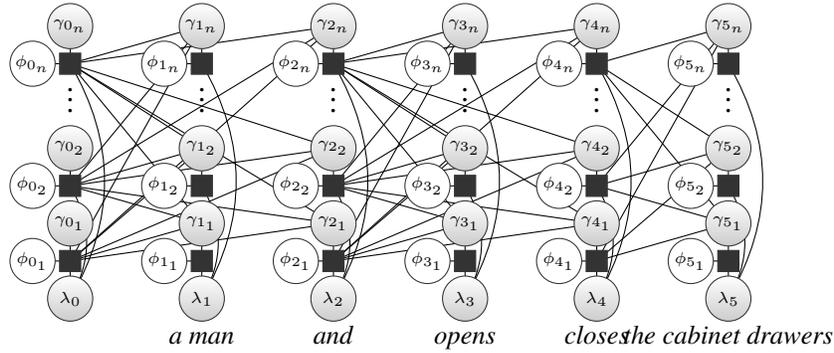
\begin{figure}[!t]
\begin{center}
\begin{tikzpicture}[textnode/.style={anchor=mid,font=\tiny},nodeknown/.style={circle,draw=black!80,fill=black!10,minimum size=6mm,font=\tiny,top color=white,bottom color=black!20},nodeunknown/.style={circle,draw=black!80,fill=white,minimum size=6mm,font=\tiny},factor/.style={rectangle,draw=black!80,fill=black!80,minimum size=2mm,font=\tiny,text=white}]
\draw[-] (0,0.5) to (0,1);
\draw[-] (0,0.5) to [bend right=30] (0,2);
\draw[-] (0,0.5) to [bend right=30] (0,3.625);
\draw[-] (-0.5,1) to (0,1);
\draw[-] (-0.5,2) to (0,2);
\draw[-] (-0.5,3.625) to (0,3.625);
\draw[-] (0,1.5) to (0,1);
\draw[-] (0,2.5) to (0,2);
\draw[-] (0,4.125) to (0,3.625);
\draw[-] (1.75,1.5) to (0,1);
\draw[-] (1.75,2.5) to (0,1);
\draw[-] (1.75,4.125) to (0,1);
\draw[-] (1.75,1.5) to (0,2);
\draw[-] (1.75,2.5) to (0,2);
\draw[-] (1.75,4.125) to (0,2);
\draw[-] (1.75,1.5) to (0,3.625);
\draw[-] (1.75,2.5) to (0,3.625);
\draw[-] (1.75,4.125) to (0,3.625);
\draw[-] (3.5,1.5) to (0,1);
\draw[-] (3.5,2.5) to (0,1);
\draw[-] (3.5,4.125) to (0,1);
\draw[-] (3.5,1.5) to (0,2);
\draw[-] (3.5,2.5) to (0,2);
\draw[-] (3.5,4.125) to (0,2);
\draw[-] (3.5,1.5) to (0,3.625);
\draw[-] (3.5,2.5) to (0,3.625);
\draw[-] (3.5,4.125) to (0,3.625);
\draw[-] (1.75,0.5) to (1.75,1);
\draw[-] (1.75,0.5) to [bend right=30] (1.75,2);
\draw[-] (1.75,0.5) to [bend right=30] (1.75,3.625);
\draw[-] (1.25,1) to (1.75,1);
\draw[-] (1.25,2) to (1.75,2);
\draw[-] (1.25,3.625) to (1.75,3.625);
\draw[-] (1.75,1.5) to (1.75,1);
\draw[-] (1.75,2.5) to (1.75,2);
\draw[-] (1.75,4.125) to (1.75,3.625);
\draw[-] (3.5,0.5) to (3.5,1);
\draw[-] (3.5,0.5) to [bend right=30] (3.5,2);
\draw[-] (3.5,0.5) to [bend right=30] (3.5,3.625);
\draw[-] (3,1) to (3.5,1);
\draw[-] (3,2) to (3.5,2);
\draw[-] (3,3.625) to (3.5,3.625);
\draw[-] (3.5,1.5) to (3.5,1);
\draw[-] (3.5,2.5) to (3.5,2);
\draw[-] (3.5,4.125) to (3.5,3.625);
\draw[-] (5.25,1.5) to (3.5,1);
\draw[-] (5.25,2.5) to (3.5,1);
\draw[-] (5.25,4.125) to (3.5,1);
\draw[-] (5.25,1.5) to (3.5,2);
\draw[-] (5.25,2.5) to (3.5,2);
\draw[-] (5.25,4.125) to (3.5,2);
\draw[-] (5.25,1.5) to (3.5,3.625);
\draw[-] (5.25,2.5) to (3.5,3.625);
\draw[-] (5.25,4.125) to (3.5,3.625);
\draw[-] (7,1.5) to (3.5,1);
\draw[-] (7,2.5) to (3.5,1);
\draw[-] (7,4.125) to (3.5,1);
\draw[-] (7,1.5) to (3.5,2);
\draw[-] (7,2.5) to (3.5,2);
\draw[-] (7,4.125) to (3.5,2);
\draw[-] (7,1.5) to (3.5,3.625);
\draw[-] (7,2.5) to (3.5,3.625);
\draw[-] (7,4.125) to (3.5,3.625);
\draw[-] (5.25,0.5) to (5.25,1);
\draw[-] (5.25,0.5) to [bend right=30] (5.25,2);
\draw[-] (5.25,0.5) to [bend right=30] (5.25,3.625);
\draw[-] (4.75,1) to (5.25,1);
\draw[-] (4.75,2) to (5.25,2);
\draw[-] (4.75,3.625) to (5.25,3.625);
\draw[-] (5.25,1.5) to (5.25,1);
\draw[-] (5.25,2.5) to (5.25,2);
\draw[-] (5.25,4.125) to (5.25,3.625);
\draw[-] (7,0.5) to (7,1);
\draw[-] (7,0.5) to [bend right=30] (7,2);
\draw[-] (7,0.5) to [bend right=30] (7,3.625);
\draw[-] (6.5,1) to (7,1);
\draw[-] (6.5,2) to (7,2);
\draw[-] (6.5,3.625) to (7,3.625);
\draw[-] (7,1.5) to (7,1);
\draw[-] (7,2.5) to (7,2);
\draw[-] (7,4.125) to (7,3.625);
\draw[-] (8.75,1.5) to (7,1);
\draw[-] (8.75,2.5) to (7,1);
\draw[-] (8.75,4.125) to (7,1);
\draw[-] (8.75,1.5) to (7,2);
\draw[-] (8.75,2.5) to (7,2);
\draw[-] (8.75,4.125) to (7,2);
\draw[-] (8.75,1.5) to (7,3.625);
\draw[-] (8.75,2.5) to (7,3.625);
\draw[-] (8.75,4.125) to (7,3.625);
\draw[-] (8.75,0.5) to (8.75,1);
\draw[-] (8.75,0.5) to [bend right=30] (8.75,2);
\draw[-] (8.75,0.5) to [bend right=30] (8.75,3.625);
\draw[-] (8.25,1) to (8.75,1);
\draw[-] (8.25,2) to (8.75,2);
\draw[-] (8.25,3.625) to (8.75,3.625);
\draw[-] (8.75,1.5) to (8.75,1);
\draw[-] (8.75,2.5) to (8.75,2);
\draw[-] (8.75,4.125) to (8.75,3.625);
\node[nodeknown] (p0) at (0,0.5) {};
\node[font=\tiny] (p0label) at (0,0.5) {$\lambda_{0}$};
\node[nodeunknown] (c01) at (-0.5,1) {};
\node[font=\tiny] (c01label) at (-0.5,1) {$\phi_{0_{1}}$};
\node[nodeunknown] (c02) at (-0.5,2) {};
\node[font=\tiny] (c02label) at (-0.5,2) {$\phi_{0_{2}}$};
\node[nodeunknown] (c0n) at (-0.5,3.625) {};
\node[font=\tiny] (c0nlabel) at (-0.5,3.625) {$\phi_{0_{n}}$};
\node[nodeknown] (g01) at (0,1.5) {};
\node[font=\tiny] (g01label) at (0,1.5) {$\gamma_{0_{1}}$};
\node[nodeknown] (g02) at (0,2.5) {};
\node[font=\tiny] (g02label) at (0,2.5) {$\gamma_{0_{2}}$};
\node[] (g0dots) at (0,3.25) {$\vdots$};
\node[nodeknown] (g0n) at (0,4.125) {};
\node[font=\tiny] (g0nlabel) at (0,4.125) {$\gamma_{0_{n}}$};
\node[factor] (f01) at (0,1) {};
\node[factor] (f02) at (0,2) {};
\node[factor] (f0n) at (0,3.625) {};
\node[textnode] (l1) at (1.75,-0) {\footnotesize{\textit{a man}}};
\node[nodeknown] (p1) at (1.75,0.5) {};
\node[font=\tiny] (p1label) at (1.75,0.5) {$\lambda_{1}$};
\node[nodeunknown] (c11) at (1.25,1) {};
\node[font=\tiny] (c11label) at (1.25,1) {$\phi_{1_{1}}$};
\node[nodeunknown] (c12) at (1.25,2) {};
\node[font=\tiny] (c12label) at (1.25,2) {$\phi_{1_{2}}$};
\node[nodeunknown] (c1n) at (1.25,3.625) {};
\node[font=\tiny] (c1nlabel) at (1.25,3.625) {$\phi_{1_{n}}$};
\node[nodeknown] (g11) at (1.75,1.5) {};
\node[font=\tiny] (g11label) at (1.75,1.5) {$\gamma_{1_{1}}$};
\node[nodeknown] (g12) at (1.75,2.5) {};
\node[font=\tiny] (g12label) at (1.75,2.5) {$\gamma_{1_{2}}$};
\node[] (g1dots) at (1.75,3.25) {$\vdots$};
\node[nodeknown] (g1n) at (1.75,4.125) {};
\node[font=\tiny] (g1nlabel) at (1.75,4.125) {$\gamma_{1_{n}}$};
\node[factor] (f11) at (1.75,1) {};
\node[factor] (f12) at (1.75,2) {};
\node[factor] (f1n) at (1.75,3.625) {};
\node[textnode] (l2) at (3.5,-0) {\footnotesize{\textit{and}}};
\node[nodeknown] (p2) at (3.5,0.5) {};
\node[font=\tiny] (p2label) at (3.5,0.5) {$\lambda_{2}$};
\node[nodeunknown] (c21) at (3,1) {};
\node[font=\tiny] (c21label) at (3,1) {$\phi_{2_{1}}$};
\node[nodeunknown] (c22) at (3,2) {};
\node[font=\tiny] (c22label) at (3,2) {$\phi_{2_{2}}$};
\node[nodeunknown] (c2n) at (3,3.625) {};
\node[font=\tiny] (c2nlabel) at (3,3.625) {$\phi_{2_{n}}$};
\node[nodeknown] (g21) at (3.5,1.5) {};
\node[font=\tiny] (g21label) at (3.5,1.5) {$\gamma_{2_{1}}$};
\node[nodeknown] (g22) at (3.5,2.5) {};
\node[font=\tiny] (g22label) at (3.5,2.5) {$\gamma_{2_{2}}$};
\node[] (g2dots) at (3.5,3.25) {$\vdots$};
\node[nodeknown] (g2n) at (3.5,4.125) {};
\node[font=\tiny] (g2nlabel) at (3.5,4.125) {$\gamma_{2_{n}}$};
\node[factor] (f21) at (3.5,1) {};
\node[factor] (f22) at (3.5,2) {};
\node[factor] (f2n) at (3.5,3.625) {};
\node[textnode] (l3) at (5.25,-0) {\footnotesize{\textit{opens}}};
\node[nodeknown] (p3) at (5.25,0.5) {};
\node[font=\tiny] (p3label) at (5.25,0.5) {$\lambda_{3}$};
\node[nodeunknown] (c31) at (4.75,1) {};
\node[font=\tiny] (c31label) at (4.75,1) {$\phi_{3_{1}}$};
\node[nodeunknown] (c32) at (4.75,2) {};
\node[font=\tiny] (c32label) at (4.75,2) {$\phi_{3_{2}}$};
\node[nodeunknown] (c3n) at (4.75,3.625) {};
\node[font=\tiny] (c3nlabel) at (4.75,3.625) {$\phi_{3_{n}}$};
\node[nodeknown] (g31) at (5.25,1.5) {};
\node[font=\tiny] (g31label) at (5.25,1.5) {$\gamma_{3_{1}}$};
\node[nodeknown] (g32) at (5.25,2.5) {};
\node[font=\tiny] (g32label) at (5.25,2.5) {$\gamma_{3_{2}}$};
\node[] (g3dots) at (5.25,3.25) {$\vdots$};
\node[nodeknown] (g3n) at (5.25,4.125) {};
\node[font=\tiny] (g3nlabel) at (5.25,4.125) {$\gamma_{3_{n}}$};
\node[factor] (f31) at (5.25,1) {};
\node[factor] (f32) at (5.25,2) {};
\node[factor] (f3n) at (5.25,3.625) {};
\node[textnode] (l4) at (7,-0) {\footnotesize{\textit{closes}}};
\node[nodeknown] (p4) at (7,0.5) {};
\node[font=\tiny] (p4label) at (7,0.5) {$\lambda_{4}$};
\node[nodeunknown] (c41) at (6.5,1) {};
\node[font=\tiny] (c41label) at (6.5,1) {$\phi_{4_{1}}$};
\node[nodeunknown] (c42) at (6.5,2) {};
\node[font=\tiny] (c42label) at (6.5,2) {$\phi_{4_{2}}$};
\node[nodeunknown] (c4n) at (6.5,3.625) {};
\node[font=\tiny] (c4nlabel) at (6.5,3.625) {$\phi_{4_{n}}$};
\node[nodeknown] (g41) at (7,1.5) {};
\node[font=\tiny] (g41label) at (7,1.5) {$\gamma_{4_{1}}$};
\node[nodeknown] (g42) at (7,2.5) {};
\node[font=\tiny] (g42label) at (7,2.5) {$\gamma_{4_{2}}$};
\node[] (g4dots) at (7,3.25) {$\vdots$};
\node[nodeknown] (g4n) at (7,4.125) {};
\node[font=\tiny] (g4nlabel) at (7,4.125) {$\gamma_{4_{n}}$};
\node[factor] (f41) at (7,1) {};
\node[factor] (f42) at (7,2) {};
\node[factor] (f4n) at (7,3.625) {};
\node[textnode] (l5) at (8.75,-0) {\footnotesize{\textit{the cabinet drawers}}};
\node[nodeknown] (p5) at (8.75,0.5) {};
\node[font=\tiny] (p5label) at (8.75,0.5) {$\lambda_{5}$};
\node[nodeunknown] (c51) at (8.25,1) {};
\node[font=\tiny] (c51label) at (8.25,1) {$\phi_{5_{1}}$};
\node[nodeunknown] (c52) at (8.25,2) {};
\node[font=\tiny] (c52label) at (8.25,2) {$\phi_{5_{2}}$};
\node[nodeunknown] (c5n) at (8.25,3.625) {};
\node[font=\tiny] (c5nlabel) at (8.25,3.625) {$\phi_{5_{n}}$};
\node[nodeknown] (g51) at (8.75,1.5) {};
\node[font=\tiny] (g51label) at (8.75,1.5) {$\gamma_{5_{1}}$};
\node[nodeknown] (g52) at (8.75,2.5) {};
\node[font=\tiny] (g52label) at (8.75,2.5) {$\gamma_{5_{2}}$};
\node[] (g5dots) at (8.75,3.25) {$\vdots$};
\node[nodeknown] (g5n) at (8.75,4.125) {};
\node[font=\tiny] (g5nlabel) at (8.75,4.125) {$\gamma_{5_{n}}$};
\node[factor] (f51) at (8.75,1) {};
\node[factor] (f52) at (8.75,2) {};
\node[factor] (f5n) at (8.75,3.625) {};
\end{tikzpicture}
\vspace{-10px}
\end{center}
\caption{The DCG for the utterance ``a man opens and closes the cabinet drawers'' constructed from the parse tree illustrated in Figure~\ref{isrr2017:fig:architecture}. This model enumerates all possible groundings for each phrase and performs inference by searching over unknown correspondences. The expressed groundings (groundings for factors with \textsc{True}-valued correspondence variables) of factors connected to $\lambda_{0}$ are used as language-based observations that are fused with the visual observation.  In this example, the model should infer a ``prismatic'' relationship between objects of semantic classes ``cabinet'' and ''drawer''.}
\label{isrr2017:fig:dcg}
\end{figure}

Following the notation from \citet{paul16a}, we formulate this problem as one of inferring a distribution of symbols ($\Gamma$) representing objects ($\Gamma^{\mathcal{O}}$), relationships ($\Gamma^{\mathcal{R}}$), and affordances ($\Gamma^{\mathcal{A}}$) in the absence of an environment model for each utterance.  Object groundings are defined by an object type $o_{i}$ from a space of object types $\mathcal{O}$, relationship groundings are defined by a relationship type $r_{k}$ from a space of relationship types $\mathcal{R}$. Affordance groundings are defined by a pair of object types $o_{i}$ and $o_{j}$, and relationship type $r_{k}$:
\begin{subequations} \label{isrr2017:eqn:groundings}
    \begin{align}
\Gamma^{\mathcal{O}} &= \{ \gamma_{o_{i}}, o_{i} \in \mathcal{O} \} \\
\Gamma^{\mathcal{R}} &= \{ \gamma_{r_{k}}, r_{k} \in \mathcal{R} \} \\
\Gamma^{\mathcal{A}} &= \{ \gamma_{o_{i},o_{j},r_{k}}, o_{i}, o_{j} \in \mathcal{O}, r_{k} \in \mathcal{R} \}
    \end{align}
\end{subequations}

Examples of object types include ``chair'', ``desk'', and ``door'' which represent semantic classes of random variables inferred by visual perception.  Examples of relationship types include ``prismatic'' and ``revolute'' that represent translational and rotational motion.  The set of all groundings is defined as the union of these symbols:
\begin{equation}
\Gamma = \{ \Gamma^{\mathcal{O}} \cup \Gamma^{\mathcal{D}} \cup \Gamma^{\mathcal{A}} \}
\end{equation}

Extracting the most probable set of groundings from language is challenging due to the diversity inherent in free-form language and the complex relationships between the articulation of different objects.  For example, the verb ``open'' can be used to describe a person's interaction with both a drawer and a door, but the motion described in the former case is prismatic with a cabinet, while it is rotational with a wall in the latter.  We address these challenges by adapting the Distributed Correspondence Graph (DCG) \cite{howard-14} to the problem of affordance inference which formulates a probabilistic graphical model according to the parse structure of the sentence that is searched for the most likely binary correspondence variable $\phi_{i,j} \in \{\textsc{True},\textsc{False}\}$ between linguistic elements in the command $\lambda_{i}$, groundings $\gamma_{i,j} \in \Gamma$, and expressed groundings of child phrases $\Gamma_{c_{i}} \in \Gamma$. The DCG encodes the factors $f_{i,j}$ in the graph using log-linear models whose weights are learned from a corpus of annotated examples. We then perform inference over this model in a space of correspondence variables to arrive at a distribution over the kinematic model structure and parameters.  Note that since inference is conducted without an expressed environment model, the symbols express inferred relationships between semantic classes of pairwise objects. These are interleaved with the model constructed from visual perception to form a probabilistic model of the environment.  An example of the structure of the DCG for the utterance ``a man opens and closes the cabinet drawers'' is illustrated in Figure~\ref{isrr2017:fig:dcg}.

\subsection{Combining Vision and Language Observations}

The final step in our framework selects the kinematic graph structure
$\hat{\mathcal{M}} = \{\hat{M}_{ij}, \forall (ij) \in E_G \}$ that best explains the vision and language observations
$D_z = \{D_v, D_l\}$ from the space of all possible kinematic
graphs. We do so by maximizing the conditional posterior over the
model type associated with each edge in the graph $(ij) \in E_G$:
\begin{subequations}
    \begin{align}
      \hat{M}_{ij} &= \argmax{M_{ij}} p(M_{ij} \vert D_z) \\
      &= \argmax{M_{ij}} \int p(M_{ij},\theta_{ij} \vert D_z)  d\theta_{ij}
    \end{align}
\end{subequations}
Evaluating this likelihood is computationally prohibitive, so we use
the Bayesian Information Criterion (BIC) score as an approximation
\begin{equation}
    BIC(M_{ij}) = -2 \log p(D_z \vert M_{ij}, \hat{\theta}_{ij}) + k \log n,
\end{equation}
where $\hat{\theta}_{ij}$ is the maximum likelihood parameter estimate
(Eqn.~\ref{isrr2017:eqn:model-fitting}), $k$ is the number of parameters of the current model and $n$ is
the number of vision and language observations. We choose the model with the
lowest BIC score
\begin{equation} \label{isrr2017:eqn:bic-error}
    \hat{M}_{ij} = \argmin{M_{ij}} BIC(M_{ij})
\end{equation}
as that which specifies the kinematics of the object.

While our previous method~\citep{pillai14} only considers visual
measurements, our new framework performs this optimization over the joint
space of vision and language observations. Consequently, the BIC score
becomes
\begin{equation}
        BIC(M_{ij}) = -2 \Bigl(\log p(D_v \vert M_{ij},
        \hat{\theta}_{ij}) + \log p(D_l \vert M_{ij}, \hat{\theta}_{ij})\Bigr) + k\log n,
\end{equation}
where we have made the assumption that the language and vision
observations are conditionally independent given the model and
parameter estimates. We formulate the conditional likelihood of the linguistic observation according to the grounding likelihood $P(\Phi = \textsc{True} \vert \gamma_1, \ldots, \gamma_n, \Lambda)$ from the DCG language model. The grounding variables $\gamma_i$ denote affordances that express different kinematic structures that encode the articulation of the object, namely the relationship between its individual parts (Eqn.~\ref{isrr2017:eqn:groundings}). For each candidate model, we use the likelihood of the corresponding groundings under the learned DCG language model to compute the BIC score for the corresponding affordance. We then estimate the overall kinematic structure by solving for the
minimum spanning tree of the graph, where we define the cost of each
edge as
$\textrm{cost}_{ij} = -\log p(M_{ij}, \theta_{ij} \vert
D_z)$.
Such a spanning tree constitutes the kinematic graph that best
describes the vision and language observations.

\section{Results} \label{isrr2017:sec:results}

We evaluate our framework using a dataset of $78$ RGB-D videos in which a user
manipulates a variety of common
household and office objects (e.g., a microwave, refrigerator, and
filing cabinet). Each video is accompanied with $5$ textual descriptions provided by different human subjects using a web-based crowd-sourcing platform.
We split the dataset into separate training and test sets consisting of $22$ and $56$ videos, respectively.
AprilTags~\cite{olson11} were placed on each of the object parts in the test set to determine ground-truth motion. We train our language grounding model on a corpus of $50$ video descriptions relative to the training set composed of $28$ unique symbols composed of different object and/or relation types. Our language grounding model requires that every word in a given training sentence is
aligned with the corresponding symbol. Such alignment is not required at test time.

Of the $56$ test videos, $25$
involve single-part objects and $31$ involve multi-part objects.
The single-part object videos are used to demonstrate that
the addition of language observations can only improve the accuracy of
the learned kinematic models. The extent of these improvements on
single-part objects is limited by the relative ease of inference of
single degree-of-freedom motion. In the case of multi-part objects,
the larger space of candidate kinematic graphs makes vision-only
inference challenging, as feature tracking errors may result in
erroneous estimates of the graph structure.

\subsection{Evaluation Metrics}

We estimate the ground-truth kinematic models by performing MAP
inference based upon the motion trajectories observed using
AprilTags. We denote the resulting kinematic graph as $G^*$. The
kinematic type and parameters for each object part pair are denoted as
$M^*_{ij}$ and $\theta^*_{ij}$, respectively. Let $\hat{G}$,
$\hat{M}_{ij}$, $\hat{\theta}_{ij} $ be the estimated kinematic
graph, kinematic type, and parameters for each object pair from the
RGB-D video, respectively.

The first metric that we consider evaluates whether the vision
component estimates the correct number of parts.  We determine the
ground-truth number of parts as the number of AprilTags observed in
each video, which we denote as $N^*$. We indicate the number of parts
(motion clusters) identified by the visual pipeline as $N_v$. We
report the average success rate when using only visual observations as
\mbox{$S_v = \frac{1}{K} \sum_{k=1}^{K} \mathbb{1}(N_v^k = N^{k*})$}, where
$K$ is the number of videos for each object type.

Next, we consider two metrics that assess the ability of each method
to estimate a graph with the same kinematic model as the ground truth
$G^*$. The first metric requires that the two graphs have the same
structure, i.e.,
$\hat{M}_{ij} = M^*_{ij}, \forall (ij) \in E_{\hat{G}} =
E_{G^*}$.
This equivalence requires that vision-only inference yields the
correct number of object parts and that the model selection framework
selects the correct kinematic edge type for each pair of object
parts. We report this ``hard'' success rate $S_h$ in terms of the
fraction of demonstrations for which the model estimate agrees with
the ground-truth. Note that this is bounded from above by fraction for
which the vision component estimates the correct number of parts. The second ``soft'' success rate (denoted by $S_{s}$) employs a relaxed
requirement whereby we only consider the inter-part relationships
identified from vision, i.e.,
$\hat{M}_{ij} = M^*_{ij}, \forall (ij) \in E_{\hat{G}} \subset
E_{G^*}$. In this way, we consider scenarios for which the visual system detects
fewer parts than are in the ground-truth model. In our experiments, we
found that $\hat{G}$ is a sub-graph of $G^*$, so we only require that
the model type of the edges in this sub-graph agree between both
graphs. The metric reports the fraction of total demonstrations for
which the estimated kinematic graph is a correct sub-graph of the ground-truth
kinematic graph.

Once we have the same kinematic models for both $\hat{G}$ and $G^*$,
we can compare the kinematic parameters $\hat{\theta}_{ij}$ to the
ground-truth values $\theta^*_{ij} $ for each inter-part model
$\hat{M}_{ij}$. Note that for the soft metric, we only compare
kinematic parameters for edges in the sub-graph, i.e.,
\mbox{$\forall (ij) \in E_{\hat{G}} \subset E_{G^*}$}. We define the parameter
estimation error for a particular part pair as the angle between the two kinematic parameter axes
\begin{equation}
    \textrm{e}_{ij} = \arccos \frac{\hat{\theta}_{ij} \cdot \theta^*_{ij}}{\lVert \hat{\theta}_{ij} \rVert \lVert \theta^*_{ij} \rVert},
\end{equation}
where we use the directional and rotational axes for prismatic and
rotational degrees-of-freedom, respectively. We measure the overall parameter estimation error $e_\textrm{param}$ for an object as the average parameter estimation error over each edge in the object's kinematic graph. We report this error further averaged over the number of demonstrations.
We do not report the estimation error for the object classes \textit{Drawer (Single-Part)} and
\textit{Door (Multi-Part)} because the demonstrations for those classes do not contain the fiducial
markers needed to compute the ground-truth.

\subsection{Results and Analysis}
\begin{table*}[!t]
    \centering
    \def\arraystretch{1.3}
    \begin{tabular}{lcccccc@{\hspace{12pt}}cccccc@{}}
      \toprule
      &&&&&& \multicolumn{2}{c}{Vision-Only} && \multicolumn{2}{c}{Ours} & \\
      \cmidrule{7-8} \cmidrule{10-11} %
      & Object & $K$ & $N^*$ & $N_v$ & $S_v$ & $S_{h}$ & $S_{s}$ && $S_{h}$ & $S_{s}$ & $e_{param}$\\
      \midrule
      \multirow{5}{*}{Single-Part} &
      Door & 9 & 1  & 1(9)  & 9/9 & 5/9  & 5/9 && \textbf{9/9} & \textbf{9/9} & $1.86^\circ$ \\
      & Chair & 5 & 1 & 1(4), 3(1) & 4/5 & 1/5 & 2/5 && \textbf{4/5}  & \textbf{5/5} & $3.34^\circ$\\
      & Refrigerator & 5 & 1 & 1(5) & 5/5 & 5/5 & 5/5 && 5/5 & 5/5 & $5.74^\circ$\\
      & Microwave & 4 & 1 & 1(3), 2(1) & 3/4 & 3/4 & 4/4 && 3/4 & 4/4 & $2.02^\circ$\\
      & Drawer & 2 & 1 & 1(2) & 2/2 & 0/2 & 0/2 && \textbf{2/2} & \textbf{2/2} & --\\
      \midrule
      \multirow{5}{*}{Multi-Part} &
      Chair & 4 & 2 & 1(2), 2(2) & 2/4 & 1/4 & 5/8 && \textbf{2/4} & \textbf{6/8} & $3.05^\circ$\\
      & Monitor & 7 & 2 & 1(7) & 0/7 & 0/7 & 6/14 && 0/7 & {\bf 7/14} & $7.27^\circ$\\
      & Bicycle & 7 & 3 & 1(1), 2(4), 3(2) & 2/7& 0/7 & 13/21 && 0/7 & 13/21 & $11.33^\circ$\\
      & Drawer & 11 & 2 & 1(6), 2(4), 3(1) & 4/11 & 3/11 & 10/22 && \textbf{4/11} & \textbf{15/22} & $0.11^\circ$\\
      & Door & 2 & 2 & 2(2) & 2/2 & 0/2 & 2/4 && \textbf{2/2} & \textbf{4/4} & --\\
      \bottomrule
      \end{tabular}
    \caption{Overall performance of our framework on video-description pairs. \label{isrr2017:tab:main}}
\end{table*}
Table~\ref{isrr2017:tab:main} provides a summary that compares the performance of our multimodal
learning method against that of the vision-only
baseline~\citep{pillai14}, which the authors have previously compared favorably to alternative state-of-the-art formulations~\cite{katz13}. We limit our comparison to methods that similarly do not assume that the number of parts is known a priori and that do not require the use of fiducial markers. The table indicates the number of demonstrations $K$,
the ground-truth number
of parts for each object $N^*$, a list of the number of parts identified using
visual trajectory clustering for each demonstration $N_v$, and the
fraction of videos for which the correct number of parts was
identified $S_v$. The number in parenthesis under $N_v$ indicates
the number of demonstrations for which a specific number of parts $N_v$ was identified.
We then present the hard $S_h$ and soft $S_s$
model selection rates for our method as well as for the baseline.
The denominators listed under $S_h$ and $S_s$ indicate the number of parts in each demonstration.
In the case of a single-part object, this matches the
number of demonstrations $K$ for that object. In the case of a multi-part object, while the denominator for $S_h$ matches the number of demonstrations $K$, the
denominator for $S_s$ indicates the total number of parts (i.e., $K * N^{*}$).
Our method bests the vision-only baseline in estimating the full kinematic
graph for five of the eight object classes, matching its performance on the
remaining three objects. Specifically, our framework yields accurate
estimates of the full kinematic graphs for thirteen more demonstrations
than the vision-only baseline, nine more for single-part objects and
four more for multi-part objects, corresponding to a $23\%$ absolute improvement. Similarly, we are able to estimate a valid sub-graph of the ground-truth kinematic graph for eighteen more demonstrations than the vision-only baseline
(eleven for single-part and seven for multi-part objects), corresponding to a $19\%$ absolute improvement. One notable object on which both
methods have difficulty is the bicycle for which the trajectory
clustering method was unable to identify the presence of the third
part (the wheel) due to the sparsity of visual features. Consequently, neither method estimated the full kinematic graph for any video.
Similarly, clustering
failed to identify the three parts comprising the monitor in all
videos, however our framework exploits language to estimate an accurate sub-graph for one more video.

We then evaluate the accuracy of the parameters estimated by our
method by reporting the parameter estimation error for each object,
averaged over the set of videos. Note that it is difficult to compare
against the error of the vision-only baseline since it does not yield
accurate kinematic graphs for several of the videos. When the kinematic graph
estimates agree, however, the parameter estimation errors are
identical for the two methods, since they both estimate the parameters
from the visual data (Eqn.~\ref{isrr2017:eqn:model-fitting}).

\section{Conclusion}
\label{isrr2017:sec:conclusion}

We have described a method that uses a joint combination of vision- and
language-based observations to learn accurate probabilistic models that define the structure and parameters of articulated objects. Our framework treats linguistic descriptions of a demonstrated motion as a complementary
observation of the structure of kinematic linkages. We evaluate our
framework on a series of RGB-D video-description pairs involving the manipulation of common household and office objects. The results demonstrate that exploiting language as a form of weak supervision improves the accuracy of the inferred model structure and parameters. While this evaluation considers videos paired with free-form descriptions, the method does not rely on any assumptions that preclude its application to robotic systems. Future work includes incorporating semantic segmentation as a means of assigning perceived labels to inferred clusters, using the
description to mitigate noise in the visual recognition. Additionally, kinematic models provide a common representation that is suited to generalization across different object instances. We are extending our model to predict kinematic models of novel (i.e., unseen) articulated objects, using natural language as a means of knowledge transfer.

    \newpage
    \chapter{\autolabsec}
    \label{iros2020:sec:main}
    This work was published in~\cite{tani20}.\\
    \begin{figure}[h]
\centering
    \includegraphics[width=0.95\columnwidth, trim = 0 0 0 1.2cm, clip]{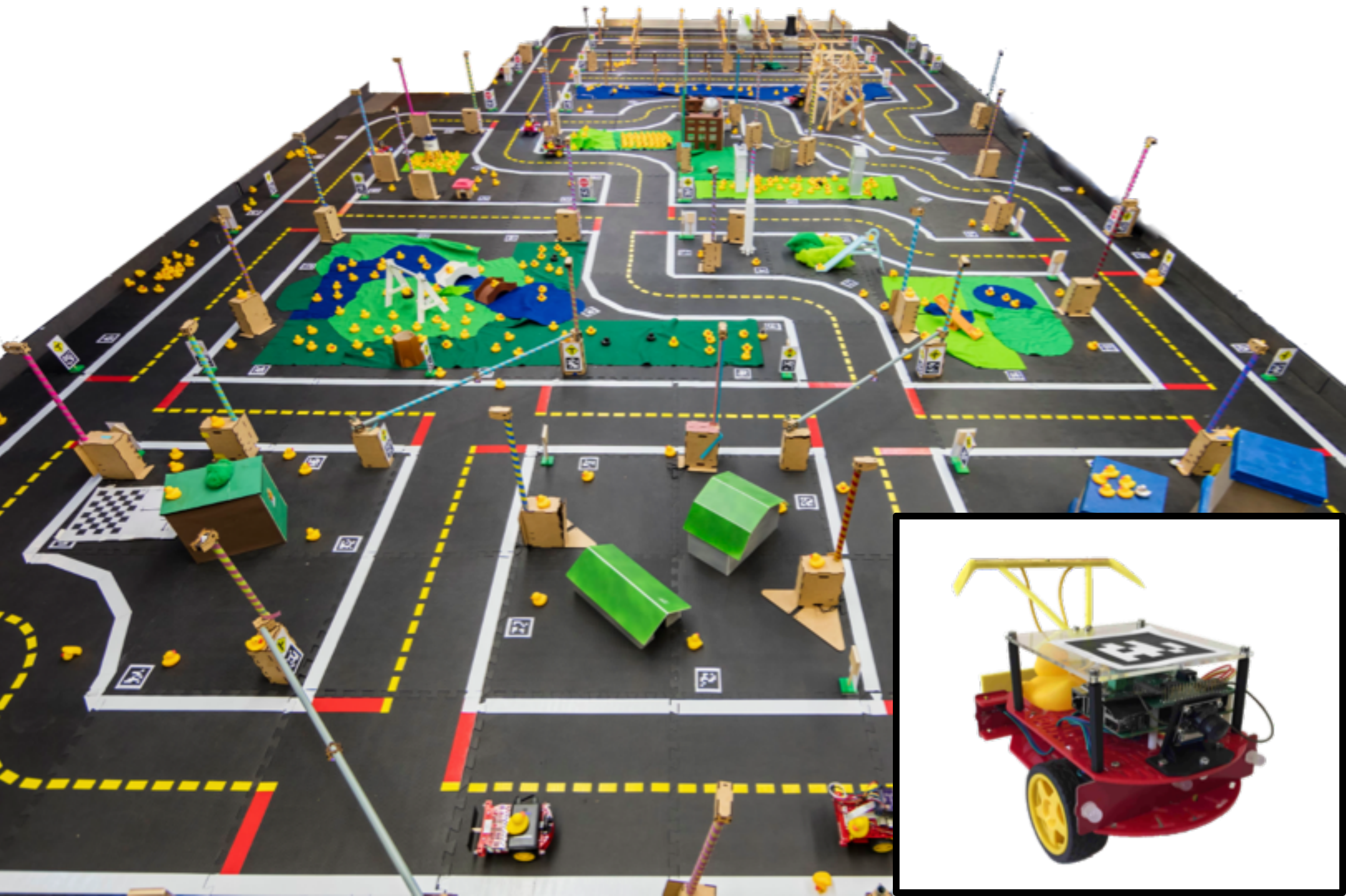}
    \caption{\textbf{The Duckietown Autolab}: We augment Duckietowns~\cite{paull2017duckietown} with localization and automatic charging infrastructure. We modify Duckiebots (inset) to facilitate detection by the localization system and auto-charging.}
    \label{iros2020:fig:db18-autolab}
\end{figure}

Mobile robotics poses unique challenges that have precluded the establishment of benchmarks for rigorous evaluation. Robotic systems are \emph{complex} with many different interacting components. As a result, evaluating the individual components is not a good proxy for full system evaluation. Moreover, the outputs of robotic systems are temporally correlated (due to the system dynamics) and partially observable (due to the perception system). Disentangling these issues for proper evaluation is daunting.

The majority of research on mobile robotics takes place in idealized laboratory settings or in unique uncontrolled environments that make comparison difficult. Hence, the value of a specific result is either open to interpretation or conditioned on specifics of the setup that are not necessarily reported as part of the presentation.
The issue of reproducibility is exacerbated by the recent emergence of \emph{data-driven} approaches, the performance of which can vary dramatically and unpredictably across seemingly identical environments (e.g., by only varying the random seed~\cite{Henderson2017}). It is increasingly important to fairly compare these data-driven approaches with the more classical methods or hybrids of the two.

Most existing methods for evaluating robotics operate on individual, isolated components of the system. For example, evaluating robot perception is comparatively straightforward and typically relies on annotated datasets~\cite{kitti, cityscapes, nuscenes, waymo-sun2019scalability}. However, performance on these benchmark datasets is often not indicative of an algorithm's performance in practice.
A common approach to analyzing robot control algorithms is to abstract away the effects of perception~\cite{yu2020introduction, robotarium} and assume that the pose of the robot is known (e.g., determined using an external localization system, such as a motion capture setup~\cite{vicon-merriaux2017study}).

Simulation environments are potentially valuable tools for system-level evaluation. Examples such as CARLA~\cite{dosovitskiy2017carla}, AirSim~\cite{shah2018airsim}, and Air Learning~\cite{krishnan2019air} have recently been developed for this purpose. However, a challenge with simulation-based evaluation is that it is difficult to quantify the extent to which the results extend to the real world.

Robotics competitions have been excellent testbeds for more rigorous evaluation of robotics algorithms~\cite{amigoni2015competitions}. For example, the DARPA urban challenge~\cite{buehler2009darpa}, the DARPA robotics challenge~\cite{johnson2015team}, and the Amazon Picking Challenge~\cite{correll2016analysis}, have all resulted in massive development in their respective sub-fields (autonomous driving, humanoid robotics, and manipulation respectively). However, the cost and multi-year effort required to join these challenges limit participation.

A very promising recent trend that was spearheaded by the Robotarium at Georgia Tech~\cite{pickem2017robotarium, wilson2020robotarium} is to provide remotely accessible lab setups for evaluation. This approach has the key advantage that it enables access for anyone to submit. In the case of the Robotarium, the facility itself cost 2.5M \$US, and would therefore be difficult to replicate. Additionally, while it does allow users flexibility in the algorithms they can run, it does not offer any standardized evaluation.

 In this work, we propose to harmonize all the above-mentioned elements (problem definition, benchmarks, development, simulation, annotated logs, experiments, etc.) in a \quotes{closed-loop} design that has minimal software and hardware requirements. This framework allows users to define benchmarks that can be evaluated with different modalities (in simulation or on real robots) locally or remotely. Importantly, the design provides immediate feedback from the evaluations to the user in the form of logged data and scores based on the metrics defined in the benchmark.

We present the \ac{DUCKIENet}, an instantiation of this design based on the Duckietown platform~\cite{paull2017duckietown} that provides an accessible and reproducible framework focused on autonomous vehicle fleets operating in model urban environments.
The \ac{DUCKIENet} enables users to develop and test a wide variety of different algorithms using available resources (simulator, logs, cloud evaluations, etc.), and then deploy their algorithms locally in simulation, locally on a robot, in a cloud-based simulation, or on a real robot in a remote lab. In each case, the submitter receives feedback and scores based on well-defined metrics.

The \ac{DUCKIENet} includes \acfp{DTA}, remote labs that are also low-cost, standardized and fully documented with regards to assembly and operation, making this approach highly scalable. Features of the \acp{DTA} include an off-the-shelf camera-based localization system and a custom automatic robot recharging infrastructure. The accessibility of the hardware testing environment through the \ac{DUCKIENet} enables experimental benchmarking that can be performed on a network of \acp{DTA} in different geographical locations.

We will: summarize the objectives of our integrated benchmarking system in Section~\ref{iros2020:sec:preliminaries}; describe an instantiation that adheres to these objectives, the DUCKIENet, in Section~\ref{iros2020:sec:dn}; validate the performance and usefulness of the approach in Section~\ref{iros2020:sec:dn-validation}; and finally present conclusions in Section~\ref{iros2020:sec:conclusion}.

\section{Integrated benchmarking and development for reproducible research}
\label{iros2020:sec:preliminaries}

\begin{figure}[!t]
\centering
    \includegraphics[width=0.6\columnwidth]{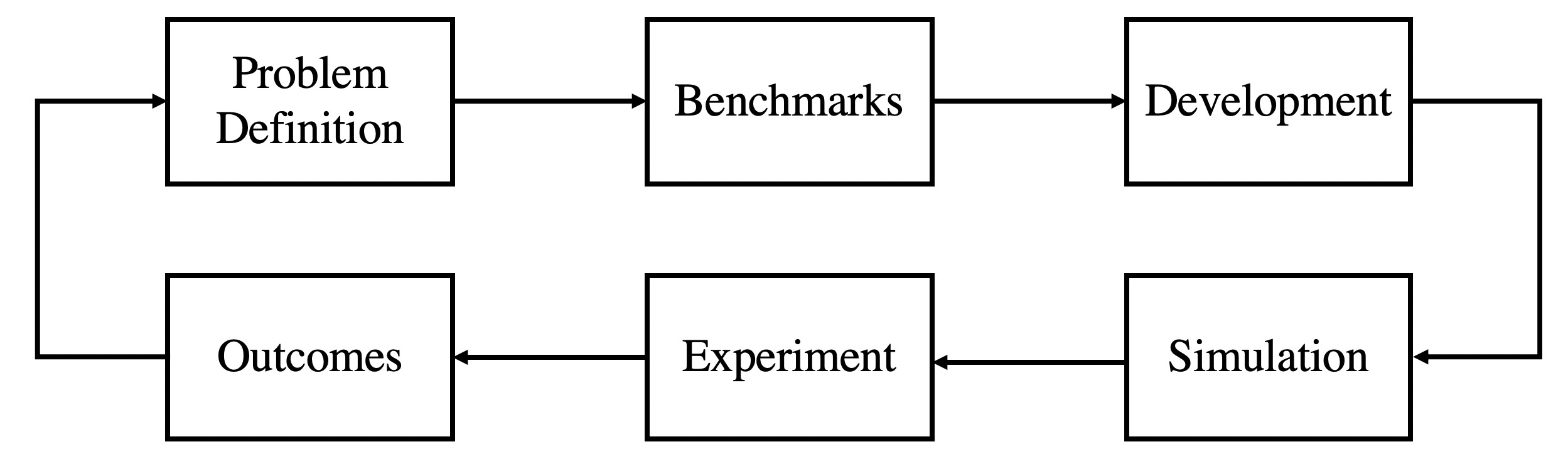}
     \caption{\textbf{Integrating Benchmark Design}: Designing robotic benchmarks should be more tightly integrating into robotic system development.}
    \label{iros2020:fig:int-loop}
\end{figure}

Benchmarking should not be considered an isolated activity that occurs as an afterthought of development. Rather, it should be considered as an integral part of the research and development process itself, analogous to test-driven development in software design. %
Robotics development should be a feedback loop that includes problem definition, benchmarks specification, development, testing in simulation and on hardware, and using the results to adjust the problem definition and the benchmarks (Figure~\ref{iros2020:fig:int-loop}).

We describe a system built to explore the benefits and the challenges of harmonizing development, benchmarking, simulation and experiments to obtain reproducibility \quotes{by design}.
In this section we give an overview of the main challenges and design choices.

\subsection{Reproducibility}
Our main objective is to achieve \quotes{reproducibility}. At an abstract level, reproducibility is the ability to reduce variability in the evaluation of a system such that experimental results are more credible~\cite{Goodman341ps12}.
More specifically, there are three distinct aspects to reproducibility, all equally important:
\begin{enumerate}
    \item An \emph{experiment} is reproducible if the results obtained at different times and by different people are similar;
    \item An \emph{experimental setup} is reproducible if it is possible to build a copy of it elsewhere and obtain comparable results;
    \item  An \emph{experimental platform} is reproducible if it is relatively easy, in terms of cost and complexity, to replicate.

\end{enumerate}

\subsubsection{Software}
Modern autonomy stacks consist of many modules and dependencies, and the reproducibility target implies being able to use the same software possibly many years later.

Standardized software archival systems such as Git solve part of the problem, as it is easy to guarantee storage and integrity of the software over many years~\cite{blischak2016quick}. However, even if the core implementation of the algorithm is preserved, the properties of the system change significantly in practice due to the external dependencies (libraries, operating system, etc.). In extreme cases \quotes{bit rot} causes software to stop working simply because the dependencies are no longer maintained~\cite{boettiger2015introduction}.

Containerization technologies such as Docker~\cite{Docker} have dramatically reduced the effort required to create reproducible software. Containerization solves the problem of compilation reproducibility, as it standardizes the environment in which the code runs~\cite{weisz2016robobench}.
Furthermore, the ability to easily store and retrieve particular builds (\quotes{images}) through the use of registries, solves the problem of effectively freezing dependencies.  This allows perfect reproducibility for a time span of 5--10 years (until the images can no longer be easily run using the current version of the containerization platform).

\subsubsection{Hardware}
In addition to issues related to software, a central challenge in the verification and validation of robot autonomy is that the results typically depend on under-defined properties of the particular hardware platforms used to run the experiments. While storage and containerization make it possible to standardize the version and build of software, hardware will never be fully reproducible, and using only one instance of a platform is impossible due to wear and tear.

As a result, we propose to make several samples from a \emph{distribution} of hardware platforms that are used for evaluation. Consequently, metrics should be considered statistically  (e.g., via mean and covariance), similarly to recent standard practices in reinforcement learning that evaluate over a set of random seeds~\cite{Henderson2017}. This incentivizes algorithms that are robust to hardware variation, and will consequently be more likely to work in practice. Also, one can define the distribution of variations in simulations such that configurations are sampled by the benchmark on each episode.
While the distribution as a whole might change (due to different components being replaced over the years), the distributional shift is easier to detect by simply comparing the performance of an agent on the previous and current distributions.

Cost can also quickly become the limiting factor for the reproducibility of many platforms. Our objective is to provide a robotic platform, a standardized environment in which it should operate, and a method of rigorous evaluation that is relatively inexpensive to acquire and construct.

\subsubsection{Environment}

\begin{wrapfigure}{r}{0.5\textwidth}
    \vspace{-1.1cm}
    \centering
    \includegraphics[width=0.48\textwidth]{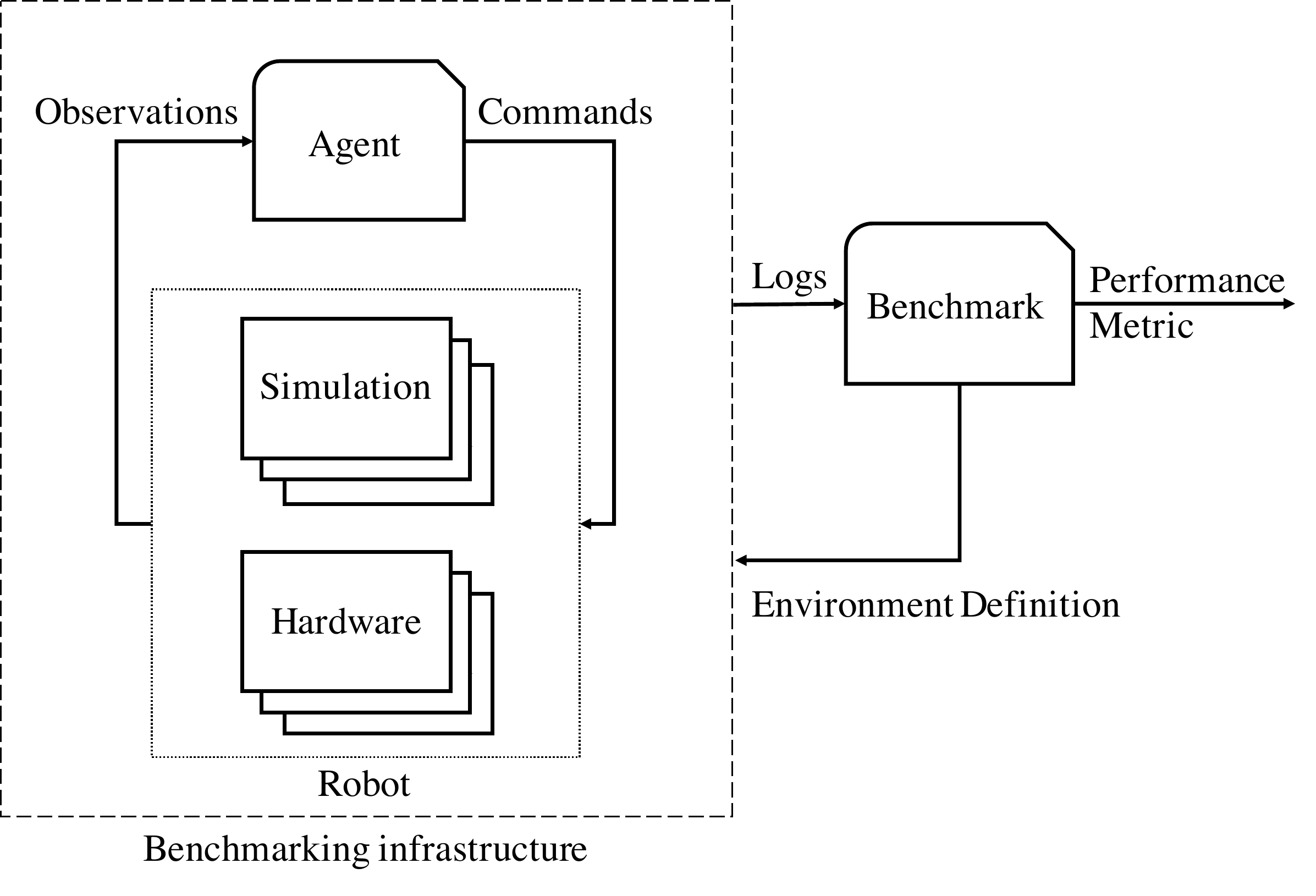}
    \caption{\textbf{Abstractions and Interfaces}: An interface must be well-specified so that components on either side of the interface can be swapped in and out. The most important is the Agent-Robot interface, where the robot can be physical or a simulator. The interface between the benchmark and the infrastructure enables a clear definition of benchmark tasks (i.e., metrics) separately from the means by which they are evaluated.}
    \label{iros2020:fig:interfaces}
    \vspace{-1cm}
\end{wrapfigure}

We use an analogous approach to deal with variability in the environment. The first and most crucial step for the reproducibility of the setup is to standardize the formal description of the environment. Nevertheless, it is infeasible to expect that even a well-described environment can be exactly reproduced in every instance.
As discussed in Section~\ref{iros2020:sec:dn}, our approach is to use a distributed set of experimental setups (the \acp{DTA}) and test agents across different installations that exist around the world. 
Similar to the way that robot hardware should be randomized, proper evaluation should include sampling from a distribution over environments.

\subsection{Agent Interoperability}
\label{iros2020:sec:interoperability}

Open-source robotics middlewares, such as ROS~\cite{ROS} and LCM~\cite{huang10}, are extremely useful thanks to the tools they provide. However, heavily leveraging these software infrastructures is an impediment to long-term software reproducibility since backward/forward compatibility is not always ensured across versions.
As a result, we propose a framework based on low-level primitives that are extremely unlikely to become obsolete, while providing \emph{bridges} to popular frameworks and middlewares. Specifically, components in our infrastructure communicate using pipes (FIFOs) and a protocol based on Concise Binary Object Representation (CBOR) (a binarized version of JavaScript Object Notation)~\cite{cbor}.

These standardized protocols  have existed  for decades and will likely persist long into the future.
It is then the job of the infrastructure to interconnect components, either directly or using current middleware (e.g., ROS) for transport over networks. In the future, the choice of middleware might change without invalidating the stored agents as a new bridge can be built to the new middleware. An additional benefit of CBOR is that it is self-describing (i.e., no need for a previously agreed schema to decode the data) and allows some form of extensibility since schema can change or be updated and maintain backward compatibility.

\subsection{Robot-Agent Abstraction}

Essential to our approach is defining the interface between the \emph{agent} and the \emph{robot} (whether the robot is in a simulator or real hardware). Fundamentally, we define this interface in the following way (Figure~\ref{iros2020:fig:interfaces}):
\begin{itemize}
\item \textbf{Robots} are nodes that consume actuation commands and produce sensor observations.
\item \textbf{Agents} are nodes that consume sensor observations and produce actuation commands.
\end{itemize}

Combining this with containerization and the low-level reproducible communication protocol introduced in \mbox{Section~\ref{iros2020:sec:interoperability}} results in a framework that is reliable and enables  the easy exchange of either  agents or robots in a flexible and straightforward manner. As described in Section~\ref{iros2020:sec:dn}, this approach is leveraged in the DUCKIENet to enable zero friction transitions from local development, to remote simulation, and to remote experiments in \acp{DTA}.   %

\subsection{Robot-Benchmark Abstraction}
A benchmarking system becomes very powerful and extensible if there is a way to distinguish the infrastructure from the particular benchmark being run. In similar fashion to the definition of the interface for the robot system, we standardize the definition of a benchmark and its implementation as containers (Figure~\ref{iros2020:fig:interfaces}):
\begin{itemize}
    \item \textbf{Benchmarks} provide environment configurations and consume robot states to evaluate well-define metrics.
    \item \textbf{Infrastructure} consumes environment configurations and provides estimates of robot state, either directly (e.g., through a simulator) or through other external means.
\end{itemize}

\section{The \ac{DUCKIENet}}
\label{iros2020:sec:dn}

In this section we describe an instantiation of the high-level objectives proposed in Section~\ref{iros2020:sec:preliminaries} we refer to as \ac{DUCKIENet}, which:
\begin{itemize}
    \item comprises affordable robot hardware and environments;
    \item adheres to the proposed abstractions to allow easy evaluation on local hardware or in simulation;
    \item enables evaluation of user-specified benchmarks in the cloud or in a remote, standardized lab;
\end{itemize}

\begin{figure}[!t]
\centering
    \includegraphics[width=0.75\columnwidth]{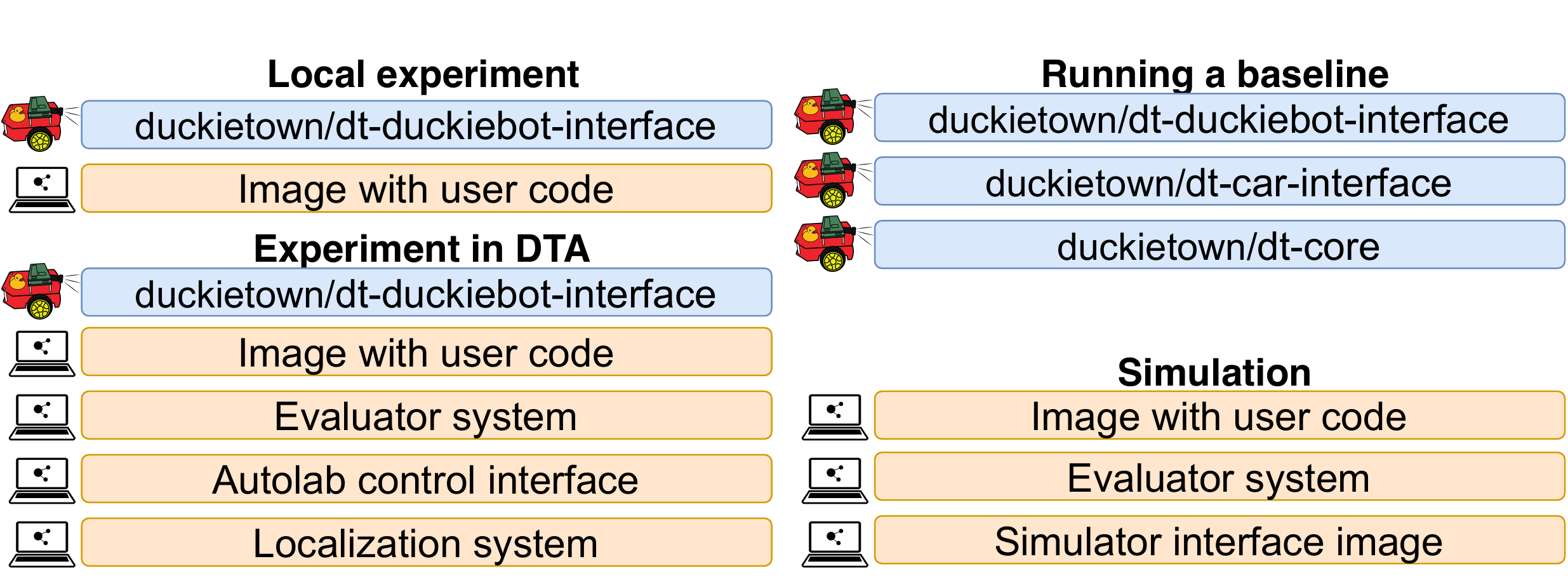}
    \caption{\textbf{Use-cases}: The well-defined interfaces enable execution of an agent in different settings, including locally in hardware, in a \ac{DTA}, with a standard baseline, or in a simulator. Different icons indicate where a container is executed (either on the robot
    \protect\includegraphics[height=0.85em]{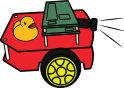} or on the laptop \protect\includegraphics[height=0.85em]{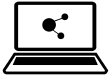}).}
    \label{iros2020:fig:dt-usecases}
\end{figure}

\begin{figure}[!t]
\centering
    \includegraphics[width=0.7\columnwidth]{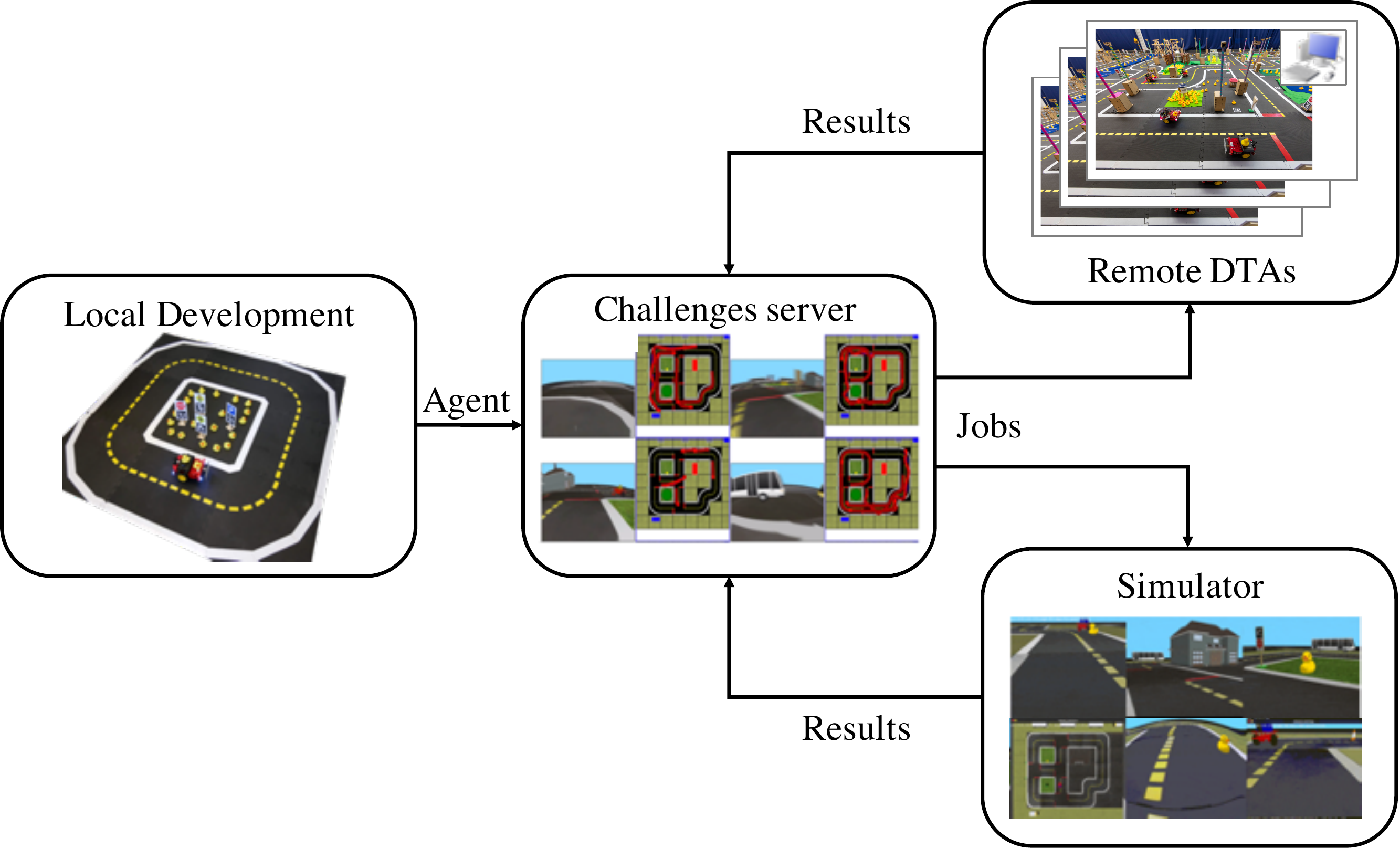}
    \caption{\textbf{The user workflow}: Local development can take place on the real hardware or in a simulator. Agents are submitted to the Challenges server, which marshals agents to be submitted in simulation challenges to the cloud simulation, or agents that should be evaluated on real hardware to a \ac{DTA}. The results of these evaluations are combined with the benchmark definition and made available to the submitter.}
    \label{iros2020:fig:duckienet-high-level}
\end{figure}

\subsection{The Base Platform}

The \ac{DUCKIENet} is built using the Duckietown platform, which was originally developed in 2016 and has since been used for education~\cite{tani2016duckietown}, research~\cite{aido} and outreach~\cite{dt-org}.

\subsubsection{The Duckietown hardware platform} \label{iros2020:sec:dt-hw}

The Duckietown hardware consists of Duckiebots and Duckietown urban environments. 
Duckiebots (bottom-right corner of Figure~\ref{iros2020:fig:db18-autolab}) are differential-drive mobile robots with a front-facing camera as the only sensor, five LEDs used for communication, two DC-motors for actuation, and a Raspberry Pi for computation. Duckietowns(Figure~\ref{iros2020:fig:db18-autolab}) are structured modular urban environments with floor and signal layers. The floor layer includes five types of road segments: straight, curve, three-way intersection, four-way intersection, and empty. The signal layer includes traffic signs and traffic lights. Traffic lights have the same hardware as the Duckiebots, excluding the wheels, and are capable of sensing, computing, and actuation through their LEDs.

\subsubsection{The Duckietown software architecture} \label{iros2020:sec:dt-sw}

We implement the Duckietown base software as ROS nodes and use the ROS topic messaging system for inter-process communication. The nodes are organized in Docker images that can be composed to satisfy various use-cases (Figure~\ref{iros2020:fig:dt-usecases}). Specifically, the components that correspond to the ``robot'' are the actuation and the sensing, and all other nodes correspond to the agent.
The Duckietown simulator~\cite{gym_duckietown} is a lightweight, modular, customizable virtual environment for rapid debugging and training of agents. It can replace the physical robot by subscribing to commands and producing sensor observations.
We provide an agent interface to the OpenAI Gym~\cite{brockman2016openai} protocol that can be called at a chosen frequency to speed up training.
In the case of reinforcement learning, rewards can be specified and are returned, enabling one to easily set up episodes for training.

The process of marshalling containers requires some specific configurations. To ease this process, we have developed the Duckietown Shell which is a wrapper around the Docker commands and provides an interface to the most frequent development and evaluation operations~\cite{dt-shell}. Additionally, the Duckietown Dashboard provides an intuitive high-level web-based graphical user interface~\cite{dt-dashboard}. The result is that agents that are developed in one modality (e.g., the simulator) can be evaluated in the cloud, on real robot hardware, or in the \ac{DTA} with one simple command or the at click of a button.

\subsection{System Architecture Overview}

The \ac{DUCKIENet} (Figure~\ref{iros2020:fig:duckienet-high-level}) is comprised of: (i) a Challenges server that stores agents, benchmarks, and results, computes leaderboards, and dispatches jobs to be executed by a distributed set of evaluators; (ii) a set of local or cloud-based evaluation machines that run the simulations; and (iii) one or more \acp{DTA}, physical instrumented labs that carry out physical experiments (Section~\ref{iros2020:sec:dn-dta}).
The system has:
\begin{enumerate}
\item \textbf{Super-users} who define the benchmarks. The benchmarks are defined as Docker containers submitted to the Challenges server. These users can also set permissions for access to different benchmarks, their results, and the scores.
\item \textbf{Regular developers} are users of the system.
They develop agents locally and submit their agent code as Docker containers. They can observe the results of the evaluation, though the system supports notions of testing and validation, which hides some of the results.
\item \textbf{Simulation evaluators} and \textbf{experimental evaluators} query the Challenges server for jobs to execute. They run jobs as they receive them and communicate the results back to the Challenges server.
\end{enumerate}

\subsection{The Duckietown Automated Laboratory (\ac{DTA})}
\label{iros2020:sec:dn-dta}

\begin{figure}[tb]
\centering
    \includegraphics[trim = 0 0 0 -30, width=\columnwidth, clip]{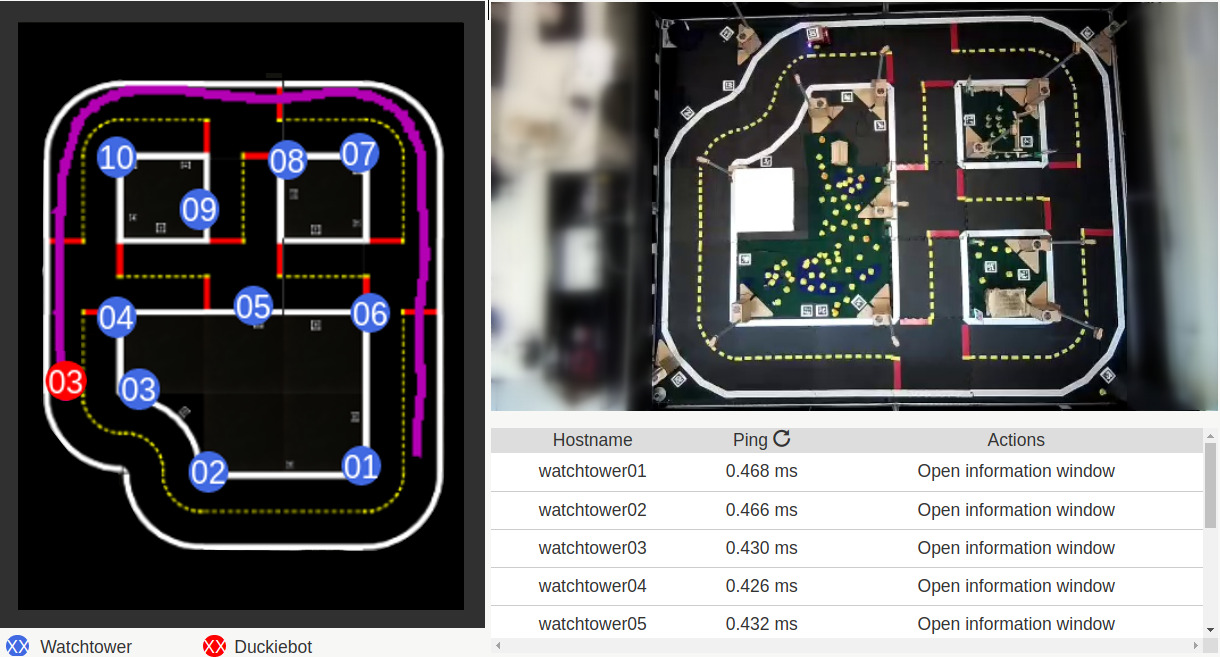}
    \caption{\textbf{The \ac{DTA} GUI} provides a control interface to human operators, including a representation of the map and the Watchtowers (left) and an overhead live feed of the physical Duckietown with diagnostic information about the experiment in progress (right). }
    \label{iros2020:fig:dta-ui}
\end{figure}

The \acp{DTA} are remotely accessible hardware ecosystems designed to run reproducible experiments in a controlled environment with standardized hardware (Section~\ref{iros2020:sec:dt-hw}). DTAs comprise: (a) a set of Duckietown environments (maps); (b) a localization system that estimates the pose of every Duckiebot in the map (Section~\ref{iros2020:sec:dn-autolocalization}); (c) a local experiment manager, which receives jobs to execute from the Challenges server and coordinates the local Duckiebots, performs diagnostics, and collects logs from the system; (d) a human operator responsible for tasks that are currently not automated (e.g., resetting experiments and
``auto"-charging); and (e) a set of Duckiebots modified with a upwards-facing AprilTag~\cite{AprilTags} and charge collection apparatus, as shown in Figure~\ref{iros2020:fig:db18-autolab}. A graphical UI (Figure~\ref{iros2020:fig:dta-ui}) makes the handling of experiments more user-friendly (Section~\ref{iros2020:sec:dta-ui}).

We evaluate the reproducibility of the \acp{DTA} themselves by evaluating agents across different labs at distributed locations worldwide, as described in Section~\ref{iros2020:sec:dn-validation}.

\subsubsection{Localization}
\label{iros2020:sec:dn-autolocalization}

The goal of the localization system is to provide pose estimates of all Duckiebots at any given time. These are then post-processed according to metrics specified in the benchmarks to evaluate agent performance. The localization system comprises a network of Watchtowers, i.e., low-cost observation structures using the same sensing and computation as the Duckiebots. 
The Watchtowers are placed such that their fields-of-view, which are restricted to a local neighbor region of each Watchtower, overlap and collectively cover the entire road network. A set of calibration AprilTags are fixed to the road layer of the map to provide reference points from the camera frame of each Watchtower to the map's fixed reference frame.

The localization system is decentralized and vision-based, using the camera streams from both the network of Watchtowers and the Duckiebots themselves. In each image frame, the system detects AprilTags, which are placed on the Duckiebots as well as in other locations around the city, e.g., on traffic signs at intersections. For each AprilTag, the system extracts their camera-relative pose. Using the resulting synchronized sequences of AprilTag-camera relative pose estimates, a pose graph-based estimate of the pose history of each AprilTag, using the $\text{g}^2 \text{o}$ library~\cite{g2o}, is returned.

\begin{figure}[t!]
\centering
    \includegraphics[trim = 0 0 0 -20,width=0.75\columnwidth]{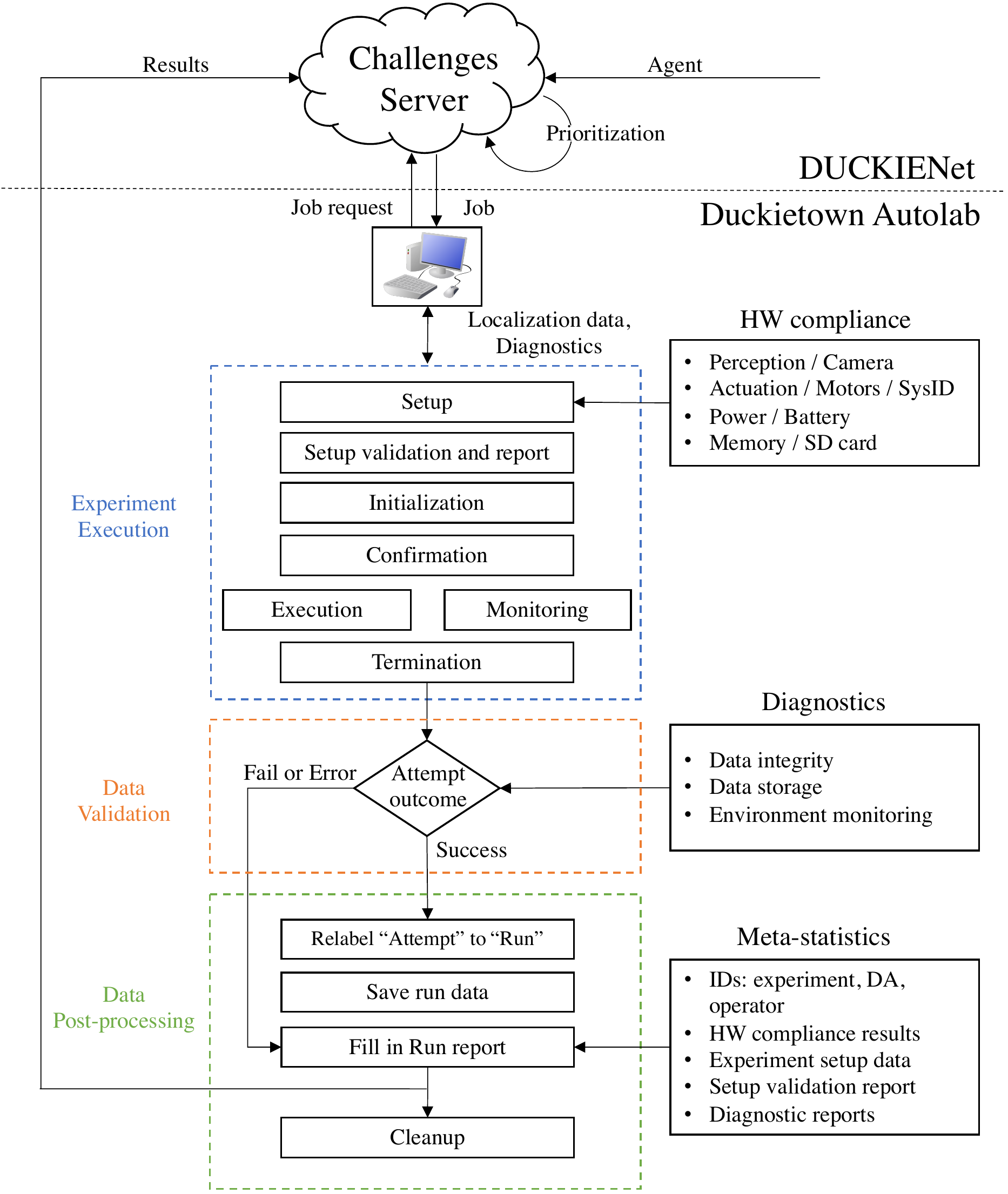}
    \caption{\textbf{Evaluation workflow}: An experiment is triggered by an agent being submitted to the Challenges server. The experiment is run in the \ac{DTA} following a predefined and deterministic procedure. The results are then uploaded to the Challenges server and returned to the submitter.}
    \label{iros2020:fig:duckienet-hw-experiment}
\end{figure}

\subsubsection{Operator console}
 \label{iros2020:sec:dta-ui}

DTAs are equipped with a graphical user interface based on the \textbackslash compose\textbackslash~web framework~\cite{compose} (Figure~\ref{iros2020:fig:dta-ui}). Similarly to the Duckietown Dashboard, it is designed to provide an accessible interface between the operator and the \ac{DTA}. Available functionalities include: various diagnostics (e.g., network latency, room illumination, camera feeds, etc.), the ability to run jobs (experiments) with a few clicks and compute, visualize and return results to the Challenges server, and the control of the environment illumination.

\subsection{Defining the Benchmarks}

Benchmarks can be defined either to be evaluated in simulation or on real robot hardware. Both types require the definition of:
(a) the environment, through a standardized map format;  (b) the \quotes{non-playing characters} and their behavior; and (c)  the evaluation metrics, which can be arbitrary functions defined on trajectories.

Each benchmark produces a set of scalar scores, where the designer can choose the final total order for scoring by describing a priority for the scores as well as tolerances (to define a lexicographic semiorder). The system allows one to define \quotes{benchmarks paths} as a directed acyclic graph of benchmarks with edges annotated with conditions for progressions. One simple use case is to prescribe that an agent be first evaluated in simulation before being evaluated on real robot hardware. It is possible to set thresholds on the various scores that need to be reached for progression. For competitions such as the AI Driving Olympics~\cite{aido}, the superusers can set up conditions such as \quotes{progress to the next stage only if the submission does better than a baseline}.

\subsection{DTA Operation Workflow}

The Challenges server sends a job to a \ac{DTA} that provides the instructions necessary to carry out the experiment (Figure~\ref{iros2020:fig:duckienet-hw-experiment}). The \ac{DTA} executes each experiment in a series of three steps: (i) experiment execution, (ii) data validation, and (iii) data post-processing and report generation.
Before the \ac{DTA} carries out an experiment, it first verifies that the robots pass a hardware compliance test that checks to see that their assembly, calibration, and the functionality of critical subsystems (sensing, actuation, power, and memory) are consistent with the specifications provided in the description of the experiment.

Having passed the compliance test, the experiment execution phase begins by initializing the pose of the robot(s) in the appropriate map, according to the experiment definition. The robot(s) that runs the user-defined agent is categorized as \quotes{active}, while the remaining \quotes{passive} robots run baseline agents. During the initialization step, the appropriate agents are transferred to each robot and initialized (as Docker containers). The localization system assigns a unique ID to each robot and verifies the initial pose. Once the robots are ready and the \ac{DTA} does not detect any abnormal conditions (e.g., related to room illumination and network performance), the \ac{DTA} sends a confirmation signal to synchronize the start of the experiment. The robots then execute their respective agents until a termination condition is met. Typical termination conditions include a robot exiting the road, crashing (i.e., not moving for longer than a threshold time), or reaching a time limit.

During the validation phase, the \ac{DTA} evaluates the integrity of the data, that it has been stored successfully, and that the experiment was not been terminated prematurely due to an invalidating condition (e.g., an unexpected change in environment lighting). Experiments that pass this phase are labeled as successful runs.

In the final phase, the \ac{DTA} compiles experiment data and passes it to the Challenges server. The Challenges server processes the data according to the specifications of the experiment and generates a report that summarizes the run, provides metrics and statistics unique to the run, as well as information that uniquely identifies the experiment, physical location, human operator, and hardware compliance. The entire report including the underlying data as well as optional comparative analysis with selected baselines are then shared with the user who submitted the agent. The Challenges server then performs cleanup in anticipation of the next job.

\subsection{The Artificial Intelligence Driving Olympics (AI-DO)}
\label{iros2020:sec:applications-aido}

One of the key applications of our infrastructure is to host competitions.  The \acf{AI-DO}~\cite{aido} is a bi-annual competition that leverages the \ac{DUCKIENet} to compare the ability of various agents, ranging from traditional estimation-planning-control architectures to end-to-end learning-based architectures, to perform autonomous driving tasks of increasing complexity. By fixing the environment and the hardware, agents and tasks can be modified to benchmark performance across different challenges. Most robotics challenges offer \emph{either} embodied or simulated experiments. The \ac{DUCKIENet} enables seamless evaluation in \emph{both} along with a visualization of performance metrics, leaderboards, and access to the underlying raw data. The \ac{DUCKIENet} makes the process of participating accessible, with open-source baselines and detailed documentation.

\section{Validation} \label{iros2020:sec:dn-validation}

We perform experiments to demonstrate (a) the repeatability of performance within a \ac{DTA}, (b) the reproducibility of experiments across different robots in the same \ac{DTA}, and (c) reproducibility across different facilities. 
To assess the repeatability of an experiment, we run the same agent (i.e., \ac{LF} ROS-based baseline), on the same Duckiebot and in the same \ac{DTA}. To evaluate inter-robot reproducibility, we run the same experiment on multiple Duckiebots, keeping everything else fixed. Finally, we demonstrate reproducibility of experiments across \ac{DTA}s.

We define two metrics to measure repeatability: \ac{MPD} and \ac{MOD}. Given a set of trajectories, \ac{MPD} at a point along a trajectory is computed as the mean lateral displacement of the Duckiebot from the center of the lane. Similarly, \ac{MOD} represents the mean orientation with respect to the lane orientation. For each set of experiments, we compute the average and the standard deviation along the trajectory itself.

\newpage
\subsection{Experiment Repeatability}

\begin{figure}[t]
\centering
    \expfigure{0.40}{0.56}{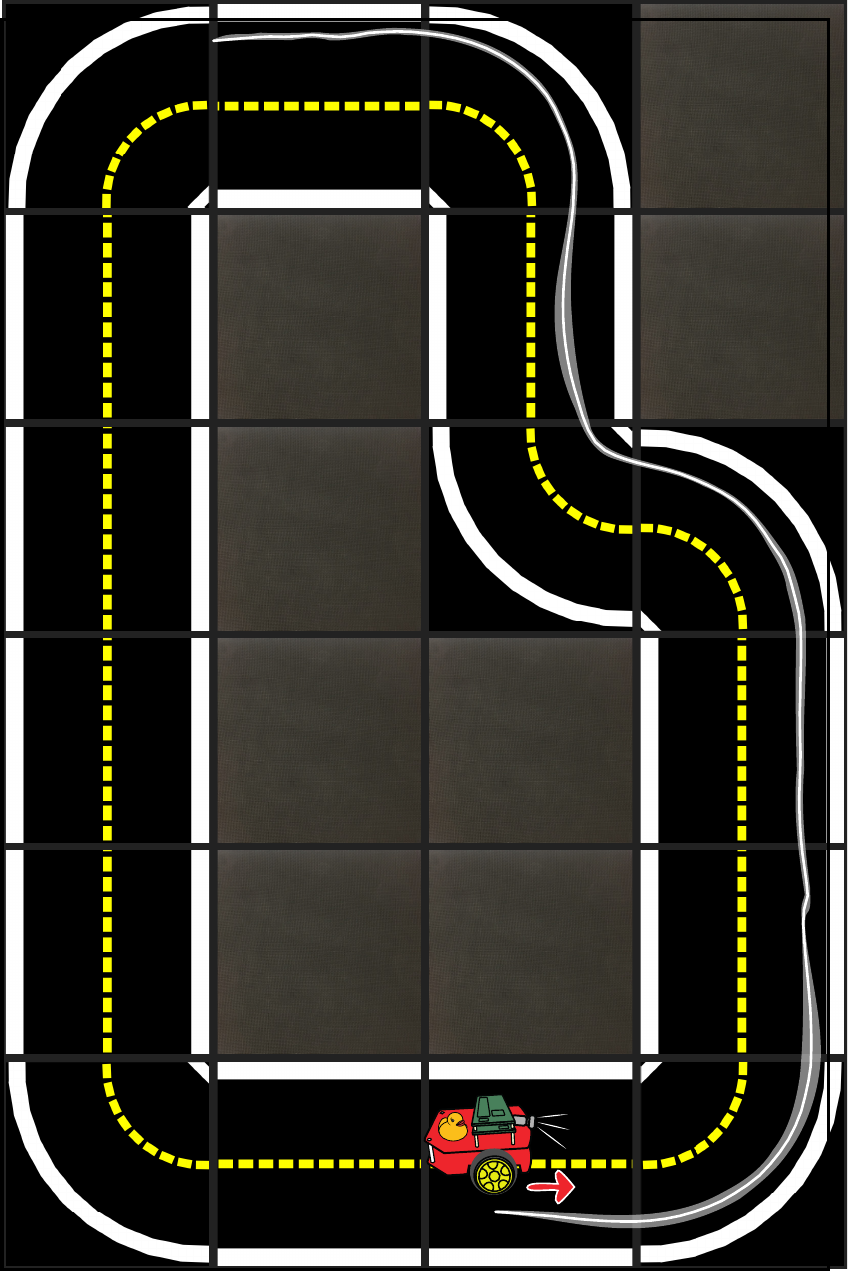}{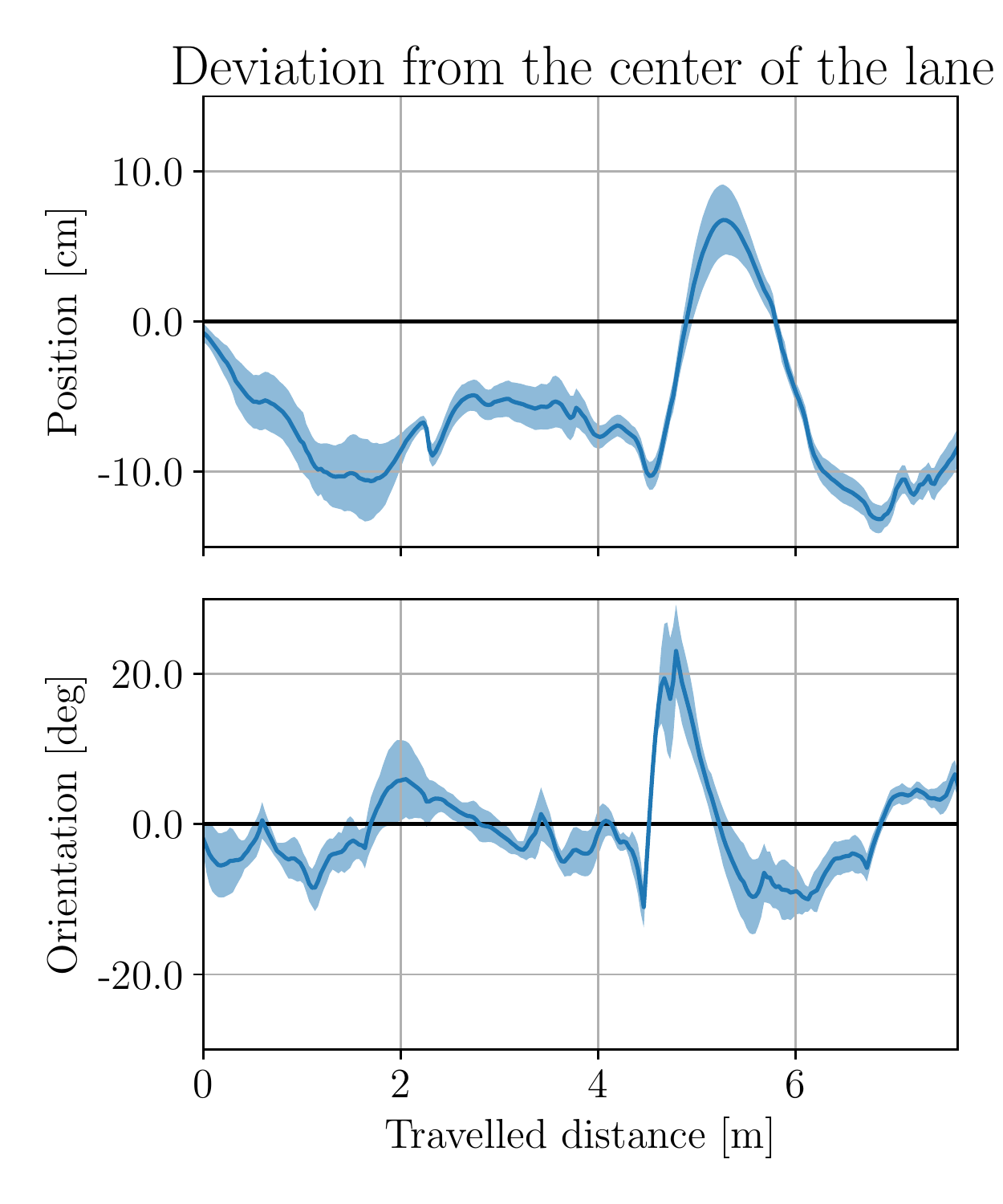}{\exptablesamebot}
    \caption{\textbf{Repeatability}: Experiment repeated on the same robot. Plots show mean and standard deviation. The higher the number of repetitions, the lower the variance.}
    \label{iros2020:fig:exp-reproducibility}
\end{figure}

To demonstrate experiment repeatability, we run a baseline agent for the \ac{LF} task nine times, on the same Duckiebot, with the same initial conditions and in the same \ac{DTA}. The results are in Figure~\ref{iros2020:fig:exp-reproducibility}, with the mean and standard deviation of the robot's position on the map. We also include plots of the \ac{MPD} and \ac{MOD} metrics. Given the vetted robustness of the baseline agent, we expect repeatable behavior. This is supported by standard deviations of the \ac{MPD} ($1.3$\,cm) and the \ac{MOD} ($2.8$\,deg), which show that there is low variability in agent performance when run on the same hardware.

\subsection{Inter-Robot Reproducibility}

\begin{figure}[t]
\centering
    \expfigure{0.40}{0.56}{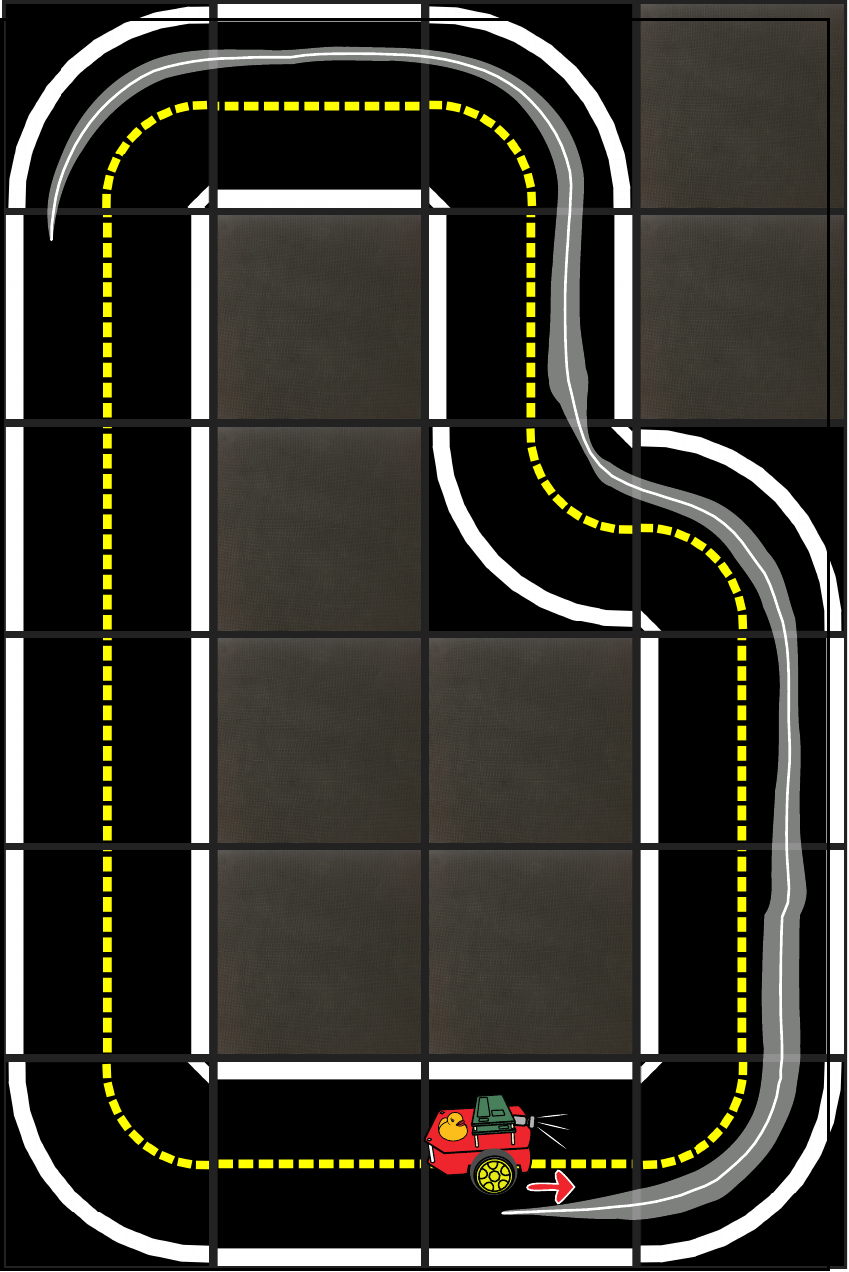}{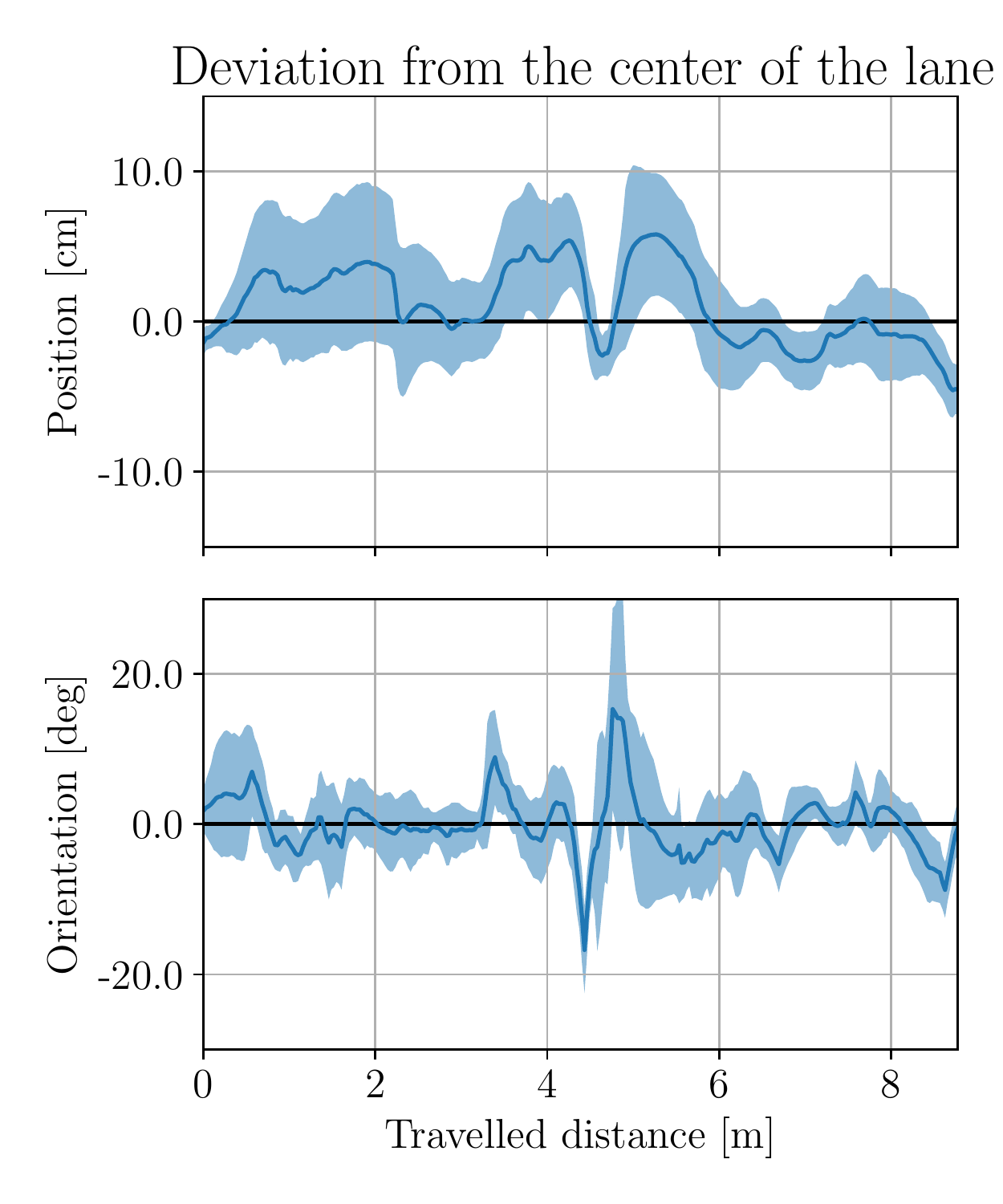}{\exptablediffbot}
        \caption{\textbf{Inter-Robot Reproducibility}: Experiments on three different robots show that there is more variation when the same agent is run on different hardware.}
        \label{iros2020:fig:Duckiebot-reproducibility}
\end{figure}

Given the low-cost hardware setup, we expect a higher degree of variability if the same agent is run on different robots. To evaluate inter-robot reproducibility, we run the \ac{LF} baseline three times each, on three different Duckiebots. Experiments are nominally identical, and performed in the same \ac{DTA}. Although the behavior of each Duckiebot is expected to be repeatable given the result of the previous section, we expect some shift in the performance distribution to hardware nuisances such as slight variations in assembly, calibration, manufacturing differences of components, etc.

Figure~\ref{iros2020:fig:Duckiebot-reproducibility} shows mean and standard deviation of the trajectories for all runs. The experiments reveal a standard deviation for the \ac{MPD} of $3.4$\,cm and a standard deviation for the \ac{MOD} of $5.2$\,deg. These results show higher deviations than for the single Duckiebot repeatability test, as expected.

\subsection{\ac{DTA} Reproducibility}

\begin{figure}[t]
\centering
    \expfigure{0.46}{0.50}{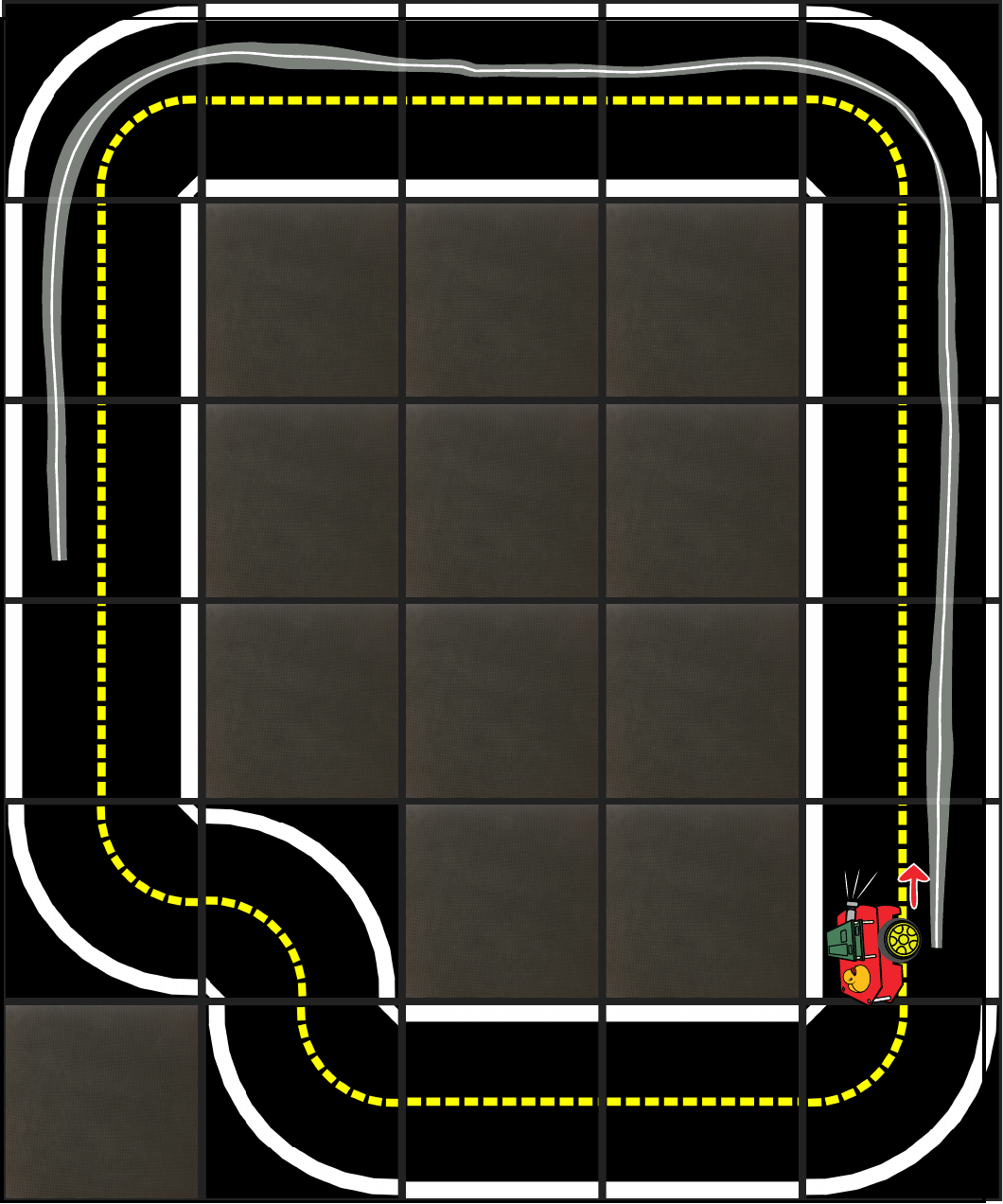}{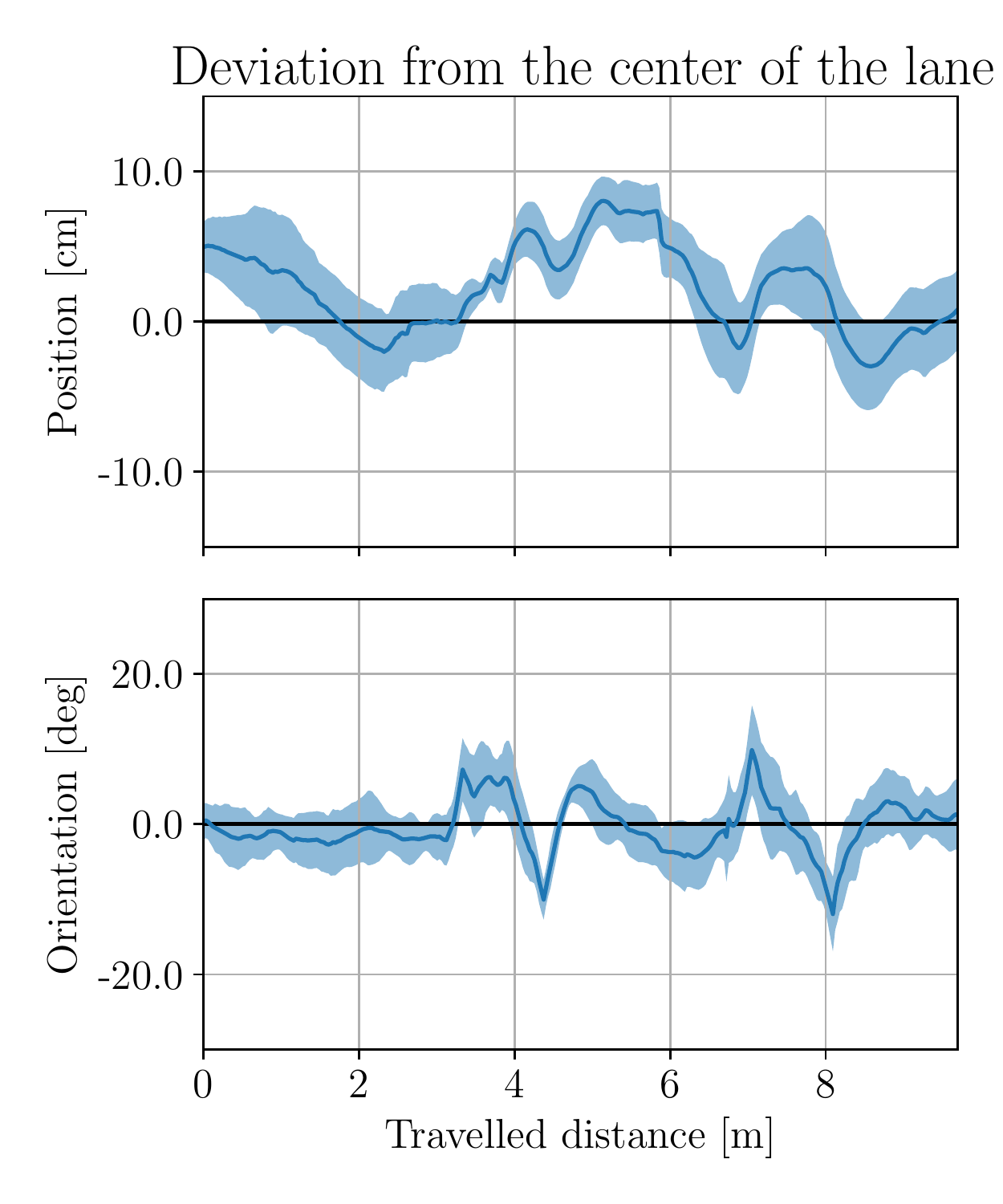}{\exptablediffsub}
    \caption{\textbf{Reproducibility}: Experiments in two different \acp{DTA}: ETH Z\"{u}rich and TTI-Chicago. Results show that the infrastructure is reproducible across setups, since experiments in two different \acp{DTA} yield similar results.}
    \label{iros2020:fig:dta-reproducibility}
\end{figure}

To demonstrate \ac{DTA} reproducibility, we run a total of twelve experiments in two different \ac{DTA}s with nominally identical conditions except for the hardware, the geographic location, the infrastructure, and the human operators. Results are shown in Figure~\ref{iros2020:fig:dta-reproducibility}. We obtain a standard deviation for the \ac{MPD} of $2.5$\,cm, and a standard deviation for the \ac{MOD} of $3.9$\,deg. Although variance is higher than the single Duckiebot repeatability test, which is to be expected, it is lower than that of the experiments run on three different robots, reinforcing the notion that hardware variability is measurable across different DTAs on the \ac{DUCKIENet}.

\newpage
\subsection{Limitations}

Finally, we discuss some limitations of the \ac{DUCKIENet} framework.

The scenarios used to evaluate the reproducibility of the platform are relatively simple. With the Duckietown setup, we are able to produce much more complex behaviors and scenarios, such as multi-robot coordination, vehicle detection and avoidance, sign and signal detection, etc. These more complex behaviors should also be benchmarked. We have also only considered here single-agent metrics, but multi-agent evaluation is also needed.

Finally, we have not analyzed the robustness to operator error (e.g., mis-configuration of the map compared to the simulation) or in case of hardware error (e.g., one camera in the localization system becomes faulty). This is necessary to encourage widespread adoption of the platform, which requires the components to be well-specified and capable of self-diagnosing configuration and hardware errors.

\section{Conclusions}
\label{iros2020:sec:conclusion}

We have presented a framework for integrated robotics system design, development, and benchmarking. We subsequently presented a realization of this framework in the form of the DUCKIENet, which comprises simulation, real robot hardware, flexible benchmark definitions, and remote evaluation. These components are swappable because they are designed to adhere to well-specified interfaces that define the abstractions.
In order to achieve this level of reproducibility, we have relied heavily on concepts and technologies including formal specifications and software containerization.
The combination of these tools with the proper abstractions and interfaces yields a procedure that can be followed by other roboticists in a straightforward manner.

Our contention is that there is a need for stronger efforts towards reproducible research for robotics, and that to achieve this we need to consider the evaluation in equal terms as the algorithms themselves. In this fashion, we can obtain reproducibility \emph{by design} through the research and development processes. Achieving this on a large-scale will contribute to a more systemic evaluation of robotics research and, in turn, increase the progress of development.

    \newpage
    \chapter{\aidosec}
    \label{aido2018:sec:main}
    This work was published in~\cite{aido2018}.\\
    \begin{figure}[!h]
\centering
    \includegraphics[width=0.95\textwidth]{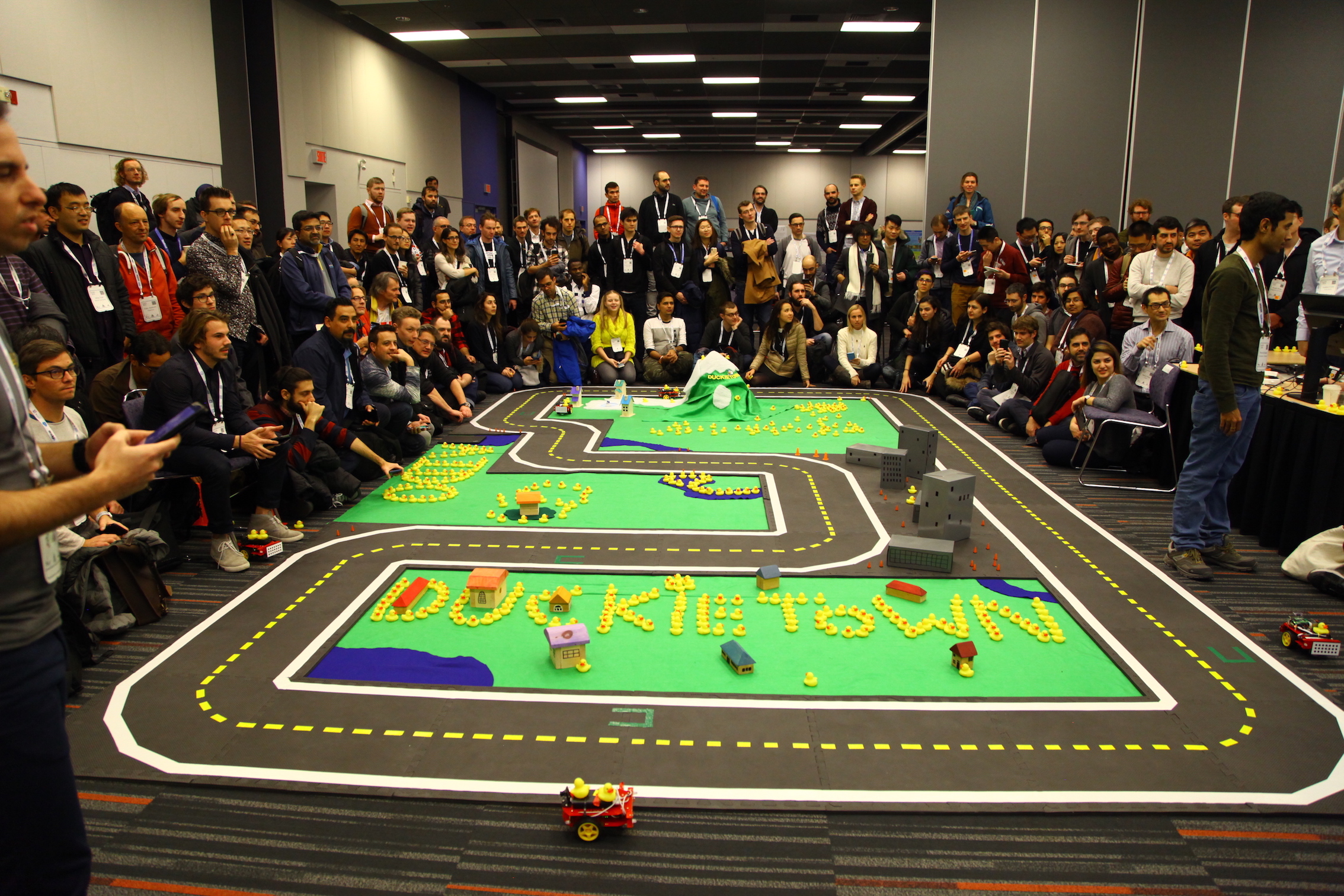}
    \caption{The AI-DO 1 competition at NeurIPS 2018, Montr\'eal, Canada.}
    \label{aido2018:fig:competition_overview}
\end{figure}

Competitions provide an effective means to advance robotics research 
by making solutions comparable and results reproducible~\cite{behnke2006robot} as without a clear benchmark of performance, progress is difficult to measure. 
The overwhelming majority of benchmarks and competitions in \ac{ML} do not involve physical robots.
Yet, the interactive embodied setting is thought to be an essential scenario to study intelligence~\cite{pfeifer2001understanding, floreano2004evolution}. 
Robots are generally built as a composition of blocks (perception, estimation, planning, control, etc.), and an understanding of the interaction between these different components is necessary to make robots achieve tasks. 
However, it is currently unclear whether these abstractions are necessary within \ac{ML} methods, or whether it is possible to learn them in an end-to-end fashion.

In order to evaluate \ac{ML} solutions on physically embodied systems, we developed a competition using the robotics platform Duckietown - a miniature town on which autonomous robots navigate~\cite{paull2017duckietown}. 
This competition is aimed as a stepping stone to understand the role of AI in robotics in general, and in self-driving cars in particular. 
We call this competition the ``AI Driving Olympics'' (AI-DO) because it comprises a set of different trials that correspond to progressively sophisticated vehicle behaviors, from the reactive task of lane-following to more complex and ``cognitive'' behaviors, such as dynamic obstacle avoidance, to finally coordinating a vehicle fleet while adhering to a set of ``rules of the road.''
The first AI-DO event (AI-DO 1) took place as a live competition at the 2018 Neural Information Processing Systems (NeurIPS) conference in Montr\'eal, Canada. AI-DO 2 will take place at the 2019 International Conference on Robotics and Automation (ICRA), again in Montr\'eal, Canada. 

Here, we describe our overall design objectives for the AI-DO (Section~\ref{aido2018:sec:purpose}). Subsequently, we detail AI-DO 1, describe the approaches taken by the top competitors (Section~\ref{aido2018:sec:results}) and then conduct a critical analysis of which aspects of AI-DO 1 worked well and which did not (Section~\ref{aido2018:sec:lessons-learned}). Section~\ref{aido2018:sec:conclusion} provides some general conclusions.
\section{The Purpose of AI-DO}
\label{aido2018:sec:vision}
\label{aido2018:sec:purpose}

\begin{table}[tb]
\centering
    \begin{tabularx}{\textwidth}{lccccccc}
    \toprule
    &
    \textbf{DARPA}  &
    \textbf{KITTI}  &
    \textbf{Robocup} &
    \textbf{RL comp.} & 
    \textbf{HuroCup} &&
    \textbf{AI-DO}\\
    &
    \cite{darpa_grand_challenge}  &
    \cite{kitti}  &
    \cite{robocup} &
    \cite{learning_to_run} & 
    \cite{hurocup} &&
    (ours)
    \\
    \midrule
    Accessibility & - & \checkmark &  \checkmark & \checkmark & \checkmark && \bf{\checkmark}   \\
    Resource constraints & \checkmark & - &  \checkmark  & - & \checkmark && \bf{$\circlearrowright$}  \\
    Modularity & - & - &  - & - & -  && $\circlearrowright$ \\
    Simulation/Realism & \checkmark & \checkmark & \checkmark & - & \checkmark && \bf{$\circlearrowright$} \\
    \ac{ML} compatibility & - & \checkmark & - &  \checkmark & - && \bf{\checkmark} \\
    Embodiment & \checkmark & - & \checkmark & - & \checkmark && \bf{\checkmark} \\
    Diverse metrics & \checkmark & - &  \checkmark & - & \checkmark && \bf{$\infty$}  \\
    Reproducible experiments & - & \checkmark & - & \checkmark & - && \bf{\checkmark} \\
    Automatic experiments & - & \checkmark & - & \checkmark & - && \bf{$\circlearrowright$} \\
    \bottomrule
    \end{tabularx}
    \caption{\textbf{Characteristics of various robotic competitions and how they compare to AI-DO} \newline
    Definitions of characteristics as they pertain to AI-DO are available in Section~\ref{aido2018:sec:considerations}.
    We identify 10 qualities we deem important to evaluate robotics algorithms and, in turn, advance the state-of-the-art in the field.
    A \checkmark signifies that a competition currently possesses a characteristic. Characteristics with  $\circlearrowright$ are in development or to be improved for AI-DO2. 
    Finally, $\infty$ symbolizes features to be added in later editions of AI-DO. With every iteration of the competition, we aim to incorporate what we believe are key characteristics of an ML-based robotics competition.}
    \label{aido2018:tab:competition_characteristics}
\end{table}

Robotics is often cited as a potential target application in \ac{ML} literature (e.g., \cite{Singh, DARLA} and many others). The vast majority of these works evaluate their results in simulation~\cite{learning_to_run} and on very simple (usually grid-like) environments~\cite{sutton2018reinforcement}. However, simulations are by design based on what we already know and lack some of the surprises the real world has to offer. 
Our experience thus far indicates that many of the implicit assumptions made in the \ac{ML} community may not be valid for real-time physically embodied systems. For instance, considerations related to resource consumption,
latency, and system engineering are rarely considered but are crucially important for fielding real robots.
Likewise, relying on a single metric to judge behavior (cf. reinforcement learning) is not realistic in more complex tasks. 
Training agents in simulation has become very popular, but it is largely unknown how to assess whether knowledge learned in simulation will transfer to a real robot.  

To some extent, \ac{ML} practitioners can be forgiven for not testing their algorithms on real robot hardware since it is time-consuming and often requires specific expertise.  
Furthermore, running experiments in an embodied setting generally limits: the amount of data that can be gathered, the ease of reproducing previous results, and the level of automation that can be expected of the experiments. To this end, we designed AI-DO as a benchmark that is both trainable efficiently %
in simulation and testable on standardized robot hardware without any robot-specific knowledge.   

What makes the AI-DO unique is that, of the many existing competitions in the field of robotics, none possess the essential characteristics that help facilitate the development of learning from data to deployment. A comparative analysis of these characteristics is given in  
Table~\ref{aido2018:tab:competition_characteristics} with the details for AI-DO given in Section~\ref{aido2018:sec:considerations}.
Ideally, the \ac{ML} community would redirect its efforts 
towards physical agents acting in the real world and help elucidate
the unique characteristics of embodied learning in the context of robotics. 
Likewise,
the robotics community should devote more effort to the use of \ac{ML} techniques where applicable.
The guaranteed impact is the establishment of a standardized baseline for comparison of learning-based and classical robotics approaches.  %

Certainly the setup for the competition is less complex than a full autonomous car. However, our objective is to provide an environment that is accessible and inexpensive, operates in the physical world, and still preserves some of the challenges of the real autonomous driving problem. Several competitions exist that use real robot logs (e.g., \cite{kitti}), but in practice many perception algorithms that do well in these contexts have actually overfit to the dataset and do not generalize well. The embodied case can mitigate this to a certain extent. Moreover, we argue that since our setup is strictly simpler than a full autonomous car, any algorithm that does not work well on our setup is guaranteed not to work on the more complex problem. 

\subsection{Important characteristics for learning-based robotics competitions}
\label{aido2018:sec:considerations}

\subsubsection{Accessibility}

No up front costs other than the option of assembling a ``Duckiebot'' are required. The Duckiebot comes with step-by-step instructions including online tutorials, a forum, and detailed descriptions within the open-source Duckiebook~\cite{Duckiebook}.

\subsubsection{Resource constraints}

Constraints on power, computation, memory, and actuator limits play a vital role within robotics.
In particular, we seek to compare contestants with access to different computational substrates such as a Raspberry PI 3 or a Movidius stick with more or less computational speed. 
Following this agenda, we want to explore which method is best given access to a fast computer, which is best using a slower computer, or which is best with more or less memory, etc.

\subsubsection{Modularity}

More often than not, robotics pipelines can be decomposed into several modules. The optimal division of a pipeline into modules however is undecided. In future editions of AI-DO we will make such modules explicit and test them according to performance and resource consumption such as computational load and memory.

\subsubsection{Simulation / Realism}

The AI-DO is an ideal setup to test sim-to-real transfer wince both simulation and real environment are readily available for testing and the software infrastructure enables toggling between the two.

\subsubsection{ML compatibility}

Logged data from hours of Duckiebot operations are available to allow for training of ML algorithms. We also provide an  OpenAI Gym~\cite{1606.01540}  interface for both the simulator and real robot, which enables easy development of reinforcement learning algorithms.

\subsubsection{Embodiment}

The AI-DO finals are run in a real robot environment (Duckietown).

\subsubsection{Diverse metrics}

Each challenge of AI-DO employs specific objective metrics to quantify success, such as:
\textit{traffic law compliance metrics} to penalize illegal behaviors (e.g., not respecting the right-of-way);
\textit{performance metrics} such as the average speed, to penalize inefficient solutions and
\textit{comfort metrics} to penalize unnatural solutions that would not be comfortable to a passenger.
These metrics are based on measurable quantities such as speed, timing and detecting traffic violations. 
Consequently,  there may be multiple winners in each competition (e.g., a very conservative solution, a more adventurous one, etc.).

\subsubsection{Reproducible experiments}

The overall objective is to design experiments that can easily be reproduced, similar to recent trends in RL~\cite{henderson2018deep}. We facilitate reproducibility by running available code on standard hardware and containerization of software.

\subsubsection{Automatic experiments}

Embodiment makes the problem of standardized evaluations particularly challenging. The ability to automatically conduct controlled experiments and perform evaluation will help to address these challenges. To that end, we are building intelligent Duckietowns called \emph{Robotariums}~\cite{pickem2017robotarium} to automate the deployment and evaluation of autonomy algorithms on standardized hardware.

\section{AI-DO 1: Competition Description}
\label{aido2018:sec:competition}

In the following, we describe the first version of the AI Driving Olympics (AI-DO). In particular we detail:
\begin{itemize}
    \item The different competition challenges and their associated scoring metrics;
    \item The physical Duckietown platform; 
    \item The software infrastructure including simulators and logs, the Docker containerization infrastructure, and various baseline implementations.
\end{itemize}

\subsection{The Challenges}
AI-DO 1 comprised three challenges with increasing order of difficulty:
\begin{enumerate}
    \item[\footnotesize{\textbf{LF}}] \textbf{Lane following} on a continuous closed course, \textbf{without obstacles}. The robot was placed on a conforming closed track (with no intersections) and was required to follow the track in the right-hand lane. 
    \item[\footnotesize{\textbf{LFV}}] \textbf{Lane following} on a continuous closed course as above, but 
	\textbf{with static obstacles (e.g., vehicles)} sharing the road. 
    \item[\footnotesize{\textbf{AMOD}}] \textbf{Autonomous mobility on demand}:\footnote{AMOD competition website \url{https://www.amodeus.science/}} Participants were required to implement a centralized dispatcher that provided goals to individual robots in a fleet of autonomous vehicles in order to best serve customers requesting rides. Low-level control of the individual vehicles was assumed. 
\end{enumerate}

\subsection{Metrics and Judging}
The rules for AI-DO 1 are described below and can be found in more detail online.\footnote{The performance \href{http://docs.duckietown.org/DT18/AIDO/out/aido_rules.html}{rules of AI-DO} \url{http://docs.duckietown.org/DT18/AIDO/out/aido_rules.html}} 

\subsubsection{Lane following with and without obstacles (LF/LFV)}
While not treated separately in AI-DO 1 for the lane following (LF) and lane following with obstacles (LFV) task, we considered the following metrics, each evaluated over five runs.

\begin{itemize}
    \item{\makebox[3.4cm][l]{\textbf{Traveled distance}} The median distance traveled along a discretized lane. Going in circles or traveling in the opposite lane does not affect the distance.}
    \item{\makebox[3.4cm][l]{\textbf{Survival time}} The median duration of an episode, which terminates when the car navigates off the road or collides with an obstacle.}
    \item{\makebox[3.4cm][l]{\textbf{Lateral deviation}} The median lateral deviation from the center line.}
    \item{\makebox[3.4cm][l]{\textbf{Major infractions}} The median of the time spent outside of the driveable zones. For example, this penalizes driving in the wrong lane.}
\end{itemize}

The winner was determined as the user with the longest traveled distance. 
In case of a tie, the above metrics acted as a tie-breaker such that a longer survival time as well as a low lateral deviation and few infractions were rewarded.

During our evaluation of the submitted entries, we noted that no submissions were performing well for LFV and therefore decided not to run any of them at the live event. We believe that this was due to the fact that avoiding moving obstacles requires a stateful representation which machine learning based methods find much more challenging. In subsequent versions of the AI-DO we will run live agents for the LFV challenge.

\subsubsection{Autonomous mobility on demand (AMOD)}

The autonomous mobility on demand challenge assumed that vehicles were already following the rules-of-the-road and were driving reasonably. 
The focus here was on deciding where and when to send each vehicle to pick up and deliver customers. 
To this end, we used three different metrics to evaluate performance.

\begin{itemize}
    \item{\makebox[3cm][l]{\textbf{Service quality}} In this arrangement, we penalized large waiting times for customers as well as large distances driven around with empty vehicles. We weighted the two components such that waiting times dominated the metric.}
    \item{\makebox[3cm][l]{\textbf{Efficiency}} Again, we penalized large waiting times for customers as well as large distances driven around with empty vehicles.  In this case, we weighted the two components such that the distance traveled by empty vehicles dominated the metric.}
    \item{\makebox[3cm][l]{\textbf{Fleet size}} This scenario considered the true/false case of whether a certain service level could be upheld with a chosen fleet size or not. The smaller the fleet the better, however only as long as service requirements were fulfilled.}
\end{itemize}

There was no live run of the AMOD challenge since it was purely in simulation. The number of entries to the AMOD challenge was significantly less than the LF and LFV challenges. Our hypothesis is that this was due to the fact that there was no embodied component to it. In future iterations of the competition we will merge the LF/LFV challenges with AMOD for a fully embodied mobility-on-demand system.

\begin{table*}[!t]
\centering
    \begin{tabularx}{0.9\linewidth}{clll}
    \toprule
    \#  &  Task &  Component & Description\\
    \midrule
    1   &   Sensing    & Fish-eye Camera      & $160^{\circ}$ FOV (front-facing), $640 \times 480$ @30Hz\\
    2   &   Computation   & Raspberry Pi 3B+    & 64bit Quad Core CPU @1.4GHz, 1 GB of RAM\\
    3   &   Actuation     & $2\times$ DC Motors & Independent, in differential drive configuration\\
    4   &   Communication & $5\times$ RGB LEDs & $3\times$ Front, $2\times$ Back for inter-bot communication \\
    5   &   Power & $1\times$ Battery & 10000 mAh battery ($>5$h operation) \\         
    \bottomrule
    \end{tabularx}
    \caption{Main components of a Duckiebot.}
    \label{aido2018:tab:duckiebot-components}
\end{table*}

\begin{figure}[!t]
    \centering
    \begin{subfigure}[t]{0.49\textwidth}
    \centering
        \includegraphics[width=0.95\textwidth]{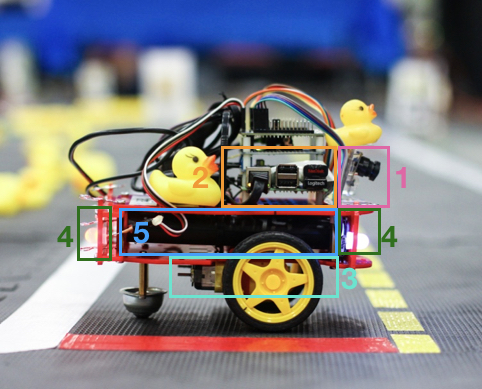}
        \caption{The main components of a Duckiebot. Numbered boxes correspond to components listed in Table~\ref{aido2018:tab:duckiebot-components}.}
        \label{aido2018:fig:duckiebot-components}
    \end{subfigure}
    ~
    \begin{subfigure}[t]{0.49\textwidth}
    \centering
        \includegraphics[width=0.95\textwidth]{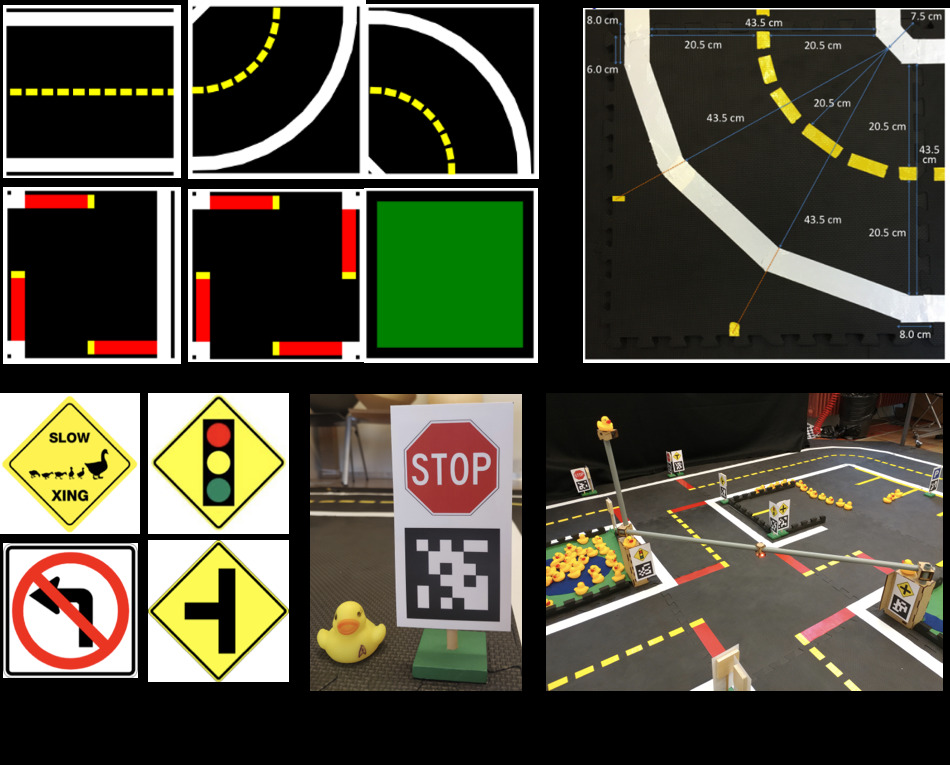}
        \caption{The Duckietown environment is rigorously defined at the road and signal level.}
        \label{aido2018:fig:duckietown-environment}
    \end{subfigure}
    
    \caption{Physical components of the competition infrastructure}
    \label{aido2018:fig:physical-components}
\end{figure}

\subsection{The Physical Competition Infrastructure}
\label{aido2018:sec:platform-physical}

The physical Duckietown platform is comprised of intelligent vehicles (\textit{Duckiebots}) and a customizable model of urban environment (\textit{Duckietown}). Both can be purchased online at low-cost to allow anyone to also physically work with the Duckietown platform.

\subsubsection{The robot - Duckiebot}

Table~\ref{aido2018:tab:duckiebot-components} lists the main components of the Duckiebot robot, which are visualized in Figure~\ref{aido2018:fig:duckiebot-components}.
We carefully chose these components in order to provide a robot capable of advanced single- and multi-agent behaviors, while maintaining an accessible cost. We refer the reader to the Duckiebook~\cite{Duckiebook} for further information about the robot hardware.

\subsubsection{The environment - Duckietown}

Duckietowns are modular, structured environments built on \textit{road} and \textit{signal} layers (Figure~\ref{aido2018:fig:duckietown-environment}) that are designed to ensure a reproducible driving experience.

A town consists of six well defined \textit{road segments}: straight, left and right $90^\circ{}$ turns, a 3-way intersection, a 4-way intersection, and an empty tile. Each segment is built on individual interlocking tiles, that can be reconfigured to customize the size and topology of a city. Specifications govern the color and size of the lane markings as well as the road geometry.

The \textit{signal layer} is comprised of street signs and traffic lights. 
In the baseline implementation, street signs are paired with AprilTags~\cite{AprilTags} to facilitate global localization and interpretation of intersection topologies by Duckiebots. 
The appearance specifications detail their size, height and location in the city. 
Traffic lights provide a centralized solution for intersection coordination, encoding signals through different frequencies of blinking LED lights.

These specifications are meant to allow for a reliable, reproducible experimental setup that would be less repeatable on regular streets. For more details about the components and the rules of a Duckietown, we refer the reader to the Duckietown operation manual~\cite{Duckiebook}.  

\subsection{The Software Infrastructure}
\label{aido2018:sec:platform-software}

\begin{figure}[!t]
    \centering
    \includegraphics[width=\textwidth]{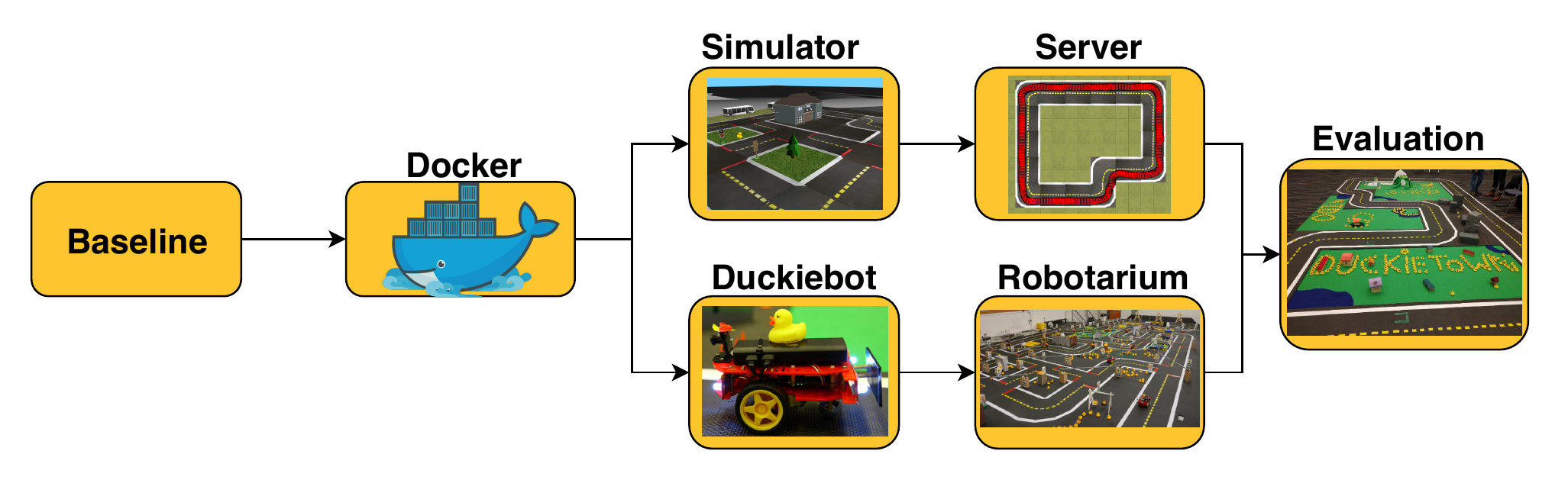}
    \caption{Illustration showing the interconnected components parts of AI-DO 1 evaluation architecture, starting from existing baseline implementations, to Docker-based deployments in simulation and on real Duckiebots, to automatic evaluations of driving behavior.}
    \label{aido2018:fig:software_overview}
\end{figure}
The overarching design principles underlying our software architecture are the following:
\begin{itemize}
    \item Minimize dependencies on specific hardware or operating systems for development, which we achieved via containerization with Docker;
    \item Standardize interfaces for ease of use, specifically to allow a participant to easily switch between developing driving solutions in simulation and deploying them on physical robots (either their own or in a Robotarium). This same interface is used to evaluate submissions;
    \item Provide interfaces to commonly used frameworks in both the ML (e.g., Tensorflow, PyTorch, and OpenAI Gym) and robotics (e.g., ROS) communities;
    \item Provide baseline implementations of algorithms commonly used in both ML (e.g., reinforcement learning, imitation learning) and robotics (see \citet{paull2017duckietown} for details) communities;
    \item Provide tools to leverage cloud resources such as GPUs.
\end{itemize}
We use Docker containerization to standardize the components as well as inter-component communication.
This has the important advantage that it allows us to deploy the same code in the simulator as well as on the real Duckiebot using an
identical interface.
Section~\ref{aido2018:sec:docker} contains more details about our different container types and the interactions between them, while Figure~\ref{aido2018:fig:software_overview} depicts the evaluation architecture.

As mentioned above, in order to facilitate fast training with modern machine learning algorithms, we developed an OpenGL-based simulator ``gym-duckietown''~\cite{gym_duckietown}. As the name suggests, the simulator provides an OpenAI gym-compatible interface~\cite{1606.01540} that enables the use of available implementations of state-of-the-art reinforcement learning algorithms. Section~\ref{aido2018:sec:simulation} discusses the simulation and its features in more detail.

Our architecture is unique in that it requires that all submissions be containerized. Rather than directly connecting to the simulator, each participant launches and connects to a Docker container that runs the simulation. In order to facilitate this interaction, we provide  boilerplate code for making submissions and for launching the simulator. We also provide baseline autonomous driving implementations based upon imitation learning, reinforcement learning, as well as a classic ROS-based lane detection pipeline. Each of these three implementations provides basic driving functionality out-of-the-box, but also includes pointers to various improvements to the source code consistent with the goal that AI-DO be educational.

Once a user submits an agent container, our cluster of evaluation servers downloads and scores the submission which is elaborated in Section~\ref{aido2018:sec:dev-pipeline}. During this process, the servers create various forms of feedback for the user. These include various numerical evaluation scores as discussed earlier (e.g., survival time, traffic infractions, etc.), plots of several runs overlaid on a map, animated GIFs that show the robot's view at any given time, and log files that record the state of the system during these runs. The evaluation server uploads this data to a leaderboard server.

We provided the code and documentation for running the submissions on the physical Duckiebot. Originally, we intended for this to happen autonomously in the Robotarium.
Since this was not completed before the first AI-DO live challenge, we instead ``manually'' ran the containers associated with the best simulation results on the physical systems.
Specifically, we ran a Docker container on our Duckiebots that
provided low-level control and perception with low latency while
exposing to the local WiFi network the same ROS interface
provided in the simulator.
From a PC connected to the same local network, we downloaded the agent
submissions (Docker images) and run it so that the agent container
was able to control the Duckiebot over the network.

\subsubsection{The Development Pipeline}
\label{aido2018:sec:dev-pipeline}

The AI-DO development process was designed with two primary goals in mind. First, it should be easy for competitors to install and run prepared ``baselines'' (Section~\ref{aido2018:subsec:baselines}) for each of the challenges. Secondly, it should be easy to test and evaluate each submission in a controlled and reproducible manner. To do so, we implemented a command line utility for building and running solutions, and a set of containerization tools (Section~\ref{aido2018:sec:docker}), enabling users to quickly fetch the necessary dependencies and evaluate their submissions locally.

For lane-following, we provided the following baseline implementations: Classic ROS, reinforcement learning (RL) with PyTorch, and imitation learning (IL) with Tensorflow. All provided baselines ran unmodified in the simulator, and some (e.g., IL) were trained using logs recorded from human drivers. We also provided docker images for producing trained RL and IL models. Following internal testing and evaluation, we released baselines to the public during the competition period.

\begin{figure}[!th]
    \centering
    \includegraphics[width=0.65\textwidth]{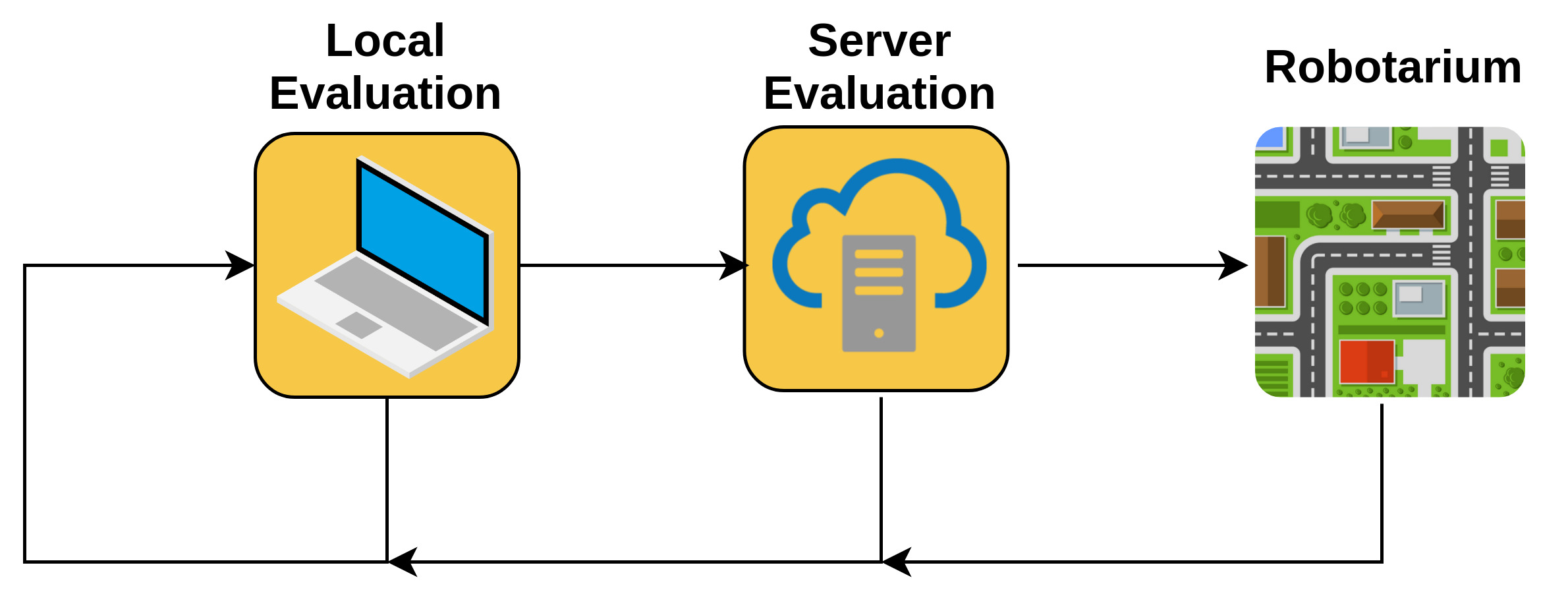}
    \caption{AI-DO submissions must pass a series of checkpoints in order to qualify for the evaluation phase.}
    \label{aido2018:fig:evaluation-checkpoints}
\end{figure}
We encouraged competitors to use the following protocol for submission of their challenge solutions (Figure~\ref{aido2018:fig:evaluation-checkpoints}). First, they should build and evaluate a baseline inside the simulator on their local development machine to ensure it was working properly. This evaluation results in a performance score and data logs that are useful for debugging purposes. If the resulting trajectory appeared satisfactory, competitors could then submit the Docker image to the official AI-DO evaluation server.
The server assigned all valid submissions a numerical score and placed them on a public leaderboard.

While the submission of competition code via containerization was straightforward, one issue we encountered during internal development was related to updates to the Docker-based infrastructure (see Section~\ref{aido2018:sec:docker}).
Originally, Docker images would be automatically rebuilt upon (1) changes to the mainline development branch and (2) changes to base layers, resulting in a cascade of changes to downstream Docker images. This was primarily due to two issues: (1) images were not version-pinned and (2) the lack of acceptance testing. Since users were automatically updated to the latest release, this would cause a significant disruption to the development workflow, where cascading failures were observed on a daily basis. To address this issue, we disabled auto builds and deployed manually, however a more complete solution would involve rolling out builds incrementally following a standard acceptance testing process. Due to a premature automation of the build pipeline, we were unable to utilize auto-builds to their full capabilities.

\subsubsection{Simulation}
\label{aido2018:sec:simulation}

\begin{figure}[!t]
    \centering
    \includegraphics[height=4cm]{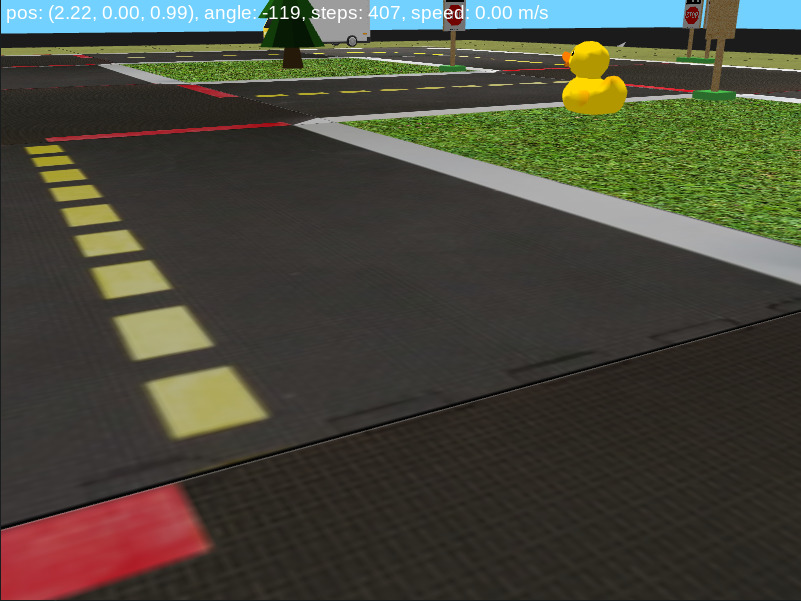}\hfil
    \includegraphics[height=4cm]{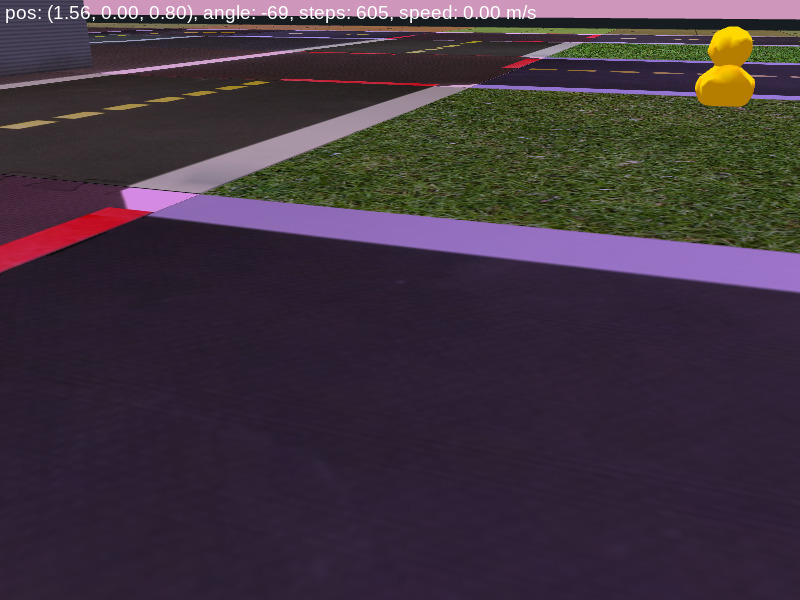}\hfil
    \includegraphics[height=4cm]{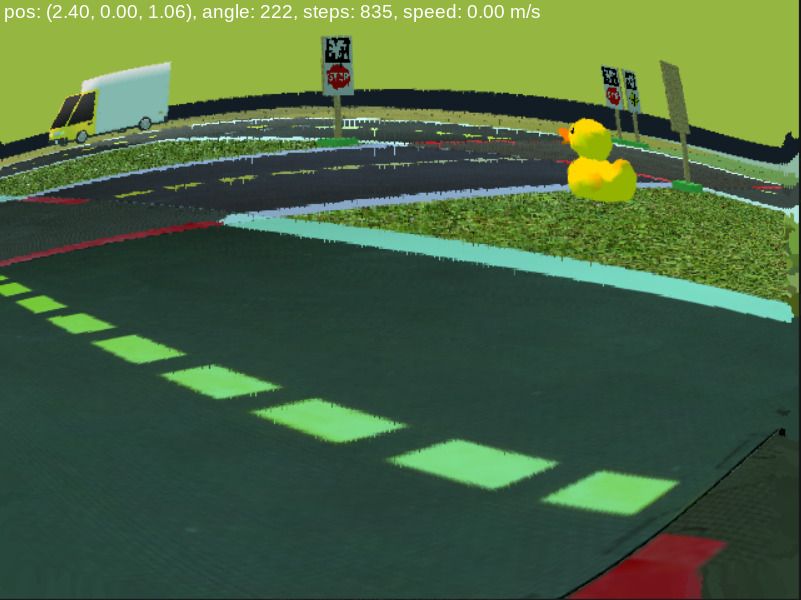}
    \caption{The lightweight OpenGL-based simulator provides (left) realistic synthetic images, (middle) visual domain randomization, and (right) lens distortion effects.}
\label{aido2018:fig:simulator}
\end{figure}

While Duckietown's software architecture allows for fast deployment on real robots, it is often easier and more efficient to test new ideas within simulated environments. Fast simulators are also important when developing and training reinforcement learning or imitation learning algorithms, which often require large amounts of data. Duckietown ships with a fast OpenGL-based simulator~\cite{gym_duckietown} (Figure~\ref{aido2018:fig:simulator}) that incorporates realistic dynamic models and simulates various Duckietown maps within a purely Python framework.  The simulator runs at hundreds of frames per second, and is fully customizable in all aspects that control it, including dynamics, environment, and visual fidelity.

The simulator provides several features that help model operation in the real Duckietown environment. These include the simulation of other agents including Duckiebots and pedestrian Duckies, as well as many of the obstacles found in physical Duckietowns, such as traffic cones and cement dividers. The simulator ships with multiple maps and a random map generator, enabling an easy way for participants to ensure that their agents do not overfit to various details of the environment. The simulator provides high-fidelity simulated camera images that model realistic fish-eye lens warping.

However, it is often the case that algorithms trained purely in simulation fall prey to the \textit{simulation-reality gap}~\cite{jakobi1995noise} and fail to transfer to the real Duckiebot. This gap is particularly problematic for methods that reason over visual observations, which often differ significantly between simulation and the real environment. An increasingly common means of bridging the simulation-to-reality gap is to employ domain randomization~\cite{Tobin2017DomainWorld}, which randomizes various aspects of the simulator, such as colors, lighting, action frequency, and various physical and dynamical parameters, in the hope of preventing learned policies from overfitting to simulation-specific details that will differ after transfer to reality.
The Duckietown simulator provides the user with hooks to dozens of parameters that control domain randomization, all with configurable ranges and settings through a JSON API.

Simulators are integral to the development process, particularly for learning-based methods that require access to large datasets. In order to ensure fairness in the development process, we also provide wrappers that allow traditional (non-learning-based) robotic architectures to benefit from the use of simulation. Participants are encouraged to try combinations of various approaches, and are able to quickly run any embodied task such as lane following (LF) and lane following with obstacles (LFV) within the simulator by changing only a few arguments. For the autonomous mobility on demand (AMOD) task, we provide a separate city-based fleet-level simulator~\cite{ruch2018amodeus}.

\subsubsection{Containerization}
\label{aido2018:sec:docker}

One of the challenges of distributed software development across heterogeneous platforms is the problem of variability. With the increasing pace of software development comes the added burden of software maintenance. As hardware and software stacks evolve, so too must source code be updated to build and run correctly. Maintaining a stable and well documented codebase can be a considerable challenge, especially in a robotics setting where contributors are frequently joining and leaving the project. Together, these challenges present significant obstacles to experimental reproducibility and scientific collaboration.

In order to address the issue of software reproducibility, we developed a set of tools and development workflows that draw on best practices in software engineering. These tools are primarily built around containerization, a widely adopted virtualization technology in the software industry. In order to lower the barrier of entry for participants and minimize variability across platforms (e.g. simulators, robotariums, Duckiebots), we provide a state-of-the-art container infrastructure based on Docker, a popular container engine. Docker allows us to construct versioned deployment artifacts that represent the entire filesystem and to manage resource constraints via a sandboxed runtime environment.

\begin{figure}[ht]
    \centering
    \includegraphics[width=0.45\textwidth]{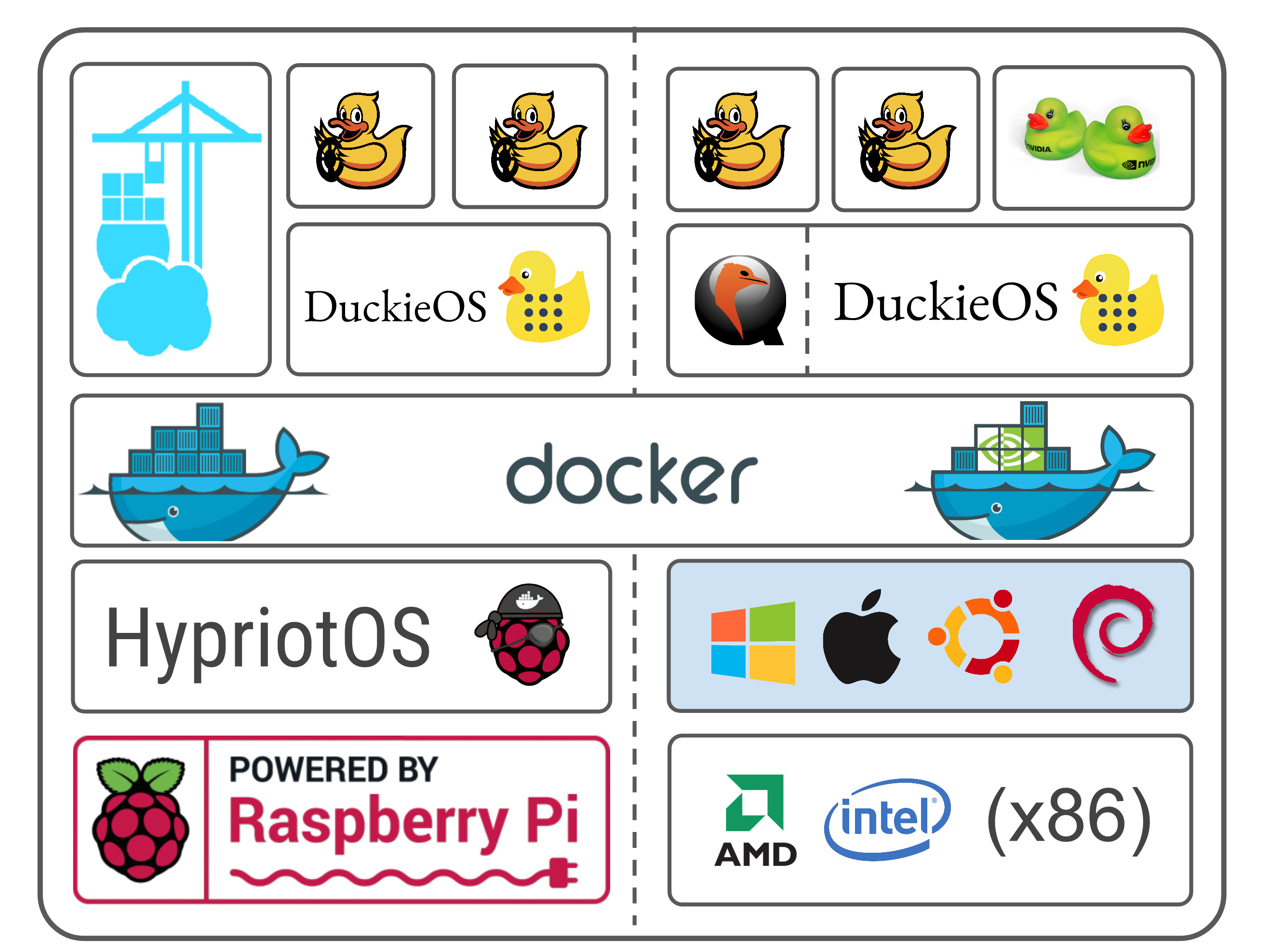} \hfil
    \includegraphics[width=0.45\textwidth]{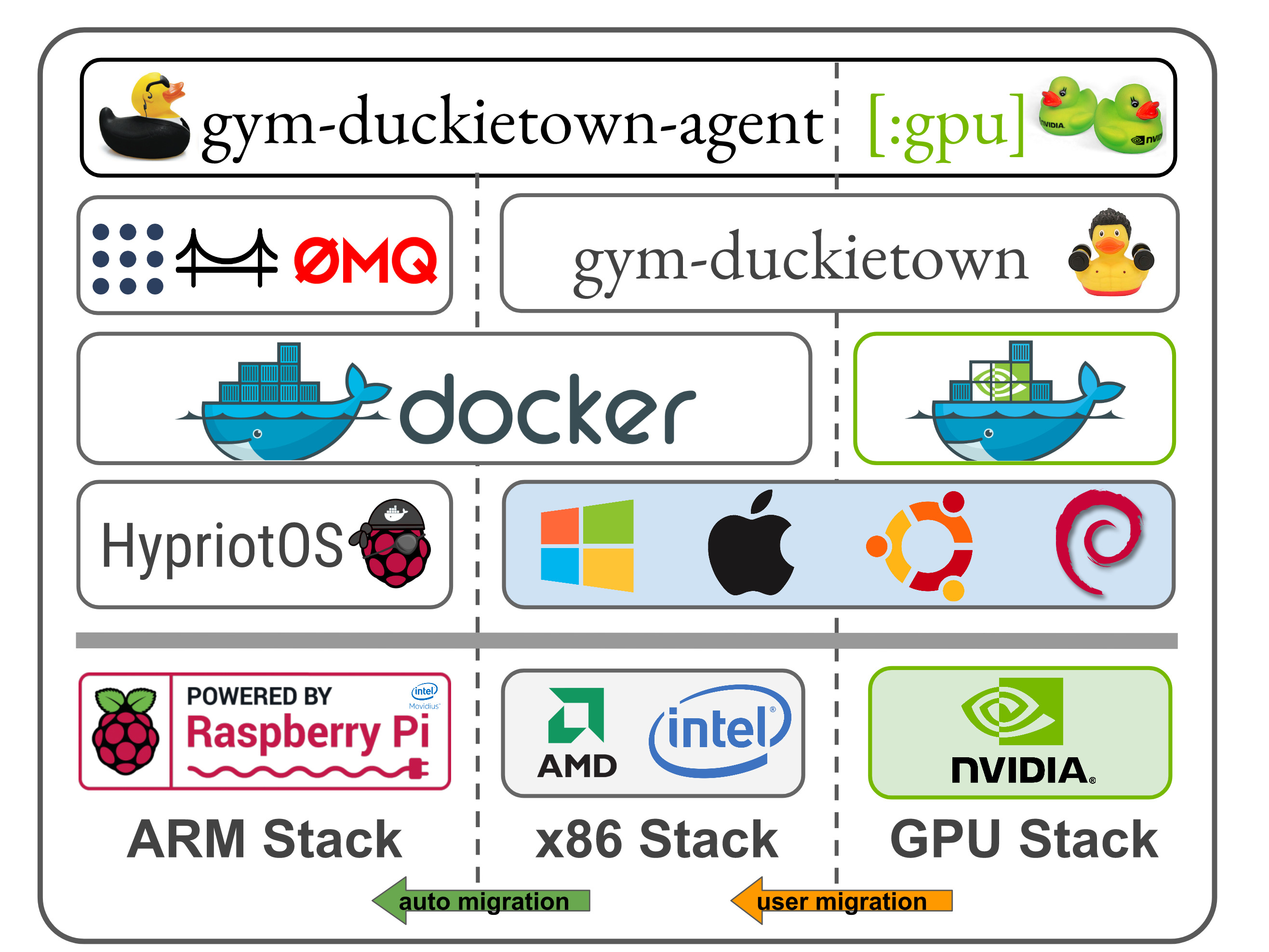}
    \caption{AI-DO container infrastructure. Left: The ROS stack targets two primary architectures, x86 and ARM. To simplify the build process, we only build ARM artifacts, and emulate ARM on x86. Right: Reinforcement learning stack. Build artifacts are typically trained on a GPU, and transferred to CPU for evaluation. Deep learning models, depending on their specific architecture, may be run on an ARM device using an Intel NCS.}
\label{aido2018:fig:docker}
\end{figure}

The Duckietown platform supports two primary instruction set architectures: x86 and ARM. To ensure the runtime compatibility of Duckietown packages, we cross-build using hardware virtualization to ensure build artifacts can be run on all target architectures. Runtime emulation of foreign artifacts is also possible, using the same technique.\footnote{For more information, this technique is described in further depth at the following URL: \url{https://www.balena.io/blog/building-arm-containers-on-any-x86-machine-even-dockerhub/}.} For performance and simplicity, we only build ARM artifacts and use emulation where necessary (e.g., on x86 devices). On ARM-native, the base operating system is HypriotOS, a lightweight Debian distribution with built-in support for Docker. For both x86 and ARM-native, Docker is the underlying container platform upon which all user applications are run.

Docker containers are sandboxed runtime environments that are portable, reproducible and version controlled. Each environment contains all the software dependencies necessary to run the packaged application(s), but remains isolated from the host OS and file system. Docker provides a mechanism to control the resources each container is permitted to access, and a separate Linux namespace for each container, isolating the network, users, and file system mounts. Unlike virtual machines, container-based virtualization like Docker only requires a lightweight kernel, and can support running many simultaneous containers with close to zero overhead. A single Raspberry Pi is capable of supporting hundreds of running containers.

While containerization considerably simplifies the process of building and deploying applications, it also introduces some additional complexity to the software development lifecycle. Docker, like most container platforms, uses a layered filesystem. This enables Docker to take an existing ``image'' and change it by installing new dependencies or modifying its functionality. Images may be based on a number of lower layers, which must periodically be updated. Care must be taken when designing the development pipeline to ensure that such updates do not silently break a subsequent layer as described earlier in Section~\ref{aido2018:sec:dev-pipeline}.

One issue encountered is the matter of whether to package source code directly inside the container, or to store it separately. If source code is stored separately, a developer can use a shared volume on the host OS for build purposes. In this case, while submissions may be reproducible, they are not easily modified or inspected. The second method is to ship code directly inside the container, where any changes to the source code will trigger a subsequent rebuild, effectively tying the sources and the build artifacts together. Including source code alongside build artifacts also has the benefit of reproducibility and diagnostics. 

\begin{figure}[!t]
    \centering
    \includegraphics[height=3.9cm]{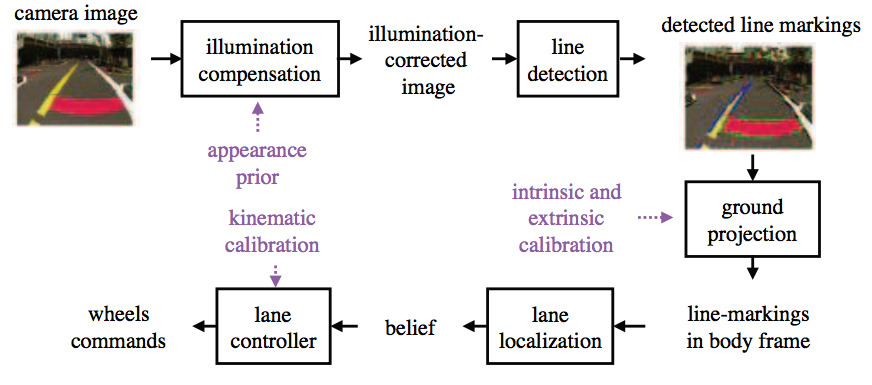} \hfil
    \includegraphics[height=3.75cm]{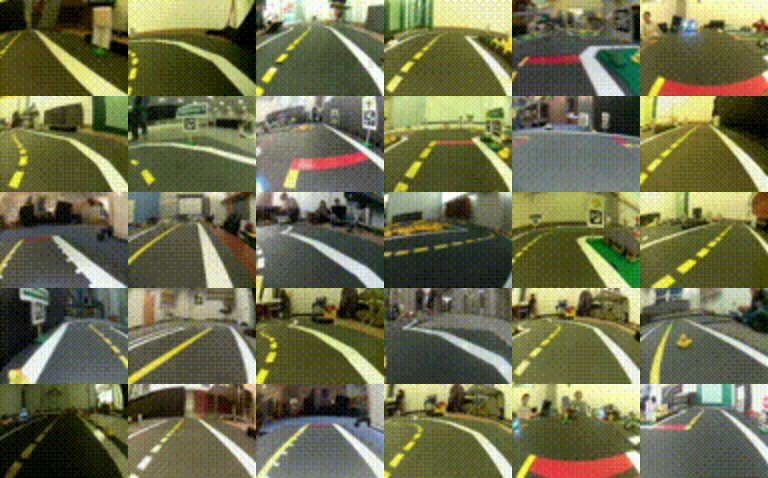}
    \caption{We provided several baselines including (left) a ROS-based Duckietown lane detection and tracking pipeline, and (right) an imitation learning baseline trained on 17 hours of logged data.}
    \label{aido2018:fig:logs}
\end{figure}

\subsubsection{Baselines}
\label{aido2018:subsec:baselines}

A important aspect of AI-DO is that participants should be able to quickly join the competition, without the significant up-front cost that comes with implementing methods from scratch. To facilitate this, we provided baseline ``strawman'' solutions, based both on ``classical" and learning-based approaches.
These solutions are fully functional (i.e., they will get a score if submitted) and contain all the necessary components, but should be easily beatable.

\begin{itemize}
    \item{\textbf{ROS baseline}
    We provided a simplified implementation of the classical Duckietown ROS-based baseline (Figure~\ref{aido2018:fig:logs} (left)), which consists of a simple lane detection algorithm that maps to wheel velocities.
    It is provided in order to show participants how to quickly use combinations of existing Duckietown software with their own enhancements.
    The ability to mix classical methods into a software pipeline enables participants to overcome various issues that often plague purely learning-based methods.
    For example, when training an end-to-end reinforcement learning policy in simulation, the policy may overfit to various visual simulation details, which upon transfer, will no longer be present. In contrast, running a learning algorithm on the top of the output of a classical module, such as a lane filter, will transfer more smoothly to the real world.
    }
    \item{\textbf{Imitation Learning from Simulation or Logs}
    Learning complex behaviors can be a daunting task if one has to start from scratch.
    Available solutions such as the above ROS-based baseline help to mitigate this.
    As one way to make use of this prior information, we are offering simple imitation learning baselines (behavioral cloning) that utilize on driving behavior datasets collected either in simulation or from recorded logs\footnote{Duckietown logs database: \url{http://logs.duckietown.org/}} as depicted on the right in Figure~\ref{aido2018:fig:logs}.
    In our experience, policies learned using imitation learning within Duckietown are frequently very skilled at driving as long as sufficient and representative data is selected to train them.
    }
    \item{\textbf{Reinforcement Learning Baseline}
    For participants who want to explore reinforcement learning approaches within AI-DO, we supply a lane following baseline using the Deep Deterministic Policy Gradients (DDPG) method~\cite{lillicrap2015continuous} trained in the Duckietown simulation environment.
    DDPG is an actor-critic method that can deal with continuous output spaces based on the Deep Q-learning architecture~\cite{mnih2015human} originally developed for discrete action spaces.
    To aid in improving the baseline, we also provide a list of possible changes that participants can try to improve the baseline.
    The distinguishing factor of such a reinforcement learning solution is that, at least in principle, it has the chance to find unexpected solutions that may not easily be found through engineering insights (ROS baseline) or copying of existing behaviors (imitation learning baseline).
    }
\end{itemize}

\section{AI-DO 1: Results}
\label{aido2018:sec:results}

The first edition of the AI Driving Olympics included an initial online phase in which submission were evaluated in simulation. 
We received our first submissions for AI-DO 1 in October 2018. Participants competed for a spot in the final competition up until the day before the final competition on 9 December at NeurIPS 2018 in Montr\'eal, Canada.

\subsection{Qualifications}
\label{aido2018:sec:results-quals}

We received a total of around 2000 submissions from around 110 unique users, with the majority focused on the lane following (LF) task. Only a small number of participants considered the lane following with obstacles (LFV) and autonomous mobility on demand (AMOD) tasks.
As a consequence, we limited the final competition to the lane following task and included the top five participants from the lane following simulation phase (see Table~\ref{aido2018:tab:results} for results).
These will be described in the following pages. 
We were interested in understanding the different approaches that the participants took and the various design decisions that they made to improve performance.
Afterwards we will share our experiences at the live finals, where we saw the code of the finalists in action on real Duckiebots.  

\begin{figure}
    \centering
    \includegraphics[width=0.33\textwidth, height=4cm, trim=0.1cm 0.1cm 0.2cm 0.1cm, clip]{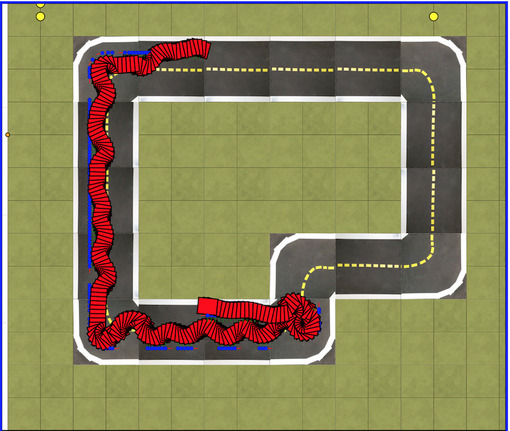}\hfil  %
    \includegraphics[width=0.33\textwidth, height=4cm]{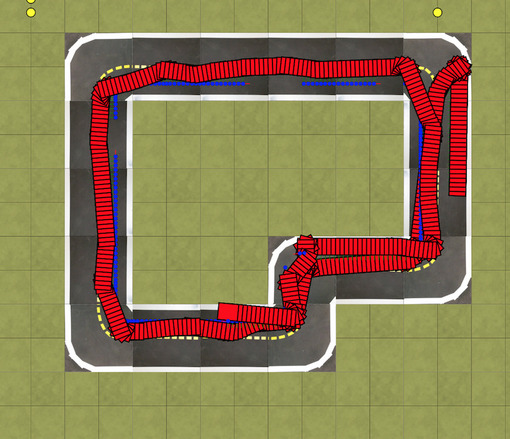}\hfil
    \includegraphics[width=0.33\textwidth, height=4cm]{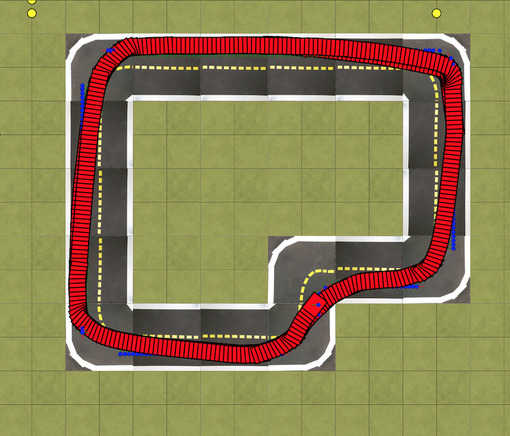} \\
    \vspace{0.5cm}
    \includegraphics[width=0.33\textwidth, height=4cm, trim=0.2cm 0.1cm 0.2cm 0.1cm, clip]{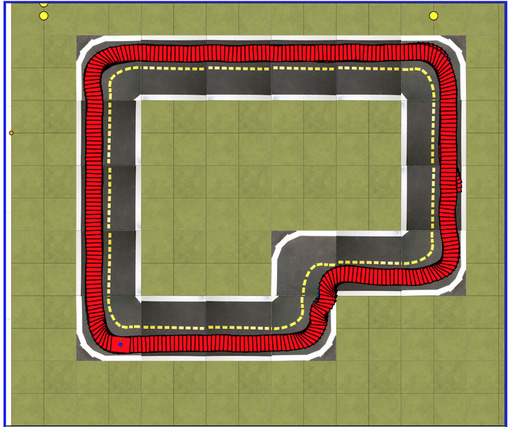}\hfil
    \includegraphics[width=0.33\textwidth, height=4cm]{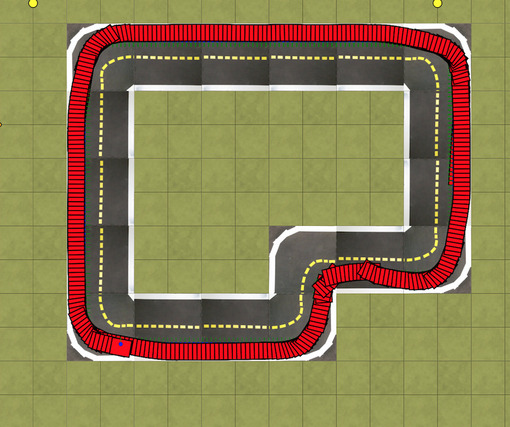}\hfil
    \includegraphics[width=0.33\textwidth, height=4cm]{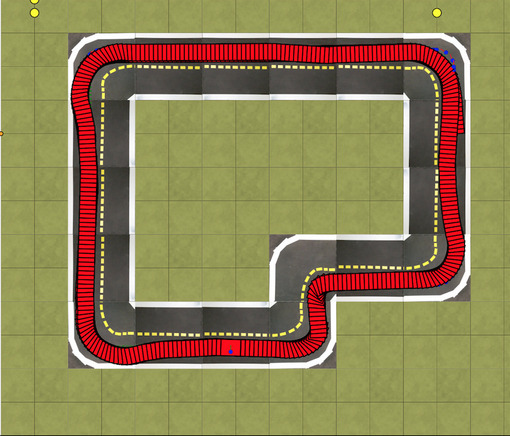}
    \caption{A bird's-eye-view of the trajectories that six different agent submissions took with performance increasing clockwise from the upper-left: a submission that did not reach the finals; that of Vincent Mai; that of Jon Plante; that of Team Jetbrains; that of Team SAIC Moscow; and the winning entry by Wei Gao.}
    \label{aido2018:fig:driving_behavior}
\end{figure}

To put the competitor submissions into context, we ask the questions of what constitutes bad driving and what good driving is?
As an example of challenges already present when driving in simulation refer to Figure~\ref{aido2018:fig:driving_behavior}.
Depicted is a bird's-eye-view visualization of the driving path of six different submissions. 
We include this figure to demonstrate the spread of behaviors between the submissions of the finalists and another submission (left-top) that did not advance to the finals.

\subsection{Finals}
\label{aido2018:sec:results-finals}

\begin{figure}[ht]
    \centering
    \includegraphics[width=0.32\textwidth, height=4cm, trim=0 0 0 2cm, clip]{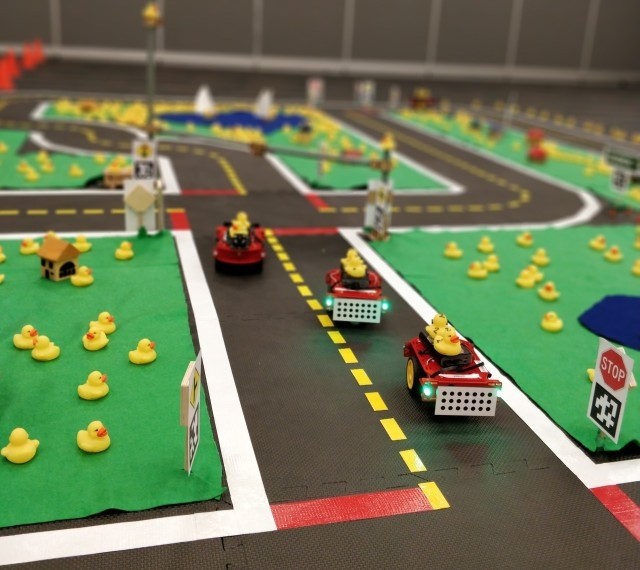}
    \includegraphics[width=0.32\textwidth, height=4cm, trim=0 0 0 0, clip]{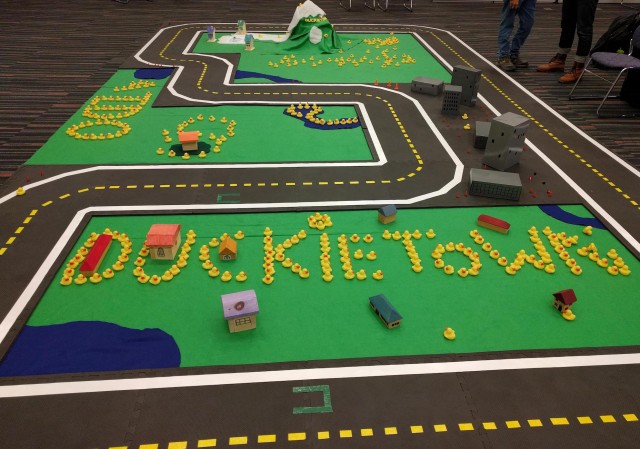}
    \includegraphics[width=0.32\textwidth, height=4cm]{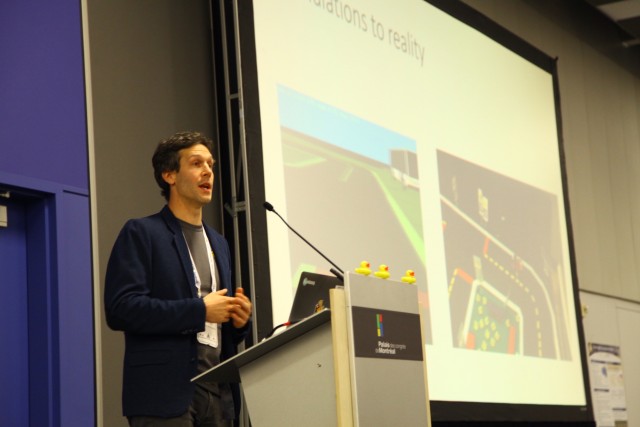}
    \caption{Impressions from the finals of AI-DO 1 at NeurIPS 2018}
\end{figure}
As the peak of the competition, the finals let the best five contestants in simulation compete on real Duckiebots at NeurIPS 2018. The
Duckiebots were placed in five starting positions spread out across the competition Duckietown. 
A judge assessed the performance of each robot in terms of the duration of time that it remained in the driveable area and the number of tiles that it traversed (Table~\ref{aido2018:tab:results}).
The best submission of each finalist was uploaded onto our competition computer, which streamed images from the Duckiebot via WiFi. 
The algorithm then computed the desired actions and sent them back to the Duckiebot.

\begin{table*}[ht]
\centering
	 {\small
	 \begin{tabular}{lccccccccc}
	   \toprule
	   \textbf{Contestant} &
	   \textbf{Simulation} &
	   \textbf{Round 1} &
  	   \textbf{Round 2} &  
 	   \textbf{Round 3} &
	   \textbf{Round 4}  &
	   \textbf{Round 5}  &
	   \textbf{Cumulative} 
      \\
	   \midrule
	   \emph{Wei Gao}          &  18 / 18.6    & 3.3 / 1   & 3.9 / 1  & 2.0 / 1   &  23.0 / 12  &  5.0 / 3     & \textbf{37 / \underline{18}}  \\
	   \emph{SAIC Moscow}      &  18 / 18.6    & 6.0 / 3   & 2.0 / 1  & 2.0 / 1   &   3.0 / 1   &  2.0 / 1     & 15 / \underline{7}            \\
	   \emph{Team JetBrains}   & 18 / 18.6     & 16.0 / 2  & 1.0 / 0  & 4.0 / 1   &   0.0 / 0   &  8.0 / 1     & 29 / \underline{4}            \\
	   \emph{Jon Plante}       & 18 / 16.7     & 18.0 / 2  & 1.0 / 0  & 7.0 / 3   &   3.0 / 1   &  5.0 / 3     & 34 / \underline{9}            \\
	   \emph{Vincent Mai}      & 18 / 14.9     & 2.0 / 1   & 1.0 / 0  & 3.0 / 2   &  14.0 / 1   &  3.0 / 2     &  23/ \underline{9}            \\
	   \bottomrule
	 \end{tabular}}
	 \caption{Results of the simulation and live competition in terms of the number of seconds before the vehicle exited the lane and the number of tiles that were traversed. Table entries of "rounds" denote the Duckiebot performance from five different starting positions. Participants were ranked in terms of distance traveled, with travel time breaking any ties with results accumulated over the five live rounds.}
	 \label{aido2018:tab:results}
\end{table*}

\subsubsection{Outcomes}
Thanks to an exceptionally long run in round four, contestant Wei Gao won the AI-DO 1 live finals. 
Most runs however did not travel more than three tiles---likely due to differences between simulation and real Duckietown, such as lighting and the cheering crowd in the background during the competition. 
Additionally, as may be observed in Figure~\ref{aido2018:fig:driving_behavior}, some of the finalist submissions appear to have overfit to the simulation environment.
We conclude that methods developed in simulation may have considerable difficulty in reaching the performance of methods developed on physical hardware. 

\subsubsection{Awards}
The finalists were awarded \$3000 worth of Amazon Web Services (AWS) credits and a trip to the nearest nuTonomy (Aptiv) location for a ride in one of their self-driving cars.

\section{AI-DO 1: Lessons Learned}
\label{aido2018:sec:lessons-learned}
Duckiebots were not the only entities that went through a learning process during AI-DO 1. 
Below we identify which aspects were beneficial and should be continued as part of subsequent competitions, and which should be revised. We separate the discussion into two categories: technical and non-technical.

\subsection{Technical}
The following reflects upon what we learned making AI-DO 1 from a technical point of view. 

\subsubsection{Successes}

\paragraph{Software infrastructure}  The use of Docker allowed us to run submission code on many different computational platforms and operating systems.
A central and beneficial feature of Docker is that any submission is reproducible, it can still be downloaded, tried and tested today. 
Likewise, it was helpful that all submissions were evaluated on a server yet could also be run locally. This allowed participants to rapidly prototype solutions locally in an environment that was an exact functional match of the evaluation environment. As a result, competitors could do many local evaluations and only submit an entry when they were confident that they would surpass previous results, which alleviated stress on our server infrastructure. 

\paragraph{Simulation}  The custom simulator was used for initial benchmarking. Since it was developed in lightweight Python/OpenGL, it enabled rapid training (up to 1k frames/sec) for quick  prototyping of ML-based solutions.
Online evaluations and comprehensive visualizations\footnote{Submission are visualized on \url{https://challenges.duckietown.org/v4/} by clicking the submission number.}  were especially helpful in assessing performance and better understanding behavior.
The fact that the Duckietown environment is so structured and well-specified, yet simple, allowed for easy generation of conforming map configurations in simulation with relatively little effort. 

\paragraph{Baselines and templates}  The containerized baselines and templates provided entry points that made it easy for competitors from different communities to begin working with the platform, essentially with just a few lines of code, as described in Section~\ref{aido2018:subsec:baselines}. These baselines made the competition accessible to roboticists and the ML communities alike, by providing interfaces to the standard tools of each.
These learning resources were reinforced further through Facebook live events and a question forum that explained the AI-DO software in more detail.

\subsubsection{To Be Improved}

\paragraph{Software infrastructure}  A large amount of effort was dedicated to the use of containerization via Docker. While containerization provides several benefits including reproducibility, the Docker infrastructure still poses challenges. For example,  %
Docker changed versions in the middle of the competition wreaking temporary havoc to our dependencies. 

An awareness of, and strategy for, dealing with resource constraints are essential to robotic systems. However, these constraints still have to be implemented using Docker. 
This gave an advantage to approaches that were computationally more expensive, when the inverse was desired. For example, a submission that ran on a Raspberry Pi should have been preferred over a submission that ran off-board on an expensive GPU. 

\paragraph{Submission evaluation}  All submissions were evaluated on one local server, which made the submission infrastructure fragile. There were instances when participants were unable to make submissions because the server was down or we had too many submissions entered at the same time. In the future, we will aim to move evaluations to the cloud to offset these issues.

Enhanced debugging and evaluation tools would have made it considerably easier to diagnose issues during the local evaluation phase. Users could be shown a richer set of diagnostics and logs from the simulator directly after their submission was evaluated, or from recorded logs of an on-Duckiebot trial. 

All submissions were publicly visible. Competitors were able to download, view, and re-submit other images using the public hash code. While we are not aware of any participants accessing another team's submissions during the competition, we intend to change the submission infrastructure to make the process more secure.

Furthermore, evaluations were not statistically significant since the simulations considered 20 episodes in only 1 map. Subsequent competitions will include more maps as well as more evaluation episodes and variability in the simulated environments. 
Bringing submissions tested in simulation to testing on real robots is a major component of AI-DO. While we provided ample opportunity for testing in simulation in AI-DO 1, there was insufficient opportunity for competitors to test on the hardware, which was apparent in the quality of the runs in the live event. 

\paragraph{Simulation}  While the finals of AI-DO 1 proved to be a good demonstration that the simulator-reality gap is large, we did not have a metric for quantifying it.
The Duckietown codebase has had a working lane following solution since 2016 \cite{paull2017duckietown}, so we know that a robust solution to the LF challenge was feasible. 
However, in simulation this robust solution did not perform well. In contrast, solutions that did perform well in simulation did not work on the real hardware. Since most competitors did the majority of their development in the simulation environment, this resulted in sub-optimal performance at the live event. In our analysis since the AI-DO 1 event we have concluded that this was primarily attribuatable to two causes: 1) Unmodeled and incorrect dynamics and extrinsic calibration - there were elements of the simplified dynamics model in the simulator that introduced biases that were difficult to overcome when the agent was transferred to the real hardware, and 2) The lack of a well-documented domain randomization \cite{Tobin2017DomainWorld} API - by randomizing over unknown parameters it allows the agents to become more robust to the reality gap.

We \textit{must} provide better tools for competitors to be able to bridge the reality gap more easily. These tools should include: better API to modify and/or randomize parameters in the simulator, easier access to real robot logs, and easier access to real robot infrastructure either remotely (through \acp{DTA}) or locally.

\paragraph{Baselines and templates}  Duplication in the boilerplate code resulted in diversity in the interface among various templates and baselines. We plan to reduce this duplication to make it easier to change protocols between different challenges.

\paragraph{Logs}  Logs represent valuable data from which to train algorithms, yet we are unaware of any competitor that made use of the $\sim$16 hours of logs that were made available. Our conclusion was that the log database was not sufficiently advertised and that the workflow for using the logs in a learning context was not explicit.

\subsection{Non-Technical}

The following discusses the non-technical aspects of AI-DO 1. 

\subsubsection{Successes}

\paragraph{Competition logistics}  The Duckietown platform~\cite{paull2017duckietown} has proven to be an ideal setup for benchmarking algorithms as it is simple, highly structured and formally specified, yet modular and easily customised. It is also accessible to competitors as the robot and city hardware is readily available and inexpensive.

\paragraph{Available resources}  The Duckietown platform contains a rich assortment of learning resources in the Duckiebook~\cite{Duckiebook} and on the Duckietown website~\cite{duckietown_website}, as well as a large body of open-source software for running Duckiebots within Duckietown. The wealth of existing resources reduced the technical support demand on the organizers.

\begin{figure}[tb]
    \centering
    \includegraphics[width=0.49\textwidth, height=5cm]{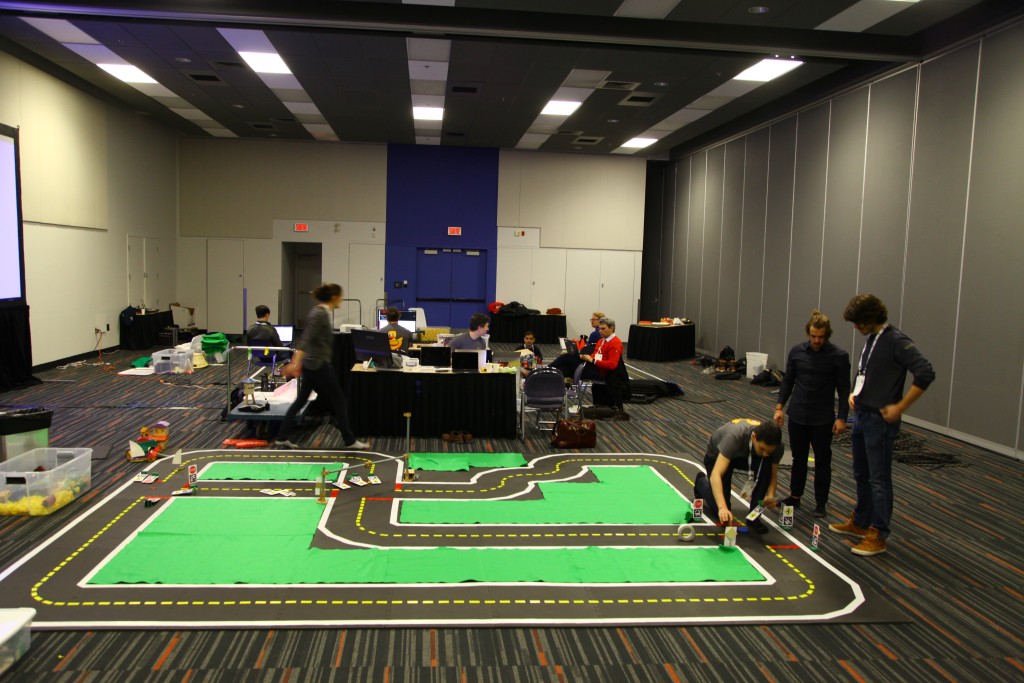}
    \includegraphics[width=0.49\textwidth, height=5cm]{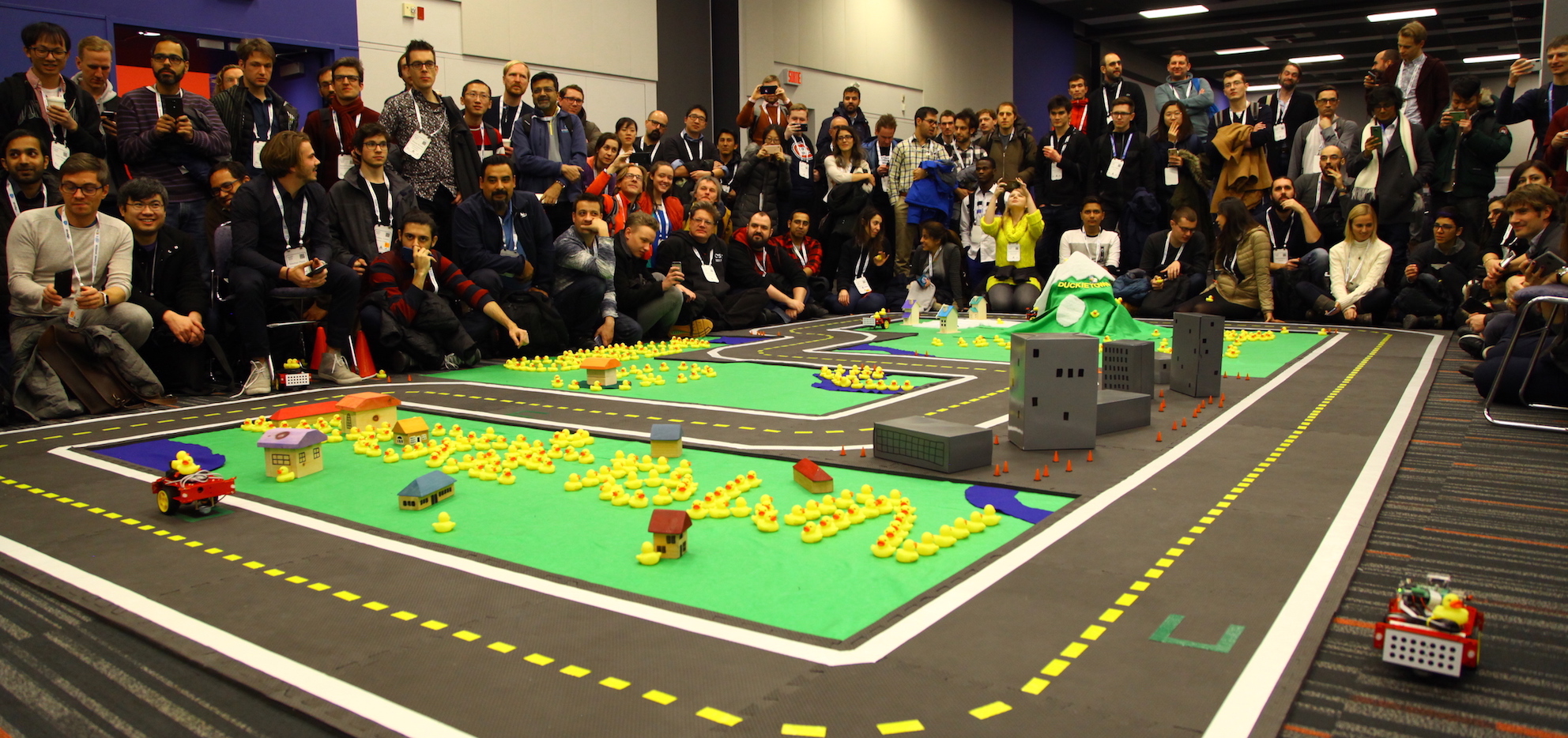}
    \caption{The AI-DO 1 event. \textbf{Left:} The setup required an entire day and a small army of volunteers. \textbf{Right:} Audience at the finals of AI-DO 1 at NeurIPS 2018 in Montr\'eal, Canada. For applause, the public squeezed rubber duckies instead of clapping.
    }
    \label{aido2018:fig:public_involvement}
\end{figure}

\paragraph{Engagement}  Especially positive was the public's emotional reception at the AI-DO 1 finals, as shown in Figure~\ref{aido2018:fig:public_involvement}-right.
Both the Duckie-theme and competition were well-received. Instead of clapping between rounds the audience squeezed yellow rubber duckies.

\subsubsection{To Be Improved}

Similar to many larger projects, AI-DO 1 fell victim to the \emph{planning fallacy}~\cite{planning_fallacy} wherein we were more optimistic in our timeline than our general experience should have suggested. 

\paragraph{Competition logistics}  The amount of setup required for the 1.5 hour event was significant. We required access to the competition space for the entire day  and needed to move a significant amount of materials into the competition venue (see Figure~\ref{aido2018:fig:public_involvement}-left). Making this process smoother and simpler would be an important step towards making the competition easily repeatable.

\begin{figure}[tb]
    \centering
    \includegraphics[width=\textwidth, trim=0.2cm 0 0.2cm 0.2cm, clip]{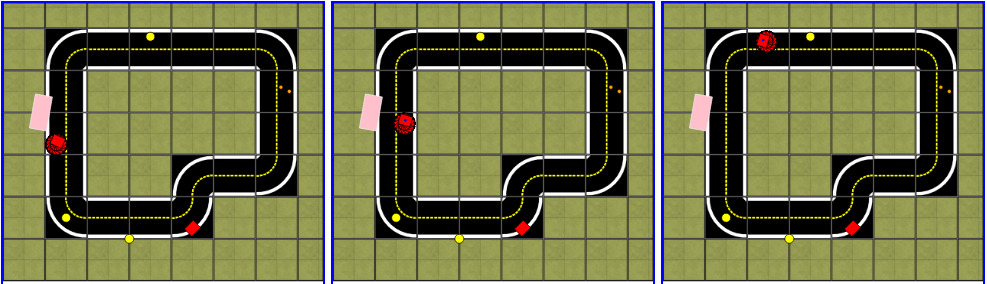}
    \caption{Submission to Challenge LFV by user ``sodobeta''. Example of degenerate solution found by reinforcement learning as a result of poorly specified evaluation metrics. Submission originally got a high score for ``Traveled Distance'' even though it is just spinning in a circle. 
    }
    \label{aido2018:fig:spinning}
\end{figure}

\paragraph{Rules}  As the competition infrastructure and our understanding of the tasks evolved, the rules evaluating competitors changed too frequently and too close to the end of the competition. This may have made it harder to participate easily and improve consistently.
It was surprisingly difficult to define the evaluation metrics in a way that promoted the kind of behavior we sought. Reinforcement learning algorithms are well-known for finding these types of loopholes. As one example, the evaluation of the ``Traveled Distance'' did not consider progress down the lane but only linear speed at first. As a result, it was possible to get a high score by spinning in circles as fast as possible (Figure~\ref{aido2018:fig:spinning}).

\section{Subsequent editions of the AI-DO}
\label{aido2018:sec:instances}

At the time of writing, the AI-DO is at its 6th edition.\\

The \textbf{second edition} of AI-DO took place at the 2019 International Conference on Robotics and Automation (ICRA), with finals held in Montréal, Canada, in May 2019.
AI-DO 2 comprised again only the Urban league, but additional challenges were added, such as Lane following with other vehicles (LFV) and Lane following with other Vehicles and Intersections (LFVI).\\

The \textbf{third edition} was held at NeurIPS 2019 with finals held in Vancouver, Canada. In AI-DO 3 we introduced the advanced perception and racing leagues and received more than $2$ thousand submissions across all of the leagues.\\

The \textbf{fourth edition} was scheduled for ICRA 2020 in Paris, France, but was unfortunately canceled due to the COVID-19 pandemic.\\

The \textbf{fifth edition} of AI-DO was organized in conjunction with NeurIPS 2020. Due to the COVID-19 pandemic, the conference and finals were held virtually. AI-DO 5 was organized in two leagues: Urban Driving and Advanced Perception, with novel challenges added to both of them.\\

The \textbf{sixth edition} of AI-DO was organized in conjunction with NeurIPS 2021. It featured three leagues: Urban Driving, Advanced Perception, and Racing. A new challenge was included in the Urban Driving league, and the new robot DB21J (see Section~\ref{mooc2021:sec:db21j}) was released to support larger neural models. 

\section{Conclusion}
\label{aido2018:sec:conclusion}

We have presented an overview of the AI Driving Olympics. The competition seeks to establish a benchmark for evaluating machine learning and ``classical'' algorithms on real physical robots. The competition leverages the Duckietown platform: a low-cost and accessible environment for autonomous driving research and education. 
We have described the software infrastructure, tools, and baseline algorithm implementations that make it easy for robotics engineers and machine learning experts alike to compete. 

For science to advance we need a reproducible test-bed to understand which approaches work and what their trade-offs are. 
We hope that the AI Driving Olympics can serve as a step in this direction in the domain of self-driving robotics. 
Above all, we want to highlight the unique design trade-offs of embodied intelligence due to resource constraints, non-scalar cost functions,
and reproducible robotic algorithm testing.

    \newpage
    \chapter{\sharcsec}
    \label{sharc2021:sec:main}
    \begin{figure}[h]
\centering
    \includegraphics[width=\textwidth, trim=0.5cm 0 1.2cm 2.5cm, clip ]{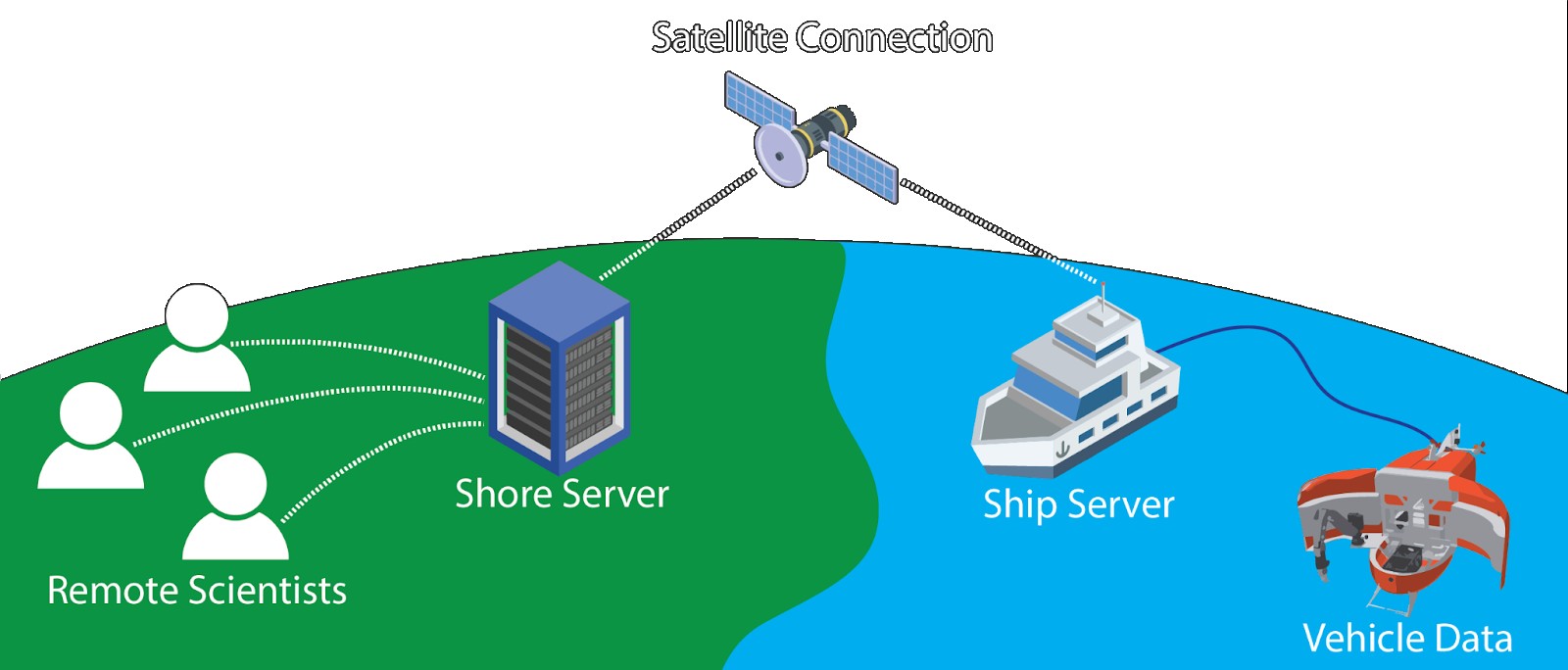}
    \caption{\textbf{SHARC}: SHared Autonomy for Remote Collaboration framework}
    \label{sharc2021:fig:components}
\end{figure}

Robotic manipulation plays a crucial role in underwater manipulation and exploration.
Divers are capable of completing precision tasks requiring a high degree of dexterity, such as taking measurements with delicate handheld instrumentation.
However, they are physiologically limited by depth, and thus cannot operate outside of coastal regions. Divers are also limited by operation time, as they can only spend up to eight hours underwater at a time. 

\acfp{ROV} are able to bring dexterity to deeper parts of the ocean and can stay underwater for days at a time, however, operating them is expensive due to the 
need for a support vessel and on-board crew.
While \acfp{AUV} reduce the operation costs by being autonomous, conducting fully 
autonomous underwater intervention tasks remains an open challenge.
In fact, 3D scene understanding in unstructured environments, optimal sample site selection, and precise control of hydraulic manipulators are some of the open problems preventing full-scale adoption of fully autonomous underwater interventions.

Fully autonomous underwater manipulation has so far only been achieved in testing tanks and structured environments. While the long-term goal is fully autonomous manipulation in unstructured environments, short-term field operations can benefit from recent advancements in the field of \textbf{shared autonomy}.
In fact, shared autonomy allows us to leverage human capabilities in perception and semantic understanding of an unstructured environment, while relying on well-tested robotic capabilities, such as planning trajectories around a perceived obstacle.

In this work, we propose SHARC: SHared Autonomy for Remote Collaboration framework.
SHARC allows multiple remote scientists to efficiently plan and execute high-level sampling procedures using an underwater manipulator while deferring low-level control to the robot.
SHARC consists of four parts: on-shore scientists, a shore server, a ship server, and a robot (see Figure~\ref{sharc2021:fig:components}).
The data from the robot is streamed up to the ship server via a fiber optic cable. The pilot onboard the ship can choose which data streams to forward from the ship server onto the shore server, usually via satellite connection.
The shore server relays the incoming data to the on-shore scientists that are 
connected to it.
The stream of data is by default unidirectional, and goes from the robot to the scientists.
The pilot on-board the ship has the ability to temporarily allow bidirectional
stream of data for a specific scientist, allowing them to contribute to the mission by providing input. In doing so, the pilot
orchestrates the assignment of smaller parts of a more complex manipulation 
task to the on-shore scientists, effectively allowing scientists with different
fields of expertise to collaborate in performing the mission.

When a scientist is allowed the temporary ability to
provide inputs to the task at hand, it does not mean that the pilot has handed 
full-control of the robot to the scientist. In fact, the pilot receives the
input from the scientists (e.g., a request to move the manipulator to a specific
point of the seafloor), validates it, and only then forwards it to the robot for
execution.

\section{XRF sampling of seafloor sediments}

\begin{figure}[!t]
\centering
    \includegraphics[width=\textwidth]{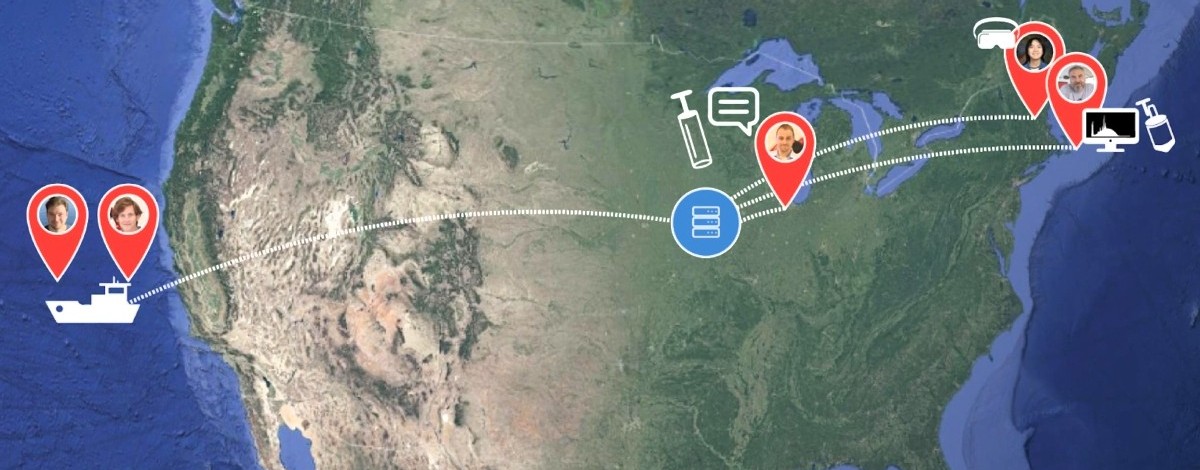}
    \caption{Geographically distributed team of scientists during the OECI expedition in September 2021}
    \label{sharc2021:fig:map-expedition}
\end{figure}
\begin{figure}[!t]
    \centering
    \begin{subfigure}[t]{0.49\textwidth}
    \centering
         \includegraphics[width=\textwidth, trim=4cm 0 1cm 0, clip]{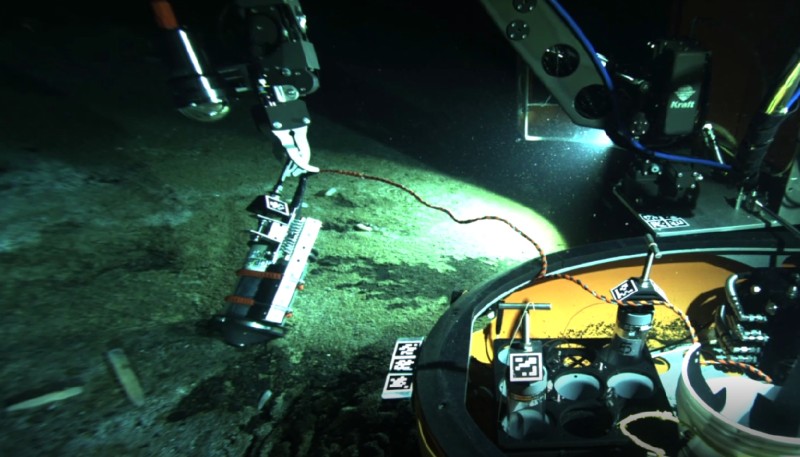}
         \caption{Robot placing the XRF tool on the seafloor}
         \label{sharc2021:fig:xrf-sampling}
    \end{subfigure}
    ~
    \begin{subfigure}[t]{0.49\textwidth}
    \centering
        \includegraphics[width=\textwidth, trim=0.5cm 0.5cm 0.5cm 0.5cm, clip]{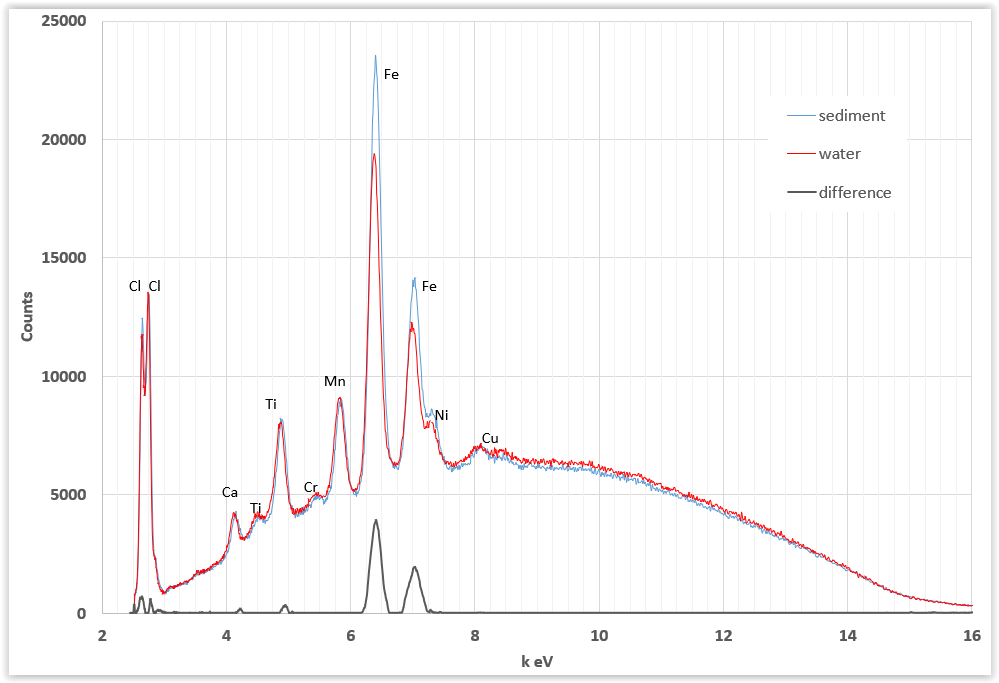}
        \caption{Real-time XRF data received by the scientist}
        \label{sharc2021:fig:xrf-data}
    \end{subfigure}

    \caption{XRF sampling procedure executed using SHARC}
    \label{sharc2021:fig:xrf-procedure}
\end{figure}

\ac{XRF}, is commonly used for elemental analysis of sediments and it is particularly suited for detecting the presence of heavy metals. 
This technique is usually preferred to others because of its non-destructive 
quality and minimal need for sample preparation.
Despite its utility, this analysis is largely conducted ex-situ on samples brought back to labs. 

In-situ sampling would enable scientists to have immediate feedback on the composition of sediments, and allow for real-time, iterative exploration of the environment. However, taking XRF samples underwater is challenging as the high attenuation rate of xrays in the water requires the delicate sensor to be in direct contact with sediment during measurement.

Previous work attempted in-situ XRF sampling.
\citet{1154232} lowered an XRF off the stern of a ship,
\citet{Breen2012AnalysisOH} took XRF measurements by landing an AUV on patches of sediments.
However, these approaches have a coarse spatial resolution, with a precision on the order of 10s of meters. 
Robotic manipulators on-board \acp{ROV} can enable site selection with a spatial resolution on the order of centimeters. Unfortunately, this approach requires high degrees of dexterity typically only achievable by a trained \ac{ROV} pilot on-board a support vessel with high-bandwidth connection to the 
robot.

\section{XRF sampling using SHARC}

\begin{figure}[t]
\centering
    \includegraphics[width=\columnwidth]{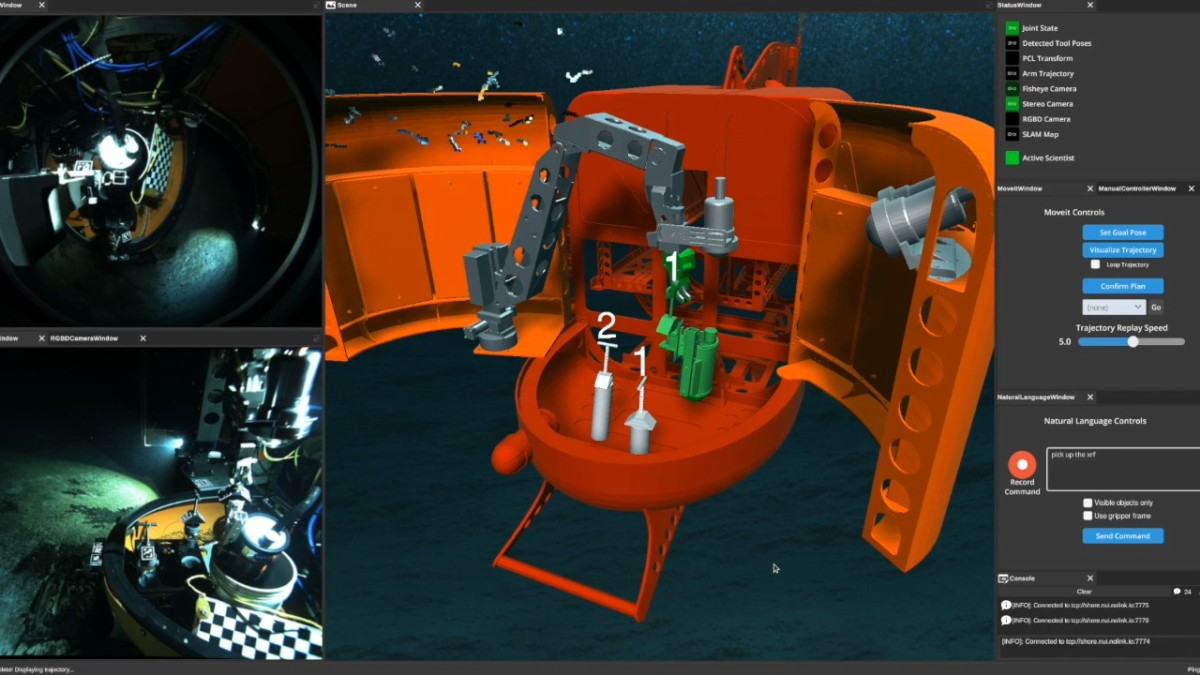}
    \caption{SHARC multimodal interface}
    \label{sharc2021:fig:interface}
\end{figure}

In September 2021, we deployed SHARC during the ``OECI Technology Demonstration: Nereid Under Ice (NUI) Vehicle + Mesobot'' expedition off the coast of San Pedro Basin, USA, aboard the E/V Nautilus support ship.
A team of five scientists, two on-board the support vessel and three on-shore, successfully employed SHARC to collect an in-situ \ac{XRF} measurement about 
$1000$ meters below the surface (see Figure~\ref{sharc2021:fig:xrf-procedure}). 
While the vessel was operating off the west coast of the US, the on-shore team 
of scientists was geographically distributed on the central and eastern part of the US at more than $3000$Km from the ship (see Figure~\ref{sharc2021:fig:map-expedition}).

\newpage
SHARC provides a multi-modal user interface for the scientists to collaborate
and interact with the robot. Available modalities are \ac{VR}, desktop
application, and natural language understanding (see Figure~\ref{sharc2021:fig:interface}).
While the scientists on-board the ship served as pilots, the on-shore scientists
were assigned different parts of the XRF sampling procedure which they performed
using the three different interfaces provided by SHARC.
Under the constant supervision of the pilot, our sampling procedure began.

The first scientist, located in Chicago (IL), commanded the robot to equip the XRF by uttering the verbal command 
``pick up the \ac{XRF}''. The trajectory was computed on-board the ship, 
then sent back to the scientist for visualization. The scientist confirmed 
the trajectory, hence commanding the arm to execute the planned
trajectory.

\begin{figure}[t]
\centering
    \includegraphics[width=\columnwidth]{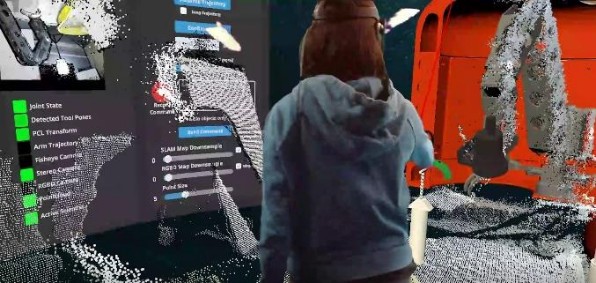}
    \caption{On-shore scientist using a VR system to choose a sampling location while immersed in the reconstructed point-cloud of the seafloor.}
    \label{sharc2021:fig:vr}
\end{figure}

Once the XRF was grasped, the second scientist, 
located in Cambridge (MA), used a 
\ac{VR} system (see Figure~\ref{sharc2021:fig:vr})
to identify the best sample location nearby the \ac{ROV}.
The use of \ac{VR} allowed the scientist to fully experience the three-dimensional
environment that was captured by cameras on-board the \ac{ROV} and reconstructed
into a point-cloud by the ship server.
The second scientist chose the sampling location by placing a virtual tool 
at the location they wanted to sample.
This triggered a planning request aboard the ship, which was executed by the 
robot and the trajectory sent back to the scientist for confirmation. The 
scientist confirmed the trajectory and the arm moved the XRF tool to the requested
location (see Figure~\ref{sharc2021:fig:xrf-sampling}).

With the XRF tool placed in the position of interest, the third on-shore 
scientist started the in-situ \ac{XRF} analysis of the sediment. 
This activated a stream of \ac{XRF} readings 
(shown in Figure~\ref{sharc2021:fig:xrf-data}) 
that started flowing from the \ac{XRF}, through the robot, the ship server, a satellite, the shore-server, and finally the scientist.
Once the scientist terminated the analysis, the XRF was returned to the tool tray,
effectively concluding the task.

\section{Conclusion}

In this work, we proposed SHARC, a framework for distributed remote shared 
autonomy. We also reported on the deployment of SHARC for an in-situ
\ac{XRF} analysis of seafloor sediments.

In our field study, we showed how, despite not having an expert in \ac{XRF}
sampling aboard the ship, SHARC enabled us to collect an in-situ XRF measurement with centimeter-level precision thanks to the combination of remote high-level human expertise and local low-level robot precision.
SHARC also enabled seamless collaboration and hand-off between multiple scientists geographically distributed.

An additional benefit of having scientists on-shore is easier access to 
computational resources that would be difficult to access on-board a ship. For example, SHARC's on-shore interface leverages cloud-based natural language processing for the parsing of verbal commands collected using a microphone.

It’s worth noting that while we tested SHARC on an \ac{XRF} sampling task, 
SHARC is a general tool that can support virtually any tool.
In fact, during our testing, two on-shore scientists used natural language commands and \ac{VR} controls to also collect a pushcore sample to validate the XRF measurements.

    \newpage
    \chapter{\moocsec}
    \label{mooc2021:sec:main}
    \begin{wrapfigure}{r!}{0.44\textwidth}
    \vspace{-0.4cm}
    \includegraphics[width=0.44\textwidth]{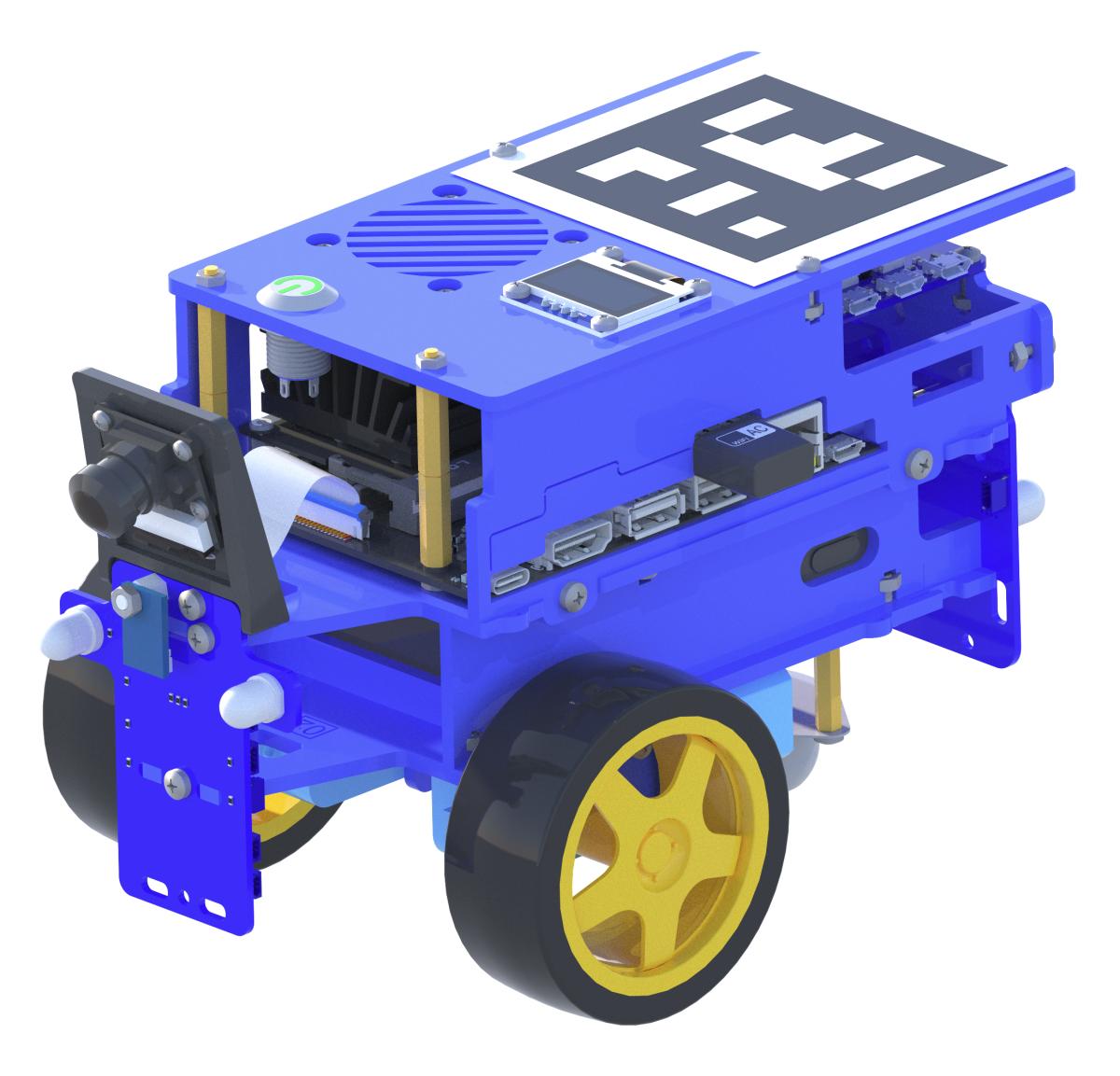}
    \caption{The DB21M robot}
    \label{mooc2021:fig:db21m-trimetric}
\end{wrapfigure}

\acfp{MOOC} are great tools for teaching to large audiences but they usually are completely virtual.
Robotics has a hardware component that cannot and must not be ignored or replaced by simulators.

``Self-driving cars with Duckietown'' is the first hardware-based \ac{MOOC} in AI and robotics. 
Offered for free on edX~\footnote{``Self-driving cars with Duckietown'' on edX: \url{https://www.edx.org/course/self-driving-cars-with-duckietown}}, it constitutes a practical introduction to vehicle autonomy with hands-on experiences.
It explores real-world solutions to the theoretical challenges of automation, including their implementation in algorithms 
and their deployment in simulation as well as on hardware. Using modern software architectures built with Python, \ac{ROS} and Docker, 
learners are exposed to the complementary strengths of classical architectures and modern machine learning-based approaches. 
The scope of this course is to go from zero to having a self-driving car safely driving on a miniature road.

Learners study autonomy hands-on by making real robots take their own decisions and accomplish broadly defined tasks. 
Step by step from the theory, to the implementation, to the deployment in simulation as well as on a real robot.
In this course, learners start from a box of parts and finish with a scaled self-driving car that drives autonomously in a miniature town.
The whole process is designed to expose learners to state-of-the-art approaches, the latest software tools and real hardware in an 
engaging hands-on learning experience.

This course offers access to a comprehensive robotics curriculum that is otherwise hard to be exposed to. By leveraging 
an easy-to-use platform for online teaching, this course aims at lowering the barrier of entry to the world of robotics for people 
with diverse backgrounds.

Presented by Professors and Scientists who are passionate about robotics and accessible education, this course uses the Duckietown robotic ecosystem~\cite{duckietown_website}.
This course was made possible thanks to the support of the Swiss Federal Institute of Technology in Zurich (ETH Zurich), in collaboration with the University of Montreal (Prof. Liam Paull), the Duckietown Foundation, and the Toyota Technological Institute at Chicago (Prof. Matthew Walter).

\section{Learning Goals}

The curriculum was built around a list of target learning goals:
\begin{itemize}
    \item recognize essential robot subsystems (sensing, actuation, computation, memory, mechanical) and their functions;
    \item make a robot drive in arbitrary user-specified paths;
    \item understand how to command a robot to reach a goal position;
    \item learn autonomous decision-making according to ``traditional approaches'' (e.g., estimation, planning, control);
    \item learn autonomous decision-making according to ``modern approaches'' (e.g., imitation learning, reinforcement learning);
    \item process streams of images;
    \item be able to set up an efficient software environment for robotics with state-of-the-art tools (Docker, \ac{ROS}, Python);
    \item program a robot to make it safely drive in empty roads lanes;
    \item program a robot to make it recognize obstacles such as rubber duckies;
    \item program a robot to make it avoid obstacles such as rubber duckies;
    \item program a robot to make it safely drive down roads with pedestrian duckies;
    \item independently assemble a robot (i.e., Duckiebot) and a Duckietown;
    \item be able to discuss differences between theory, simulation, and real word implementation for different approaches;
\end{itemize}

\section{First Edition - 2021}

In March 2021, the first edition of the \ac{MOOC} was launched on edX. It lasted four months and counted more than $7$ thousand 
learners enrolled. 

\subsection{The DB21M robot}
\label{mooc2021:sec:db21m}

\begin{figure}[!t]
    \centering
    \includegraphics[width=0.9\textwidth, trim = 0 3cm 2.6cm 1cm, clip]{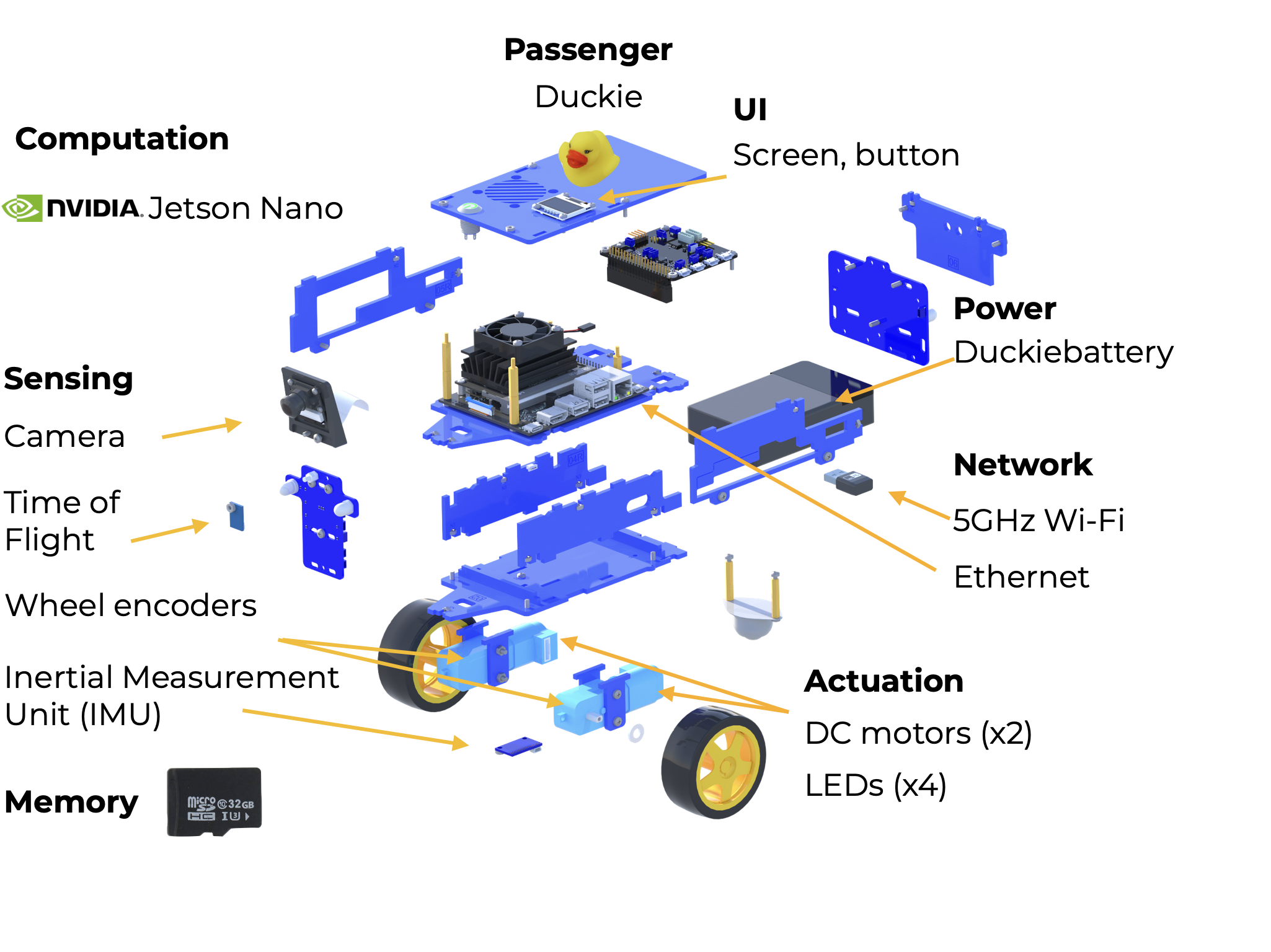}
    \caption{An exploded view of the DB21M robot}
    \label{mooc2021:fig:db21m-exploded}
\end{figure}

Within the scope of the first edition of the course, a new robot was developed, the DB21M (shown in Figure~\ref{mooc2021:fig:db21m-trimetric} and~\ref{mooc2021:fig:db21m-exploded}), 
specifically tailored to the \acs{MOOC} experience.
Powered by Nvidia’s Jetson Nano onboard computer, the DB21M was the first AI-capable robot developed by Duckietown 
and a platform for learners to see their AI algorithms come to life.\\

Unlike its predecessor (i.e., DB19), the DB21M robot featured:
\begin{itemize}
    \item a \acf{ToF} sensor mounted on the front of the chassis
pointing forward, with a maximum range of $2m$, a maximum frequency of $50Hz$, and a field-of-view of $25\deg$;
    \item an \acf{IMU} board hosting both a gyroscope and an accelerometer;
    \item an LCD display and a button on the top plate for interfacing with the user;
    \item a dedicated CUDA-capable hardware neural engine accelerator for \acf{ML} applications;
    \item a fan for active cooling;
\end{itemize}

\section{Second Edition - 2022}

In November 2022, the second edition of the \ac{MOOC} was launched on edX. Unlike the first iteration, this instance of the course is 
self-paced and will remain open for one year.

\subsection{The DB21J robot}
\label{mooc2021:sec:db21j}

In order to improve the learning experiences around the topic of \acf{ML}, a revised version of the robot was released. The DB21J
keeps all the hardware and characteristics of the DB21M, but uses an Nvidia’s Jetson Nano 4GB as onboard computer, effectively increasing
the amount of memory available to the on-board AI processor (GPU) by a factor of $3$.

\subsection{Towards a full browser-based robot programming experience}
Robotics is hard, but the path to it does not have to be. While robotics inevitably inherits the complexity of the fields it builds upon, some characteristics of the process of
developing robots contribute to an artificially higher barrier of entry for newbies.
Some of these characteristics built up over the years out of personal preferences and engineering inertia. 

\subsubsection{Linux is rare}
The most popular tools used for robots development only target \emph{specific distributions} of the Linux operating system. 
As of May 2022, \acf{ROS} is the most popular framework
for robots development, with Ubuntu Linux being the only officially supported operating system. At the same time, Linux is reported to have a global market share of
$2.09\%$ among desktop operating systems, with Ubuntu Linux having a global market share of only $0.93\%$  (source \cite{linux-statistics}).
This constitutes an artificial barrier of entry, as the technologies ROS builds upon are not unique to the Linux operating system.

\subsubsection{Web Browsers are abundant}
While Linux is rare, web browsers are abundant. They are not just available out-of-the-box on any operating system, they are available on virtually any modern device, such as smartphones, tablets, etc.
In 2017, I started developing $\backslash compose \backslash$~\cite{compose}, an open-source web-based \acf{CMS} that provides common functionalities for the fast development and deployment of web applications. 
While it started as a personal project, today, $\backslash compose \backslash$ powers the Duckietown Dashboard, a web-based
monitoring and control interface running on thousands of robots across the globe. While the Duckietown robots use ROS as the underlying programming framework, 
with the Duckietown Dashboad we are challenging the idea that Linux is compulsory to learning about robotics.

\subsubsection{The missing piece}
\begin{figure}[t!]
\centering
    \includegraphics[trim = 0 10cm 0 0, clip, width=0.94\columnwidth]{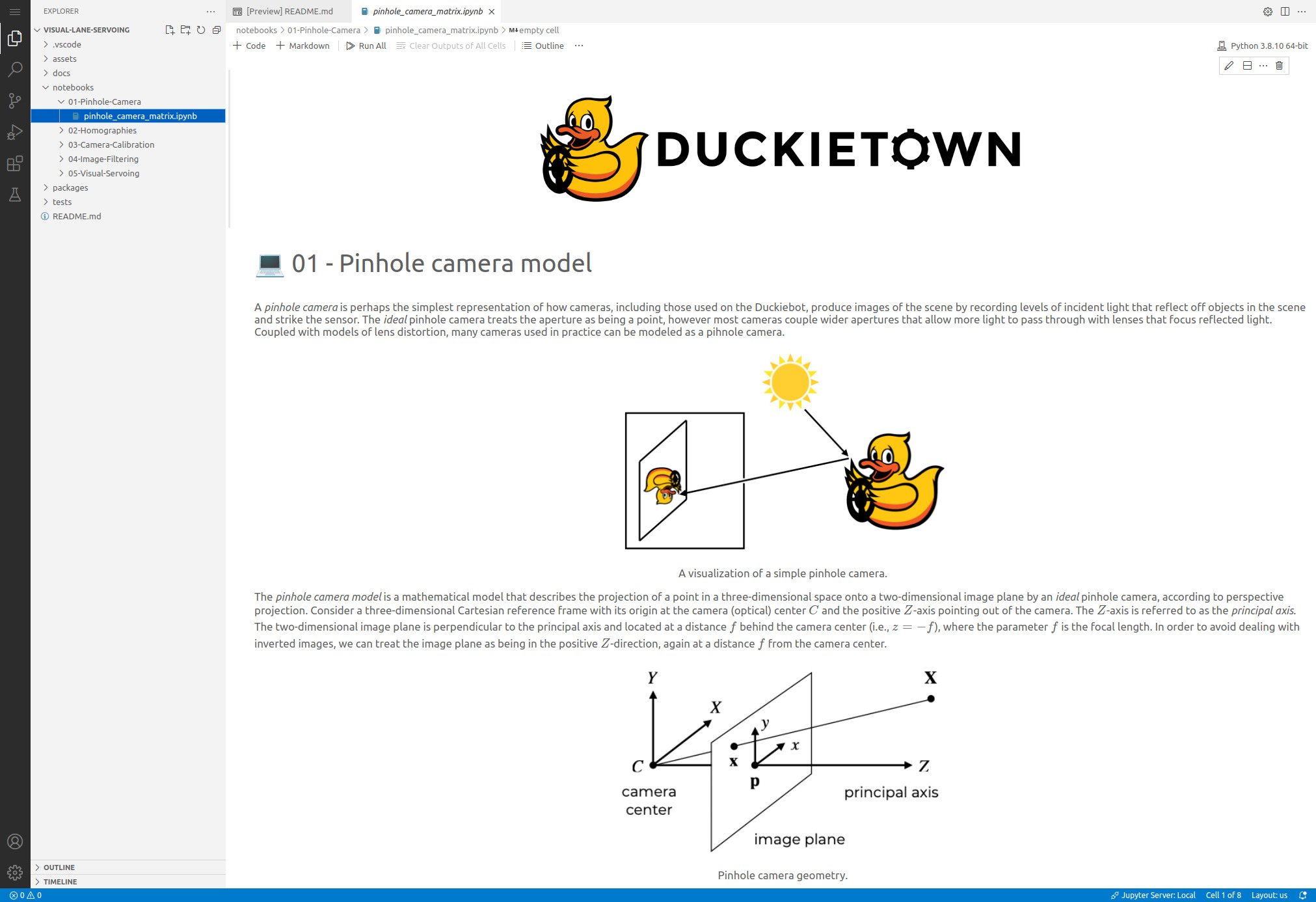}
    \caption{Instance of code-server showing a Jupyter notebook for the ''RoboVision`` exercise.}
    \label{mooc2021:vscode-notebook}
\end{figure}

\begin{figure}[t!]
\centering
    \includegraphics[trim = 0 14cm 0 0, clip, width=0.94\columnwidth]{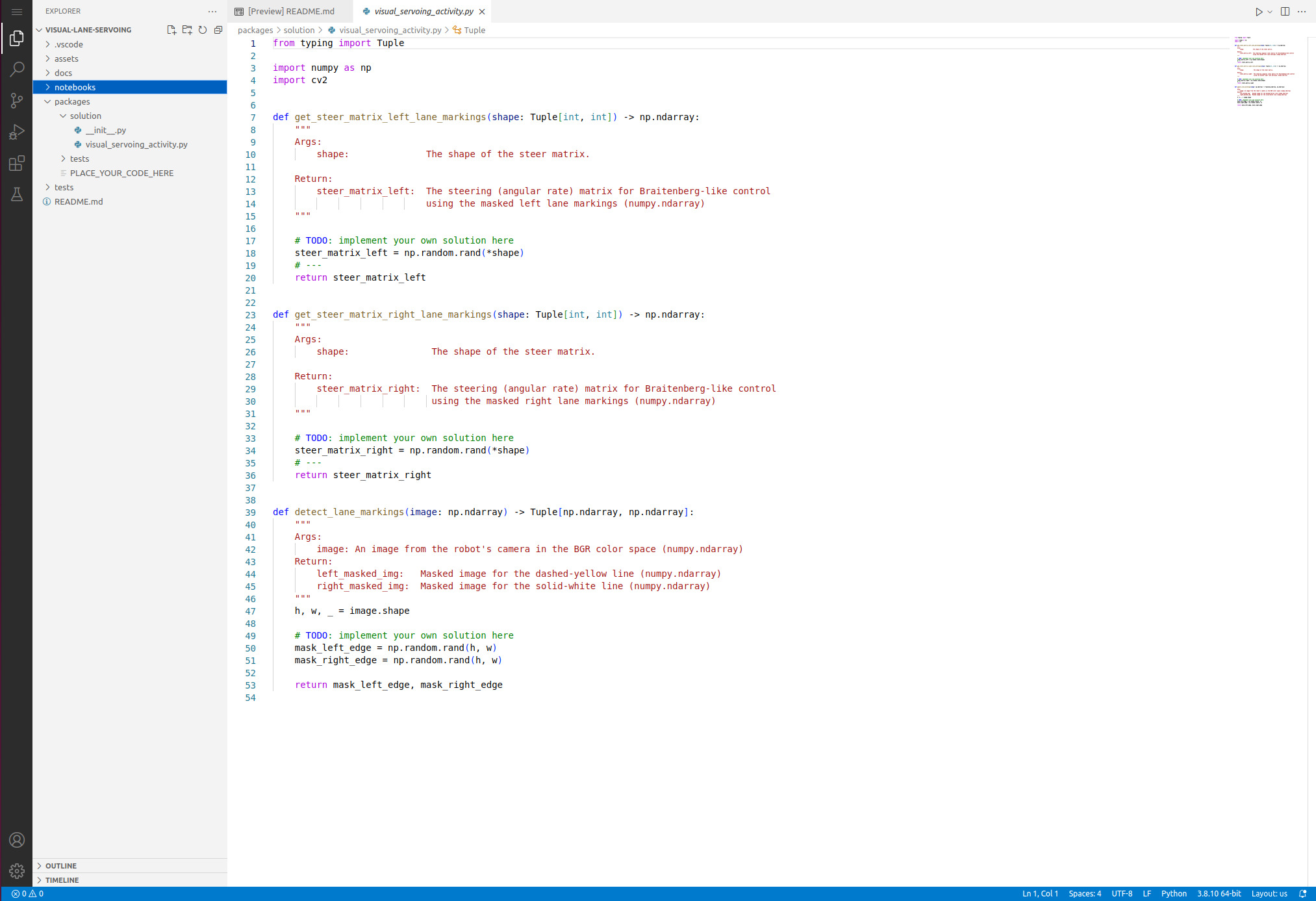}
    \caption{Instance of code-server showing a code editor tab for the ''RoboVision`` exercise.}
    \label{mooc2021:vscode-code-editor}
\end{figure}

The capstone needed to achieve a full browser-based robot programming experience was an interface that would also allow users to 
write code and deploy it to the onboard computer directly through a web browser. 
A web-based port of the popular \acf{IDE} Visual Studio Code~\footnote{\url{https://code.visualstudio.com/}}, called 
code-server~\footnote{\url{https://github.com/coder/code-server}} was integrated into the development workflow for the second
edition of the class, effectively enabling a full-featured coding experience delivered through the web browser 
(see Figures~\ref{mooc2021:vscode-notebook} and \ref{mooc2021:vscode-code-editor}).

\begin{figure}[t!]
\centering
    \includegraphics[width=0.95\columnwidth]{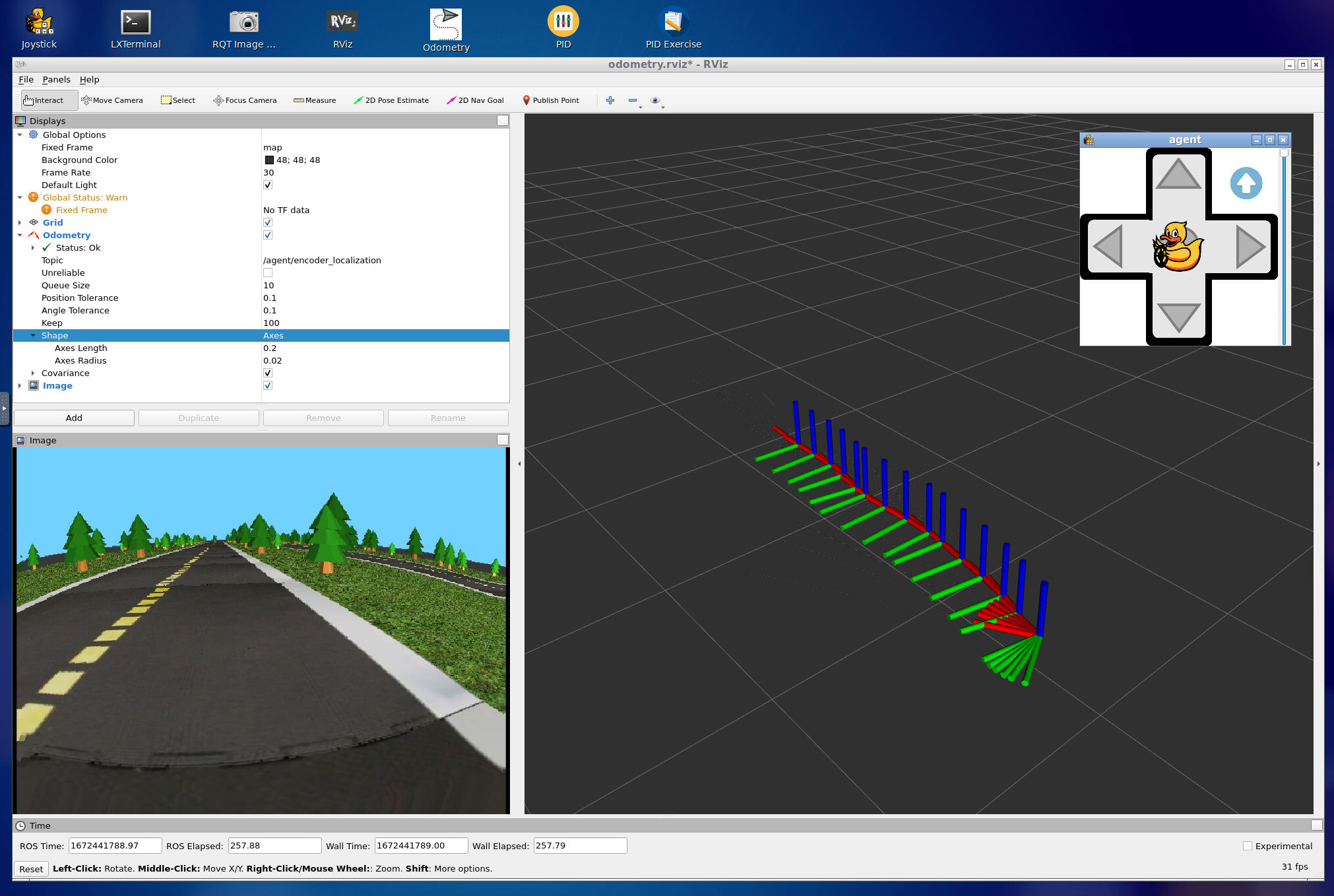}
    \caption{VNC session hosting RViz and the Duckietown Joystick for the Odometry exercise}
    \label{mooc2021:vnc-odometry}
\end{figure}

The web-based development workflow we built brings together many of the most commonly used tools for teaching and robotics
development available. VSCode provides a powerful and well-tested development environment for code editing 
(Figure~\ref{mooc2021:vscode-code-editor}), Jupyter Notebooks~\footnote{\url{https://jupyter.org/}} computational documents
(Figure~\ref{mooc2021:vscode-notebook}), version control
systems with Git, unit testing, and more. VNC allows learners to interact with a pre-packaged Linux desktop environment
running graphical interfaces for tools like RViz (a 3D viewer from the Robot Operating System framework), the Duckietown
Joystick, and more (see Figure~\ref{mooc2021:vnc-odometry}).

    \chapter{Conclusion}
    \label{sec:conclusion}
    This thesis is a collection of contributions that span across different research areas and target audiences.
At the core of all these contributions is the unwavering interest in making robots more accessible.

In Chapters~\ref{hri2017:sec:main} and \ref{isrr2017:sec:main}, we tackled the problem of improving robots'
accessibility by enabling them to ``talk'' to their human companions. We looked at the problem of translating
information that a robot possesses about how to navigate an indoor environment into natural language instructions that 
humans can successfully comprehend and follow (Chapter~\ref{hri2017:sec:main}). 
We then looked at a related problem, that is, translating natural language
sentences uttered by humans into an internal representation that the robot can process and consume while
learning about how to operate articulated objects (Chapter~\ref{isrr2017:sec:main}).

In the second part of this document, we turned our attention to the effects of accessibility in robotics research.
In Chapter~\ref{iros2020:sec:main}, we proposed a new concept for reproducible
robotics research that integrates development and benchmarking, so that reproducibility is obtained ``by design''. 
We started by providing the overall conceptual objectives to achieve 
this goal and then explored a concrete instantiation of our proposed approach: the DUCKIENet (Chapter~\ref{aido2018:sec:main}).
We then moved on to the problem of improving accessibility for underwater robotic intervention operations 
(Chapter~\ref{sharc2021:sec:main}).
We did so by proposing a new framework, called SHARC (SHared Autonomy for Remote Collaboration), which leverages
recent advancements in the field of shared autonomy and a distributed architecture to allow multiple remote 
scientists to efficiently plan and execute high-level sampling procedures using an underwater manipulator 
while being on-shore and at thousands of kilometers away from one another and the robot. 

Finally, we turned our attention to the impact of accessible robotic platforms and interfaces in the context of 
educational robotics. 
In Chapter~\ref{mooc2021:sec:main}, we discussed the development of the first hardware-based MOOC in AI and robotics. 
At the time of writing, the second edition of the MOOC is still ongoing with thousands of learners expected
to join before it ends in November 2023. We are committed to working on this effort and improving 
the accessibility of these learning materials for as long as there is an interested audience.

    \newpage
    \bibliographystyle{abbrvnat}
    \bibliography{99.bibliography.bib, papers/hri.2017/references.bib, papers/isrr.2017/references.bib, papers/aido.2018/references.bib, papers/iros.2020/references.bib, content/sharc.2021/references.bib}
\end{document}